\pdfoutput=1

\documentclass[11pt]{article}

\usepackage[preprint]{acl}
\usepackage{times}
\usepackage{latexsym}

\usepackage[T1]{fontenc}

\usepackage[utf8]{inputenc}

\usepackage{microtype}

\usepackage{inconsolata}

\usepackage{enumitem}
\usepackage{times}
\usepackage{latexsym}
\usepackage{bbding}
\usepackage{graphicx}
\usepackage{subfigure}
\usepackage{caption}
\usepackage{hyperref}
\usepackage{amsmath}
\usepackage{tablefootnote}
\usepackage{threeparttable}
\usepackage{booktabs}
\usepackage{tabularx}
\usepackage{adjustbox}
\usepackage{algorithm}
\usepackage{algorithmic}
\usepackage{multirow}
\usepackage{color}
\usepackage{colortbl}
\usepackage{microtype}
\usepackage{graphicx}
\usepackage{subcaption}
\usepackage{hyperref}
\usepackage{graphicx}
\definecolor{myblue1}{RGB}{204, 255, 255}
\definecolor{mygreen1}{RGB}{204,255,204}
\usepackage{xcolor}
\definecolor{color1}{RGB}{255,245,245}
\definecolor{color2}{RGB}{255,227,227}
\definecolor{color3}{RGB}{255,201,201}
\definecolor{color4}{RGB}{165,224,255}
\definecolor{color5}{RGB}{124,188,230}
\definecolor{color6}{RGB}{83,152,205}
\definecolor{color7}{RGB}{41,115,180}
\definecolor{color8}{RGB}{0,79,155}
\usepackage{pifont}

\newcommand{\selfpt}{\textsc{Self-Tuning}}
\newcommand{\selfteach}{\textsc{Self-Teaching}}
\newcommand{\pit}{\textsc{PIT}}

\newcommand{\llamaa}{\textsc{Llama2-7B}}

\newcommand{\ie}[0]{\emph{i.e., }}

\newcommand{\eg}[0]{\emph{e.g., }}

\newcommand{\etc}[0]{\emph{etc.}}
\newcommand{\aka}[0]{\emph{a.k.a. }}

\newcommand{\wrt}[0]{\emph{w.r.t., }}
\newcommand{\RN}[1]{%
	\textup{\lowercase\expandafter{\it \romannumeral#1}}%
}
\usepackage{inconsolata}
\usepackage{amsmath}
\usepackage{amssymb}
\usepackage{makecell}
\usepackage{soul} 
\usepackage{arydshln}
\usepackage{booktabs}
\usepackage{multirow}
\usepackage{float}
\usepackage{color, xcolor}
\definecolor{mygrey1}{RGB}{204,229,255}
\definecolor{myblue1}{RGB}{204, 255, 255}
\definecolor{mygreen1}{RGB}{204,255,204}
\definecolor{myyellow1}{RGB}{230,255,204}
\definecolor{mylightyellow1}{RGB}{255,255,204}
\definecolor{mypink1}{HTML}{e6cfec}
\newcommand{\llamaachat}{\textsc{Llama2-7B-chat}}
\newcommand{\llamaathirteen}{\textsc{Llama2-13B}}

\newcommand\addfootnote[1]{%
  \begingroup
  \renewcommand\thefootnote{}\footnote{#1}%
  \addtocounter{footnote}{-1}%
  \endgroup
}

\usepackage{tikz}

\newcommand*\circled[1]{\tikz[baseline=(char.base)]{
            \node[shape=circle,fill=black,text=white,inner sep=0.5pt] (char) {#1};}}
\usepackage{lmodern}
\usepackage{mathptmx}
\title{\selfpt{}: Instructing LLMs to Effectively Acquire New Knowledge through Self-Teaching}

\author{Xiaoying Zhang$^{1*}$, Baolin Peng$^{2}$, Ye Tian$^{2}$, Jingyan Zhou$^{1}$, Yipeng Zhang\\ 
 \bf  Haitao Mi$^{2}$, Helen Meng$^{1,3}$ \\
    $^{1}$The Chinese University of Hong Kong, Hong Kong\\
    $^{2}$Tencent AI Lab, Bellevue\\
    $^{3}$Centre for Perceptual and Interactive Intelligence, Hong Kong\\
    \{zhangxy, jyzhou, hmmeng\}@se.cuhk.edu.hk, yipengzhang97@gmail.com  \\ 
    \{baolinpeng, yaptian, haitaomi\}@global.tencent.com \\ 
}

\begin{document}
\maketitle
\begin{abstract}
Large language models (LLMs) often struggle to provide up-to-date information due to their one-time training and the constantly evolving nature of the world. To keep LLMs current, existing approaches typically involve continued pre-training on new documents. However, they frequently face difficulties in extracting stored knowledge. 
Motivated by the remarkable success of the Feynman Technique in efficient human learning, we introduce \selfpt{}, a learning framework aimed at improving an LLM's ability to effectively acquire new knowledge from unseen raw documents through self-teaching. 
Specifically, we develop a \selfteach{} strategy that augments the documents with a set of knowledge-intensive tasks created in a self-supervised manner, focusing on three crucial aspects: \textit{memorization}, \textit{comprehension}, and \textit{self-reflection}. Additionally, we introduce three Wiki-Newpages-2023-QA datasets to facilitate an in-depth analysis of an LLM's knowledge acquisition ability concerning \textit{memorization}, \textit{extraction}, and \textit{reasoning}. Extensive experimental results on various models, \eg \llamaa{} reveal that \selfpt{} consistently exhibits superior performance across all knowledge acquisition tasks and excels in preserving previous knowledge.\addfootnote{$^*$Work done during the internship at Tencent AI Lab. 
\hspace{3mm}}\footnote{Our source code and the curated new datasets can be accessed publicly at \url{https://github.com/zhangxy-2019/Effective-Knowledge-Injection}.}





\end{abstract}

\section{Introduction}
\begin{figure}[t]
\centering
\includegraphics[width=0.95\columnwidth]{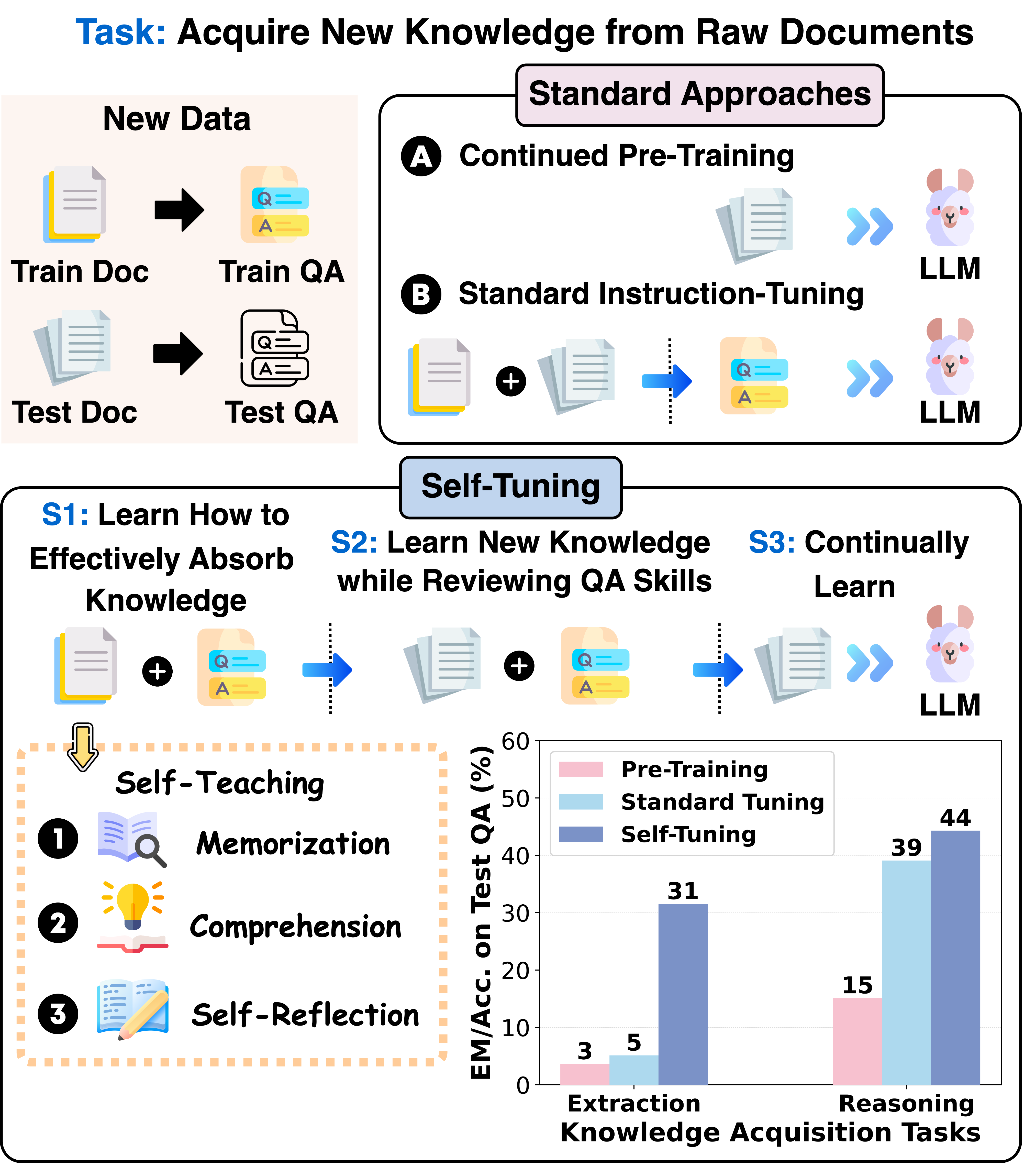}
\caption{Illustration of the knowledge acquisition task with two standard knowledge injection approaches (in the upper part). Depiction of \selfpt{} for effective knowledge acquisition from unseen raw documents, which significantly enhances factual accuracy compared to the standard approaches (in the lower part).}

\label{fig:intro_ex}
\end{figure}

Armed with a wealth of factual knowledge acquired during the pre-training phase~\cite{zhou2023lima, zhang2025will}, LLMs~\cite{touvron2023llama, openai2023gpt4} exhibit remarkable proficiency in numerous knowledge-intensive tasks~\cite{cohen-etal-2023-crawling, zhang2024self, zhang-etal-2024-question, gekhman2024does}. Despite this, the knowledge stored in LLMs can quickly become outdated due to the one-time training of LLMs and the ever-changing nature of the world~\cite{zhang-etal-2022-toward-self, zhang-etal-2023-sgp, huang2023survey, jiang2024instructiontuned}. These unavoidable knowledge limitations present notable obstacles to the trustworthiness of LLMs in real-world scenarios~\cite{liu2023trustworthy,mecklenburg2024injecting}. Thus, it is essential to equip LLMs with new knowledge to keep them up-to-date.


In this paper, we focus on injecting new knowledge into the parameters of LLMs. As depicted in the upper part of Figure~\ref{fig:intro_ex}, a standard approach involves continued pre-training (A) on a raw corpus (here, test doc) containing new information~\cite{jang2022towards}. However, it struggles to extract the embedded knowledge, potentially due to the impaired question-answering (QA) capability~\cite{allenzhu2023physics,cheng2024adapting}. 
Despite the assistance of subsequent instruction-tuning (B)~\cite{wei2022finetuned, ouyang2022training} on QA data, the knowledge retrieved from the LLMs remains notably constrained~\cite{jiang2024instructiontuned}. 
Recently, \citet{jiang2024instructiontuned} suggests fine-tuning on a mix of QA data and related documents before continuing pre-training, with the aim of teaching the model how to access knowledge from documents and answer questions. Although this method greatly outperforms standard approaches, our initial results suggest that its effectiveness in knowledge extraction remains limited.


Numerous studies \cite{ambion2020utilization, article} evidence the effectiveness of the Feynman Technique~\cite{xiaofei2017application} in promoting human learning and knowledge understanding. The remarkable success of this potent learning method is often attributed to its emphasis on ``comprehension,'' ``self-reflection'' (``identifying gaps and review''), rather than mere ``memorization''. This encourages our exploration into its potential application in improving LLMs' knowledge acquisition capabilities. As a result, we present \selfpt{}, a framework that empowers an LLM to effectively internalize and recall new knowledge. As depicted in the lower part of Figure~\ref{fig:intro_ex}, \selfpt{} consists of three stages: $(\RN{1})$ Firstly, we train the model using a mix of training documents and associated QA data, equipping it with the ability to efficiently absorb knowledge from raw documents via self-teaching, as well as question-answering skills. Specifically, we design a \selfteach{} strategy to present the training documents as plain texts for \textit{memorization} and a series of knowledge-intensive tasks derived from the documents in a self-supervised manner, without any mining patterns \cite{van-de-kar-etal-2022-dont}, for \textit{comprehension} and \textit{self-reflection}. $(\RN{2})$ Next, we deploy the model to apply the learning strategy for spontaneously acquiring knowledge from new documents while reviewing its QA skills.  $(\RN{3})$ Finally, we continue training the model using only the new documents to ensure thorough acquisition of new knowledge.

In addition, we introduce three Wiki-Newpages-2023-QA datasets to conduct an in-depth study of how an LLM acquires new knowledge \wrt \textit{memorization}, \textit{extraction}, and \textit{comprehension} (in this study, \textit{reasoning}) across single-domain, multi-domain, and cross-domain settings. These datasets are carefully curated to ensure minimal overlap with the LLM's pre-training corpora, emphasizing two key knowledge-intensive tasks, \ie open-ended generation and natural language inference (NLI) tasks. Extensive experimental results on diverse models, \eg \llamaa{}~\cite{touvron2023llama2openfoundation}, Qwen2-7B~\cite{yang2024qwen2technicalreport}, and Mistral-7B-v0.1~\cite{jiang2023mistral7b} demonstrate that \selfpt{} significantly outperforms all other compared methods on knowledge memorization and extraction tasks. In addition, \selfpt{} consistently yields high accuracy on reasoning tasks, while the performance of the compared methods largely fluctuates in different scenarios. Inspiringly, \selfpt{} exhibits exceptional performance in retaining previously acquired knowledge (\ie knowledge retention) concerning extraction and reasoning on two well-established benchmarks.


In summary, our contributions are three-fold:

\begin{itemize}\setlength{\itemsep}{0pt}
\item We present \selfpt{}, a framework designed to improve an LLM's knowledge acquisition capability via self-teaching.

\item We introduce three Wiki-Newpages-2023-QA datasets to enable a comprehensive analysis of an LLM's knowledge acquisition ability \wrt memorization, extraction, and reasoning.



\item We validate the efficacy of \selfpt{} on three crucial knowledge acquisition tasks using the Wiki-Newpages-2023-QA datasets.

\end{itemize}

\begin{table*}[!t]
\setlength\tabcolsep{0.5pt}
  \centering
  \begin{threeparttable}
  \fontsize{9}{9}
  \selectfont
    \begin{tabular}{lcccccccc} 
    \toprule
    \multirow{3}{*}{\textbf{Wiki-Newpages}}& \multirow{3}{*}{\makecell[c]{\textbf{Factual} \\ \textbf{Knowledge}}}&
     \multicolumn{2}{c}{\textbf{Open-Ended Generation (Train \& Test Sets)}}
    &\multicolumn{2}{c}{\textbf{NLI (Test Set)}}\cr\cmidrule(lr){3-4} \cmidrule(lr){5-6}
                                              &      & \textbf{Statistics} & \textbf{Avg. \# Tokens} & \textbf{Statistics} & \textbf{Answer Type} \cr 
    \midrule
    \multirow{3}{*}{\makecell[c]{Wiki-Bio \\(Single-domain)}} & \multirow{3}{*}{\makecell[c]{Birth Date,\\ Profession, \\Education, \etc}}&\multirow{3}{*}{\makecell[c]{Train: 6,136 (\# QA); 1,136 (\# Docs)\\
    Test: 663 (\# QA); 127 (\# Docs)}} & \multirow{3}{*}{\makecell[l]{ 8.34 (Q)\\ 4.24 (A) \\59.64 (Doc) }} & \multirow{3}{*}{\makecell[l]{729 (\# QA)\\ 127 (\# Docs) }} & \multirow{3}{*}{\makecell[l]{Yes (65.84\%)\\ No (33.47\%) \\ Impossible (0.69\%)}} \cr \\ \cr
    \midrule
    \multirow{3}{*}{\makecell[c]{Wiki-Multi \\(Multi-domain)}}& \multirow{3}{*}{\makecell[c]{News, \\ TV Series,\\Sports, \etc}}&\multirow{3}{*}{\makecell[c]{Train: 10,004 (\# QA); 1,823 (\# Docs)\\
    Test: 1,502 (\# QA); 281 (\# Docs)}} & \multirow{3}{*}{\makecell[l]{ 10.13 (Q)\\ 5.70 (A) \\ 69.25 (Doc) }} & \multirow{3}{*}{\makecell[l]{1,627 (\# QA)\\ 281 (\# Docs) }} & \multirow{3}{*}{\makecell[l]{Yes (60.97\%)\\ No (36.63\%) \\ Impossible (2.40\%)}}\cr \\ \cr
    \midrule
    \multirow{3}{*}{\makecell[c]{Wiki-Film\\(Single-domain)}} & \multirow{3}{*}{\makecell[c]{Genre, Language,\\Director,\\ Released Time, \etc}}&\multirow{3}{*}{\makecell[c]{
    Test: 955 (\# QA); 169 (\# Docs)}} & \multirow{3}{*}{\makecell[l]{ 8.83 (Q)\\ 4.61 (A) \\ 58.10 (Doc) }} & \multirow{3}{*}{\makecell[l]{1,387 (\# QA)\\ 169 (\# Docs) }} & \multirow{3}{*}{\makecell[l]{Yes (62.73\%)\\ No (26.53\%) \\ Impossible (10.74\%)}} \cr \\ \cr
    \bottomrule  
    \end{tabular}
  \end{threeparttable}
  
  \caption{Statistical information of three Wiki-Newpages-2023-QA datasets, \ie Wiki-Bio, Wiki-Multi, and Wiki-Film. ``Impossible'': ``It's impossible to say''. Details about token count distribution can be found in Appendix \ref{sec:token_distribution}.}
  \label{tab:wiki_data}
  \vspace{-4mm}
\end{table*}

\section{Related Work}


\paragraph{Continual Knowledge Injection.} 
The primary research approach for injecting new knowledge into LLMs \cite{DBLP:journals/corr/abs-2302-09170, ovadia2024finetuning, mecklenburg2024injecting} is through continued pre-training. This method entails the ongoing pre-training of LLMs on raw corpora containing new knowledge, carried out in a causal auto-regressive manner \cite{allenzhu2023physics, ibrahim2024simple, ovadia2024finetuning}. However, this straightforward approach often encounters hurdles in effectively enabling LLMs to extract the acquired knowledge during the inference phase \cite{allenzhu2023physics, jiang2024instructiontuned, cheng2024adapting}. To enhance knowledge extraction, instruction tuning on QA data after pre-training has been extensively employed \cite{wei2022finetuned, DBLP:conf/nips/Ouyang0JAWMZASR22}. \citet{jiang2024instructiontuned} suggests that the effectiveness of this method remains limited, and proposes fine-tuning the model on QA data before continued pre-training. This instructs the model on how to retrieve knowledge from raw corpora, thereby enhancing knowledge extraction. However, such an approach tends to underestimate the importance of comprehending the new knowledge.

Acknowledging the value of knowledge comprehension, \citet{cheng2024adapting} proposes converting raw corpora into reading comprehension texts. This approach, however, focuses on domain adaptation and preserving general prompting abilities by mining a set of instruction-following tasks from the document content.
In contrast, our work aims to equip the model with the ability to effectively absorb new knowledge from raw documents and employ the learned ability to unseen documents. Specifically, we develop a \selfteach{} strategy to present the raw document as plain texts for memorization, accompanied by a set of tasks for comprehension and self-reflection, which are created based on raw corpora in a self-supervised manner, without relying on any mining patterns.

Additionally, \textbf{knowledge editing}~\cite{DBLP:conf/iclr/MitchellLBFM22, DBLP:conf/emnlp/ZhengLDFWXC23, yao-etal-2023-editing, jiang2024learning, zhang2024comprehensive} and \textbf{retrieval-augmented generation}~\cite{lewis2021retrievalaugmented, vu2023freshllms, ovadia2024finetuning, jeong2024adaptiverag} are recognized as two related research fields. We provide more details on these initiatives in Appendix \ref{sec:additional_related_work}. 

\section{Wiki-Newpages-2023-QA: Datasets for Studying LLM Knowledge Acquisition }
To explore the knowledge acquisition capabilities of LLMs from new documents, \wrt memorization, extraction and reasoning, we introduce the Wiki-Newpages-2023-QA datasets (Table \ref{tab:wiki_data}), which are carefully designed to minimize overlap with the initial pre-training corpus. These datasets comprise new document corpora for studying knowledge memorization and associated QA datasets for two vital knowledge-intensive tasks: open-ended generation and NLI for examining extraction and reasoning, respectively. Due to space constraints, we provide a brief overview of the dataset construction process here, with the complete version available in Appendix \ref{sec:details_data_construction}.

\subsection{Document Collection and QA Pair Generation} 
\paragraph{Document Collection.}
To construct the document corpus, we collect articles from September to October 2023 (4,257 articles in total) from Wikipedia NewPages\footnote{\url{https://en.wikipedia.org/wiki/Special:NewPages}}, which include new articles from various domains published after the pre-training cut-off time of the LLMs being evaluated.\footnote{The pre-training cut-off time for the \textsc{Llama2} family models used in this study is 2022.} Following \citet{jiang2024instructiontuned}, we only use the first paragraph of each article, as it offers a comprehensive summary and contains a wealth of factual information.
\paragraph{QA Pair Generation.} 
We gather QA pairs for generation and NLI tasks using our handcrafted prompts in Tables \ref{tab:wikiqa_gen_prompt} and \ref{tab:wikiqa_nli_prompt}, aiming to cover all factual information within the given document.
\subsection{Splitting} 
We construct three datasets for single-domain, multi-domain, and cross-domain analysis, splitting them into training and testing subsets while \textit{ensuring zero knowledge overlap}.
\paragraph{Dataset Splitting.}
We generate three datasets: Wiki-Newpages-2023-10-Bio (Wiki-Bio), Wiki-Newpages-2023-10-Multi (Wiki-Multi), and Wiki-Newpages-2023-(9)10-Film (Wiki-Film) by randomly selecting 1,263 biographical documents, 2,104 multi-domain documents, and 955 film documents from the collected document corpus and their associated QA pairs.
\paragraph{Train-test Splitting.} 
We divide Wiki-Bio and Wiki-Multi datasets into training and testing subsets for single-domain and multi-domain evaluations. We use Wiki-Film as the test set for cross-domain scenarios. Note that the training QA datasets only include open-ended generation task pairs, ensuring fair comparisons. 

\section{\selfpt{}}

\begin{figure*}[!t]
\centering
\includegraphics[width=0.95\linewidth]{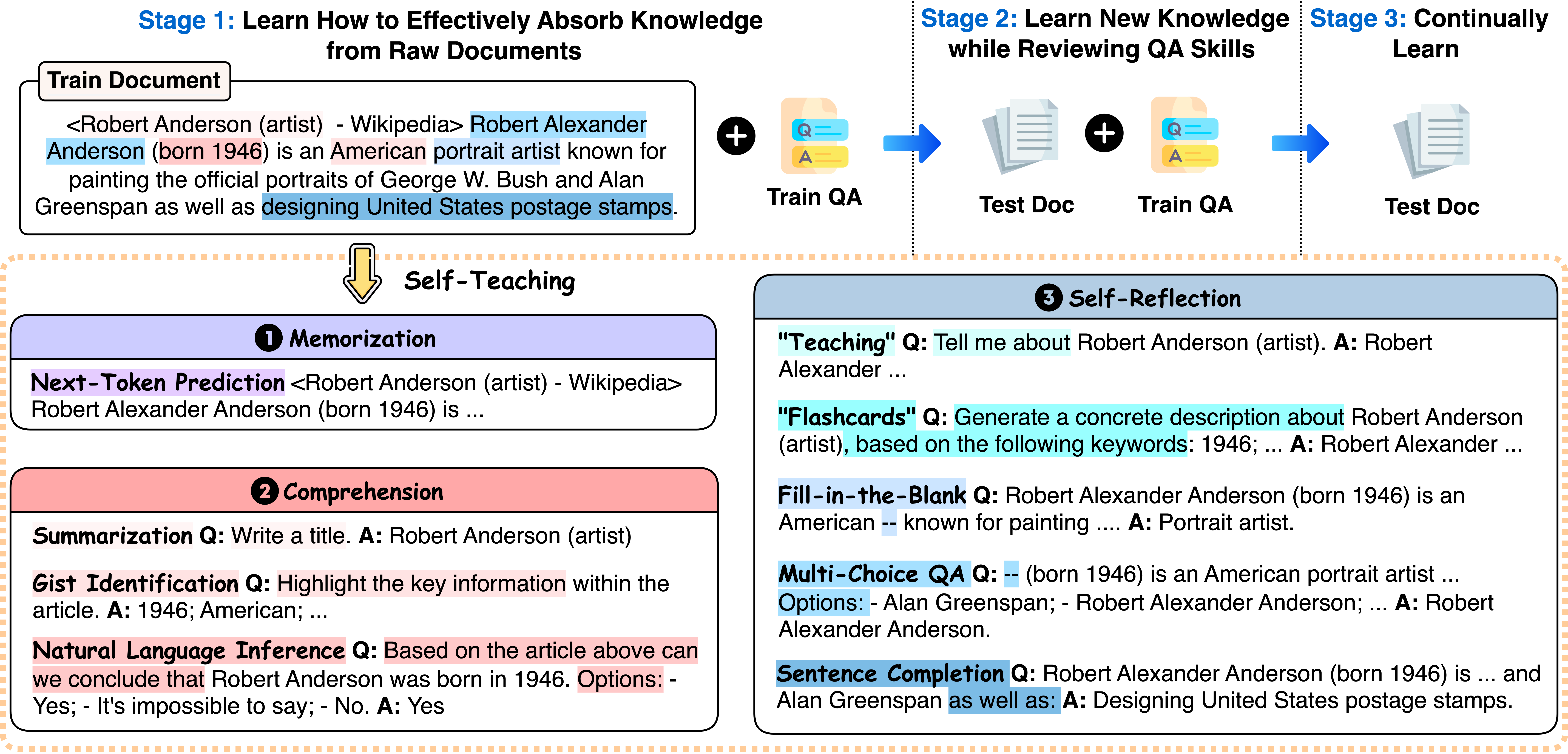}
\caption{Illustration of the proposed \selfpt{}. The framework consists of three stages (in the upper part): $(\RN{1})$ Equipping the model with the ability to deeply absorb knowledge from raw documents using the proposed \selfteach{} strategy (in the lower part), along with question-answering capabilities; $(\RN{2})$ Applying the learning strategy acquired in Stage 1 to obtain new knowledge from unseen documents and refining QA skills; $(\RN{2})$ Continuously learning from unseen documents. See Appendix \ref{sec:example_doc_task} for the full training document example in Stage 1.}
\label{fig:framework}
\vspace{-2.5mm}
\end{figure*}


\vspace{-1mm}
In this section, we introduce the \selfpt{} framework to improve the LLM's capability to acquire knowledge from new documents, with the devised \selfteach{} strategy.
We first give an overview of the training process for knowledge acquisition using the proposed \selfpt{} in Section \ref{sec:method_overview}. Then, we delve into the \selfteach{} strategy in Section \ref{sec:self-exam}.


\vspace{-1mm}
\subsection{Overview} 
\label{sec:method_overview} 
As depicted in Figure \ref{fig:framework}, the proposed \selfpt{} comprises the following three stages.
\vspace{-2mm}
\paragraph{Stage 1: Learn How to Effectively Absorb Knowledge from Raw Documents.} Our objective is to equip an LLM $M$, parameterized by $\theta$, with the ability to learn how to derive knowledge from raw documents. This is achieved by training the model using a combination of training document dataset $D_{train}^{Doc}$ and associated training QA dataset $D_{train}^{QA}$, as depicted in the upper left part of Figure \ref{fig:framework}. To enhance effective knowledge absorption, we present $D_{train}^{Doc}$ along with a series of knowledge-intensive tasks (\aka self-teaching tasks) $D_{train}^{Self}$ that are related to their content using the proposed \selfteach{} strategy (in the lower part of Figure \ref{fig:framework}). These tasks are generated in a self-supervised manner based on the contents of $D_{train}^{Doc}$, using the proposed \selfteach{} learning approach (Section \ref{sec:self-exam}). The multi-task training objective is:
\begin{equation}
{L}_{\theta}^{Stage1}= {L}_{\theta}(D_{train}^{Doc}) + {L}_{\theta}(D_{train}^{Self}) + {L}_{\theta}(D_{train}^{QA})
\end{equation}
\paragraph{Stage 2: Learn New Knowledge while Reviewing QA Skills.} Our aim is to train the model $M$ to apply the learned strategy for spontaneously extracting new knowledge from unseen documents (\ie raw test corpora $D_{test}^{Doc}$). In addition to training on $D_{test}^{Doc}$, we include $D_{train}^{QA}$, allowing the model $M$ to review and refine its question-answering ability. The objective of this stage is:
\begin{equation} 
{L}_{\theta}^{Stage2}= {L}_{\theta}(D_{test}^{Doc}) + {L}_{\theta}(D_{train}^{QA})
\end{equation}
\paragraph{Stage 3: Continually Learn.} Our goal is to ensure that the model $M$ thoroughly absorbs the new knowledge by conducting follow-up training on $D_{test}^{Doc}$ (raw corpora). The objective is as follows:
\begin{equation}
{L}_{\theta}^{Stage3}= {L}_{\theta}(D_{test}^{Doc})
\end{equation}
\subsection{\selfteach{} Learning Strategy} 
\label{sec:self-exam} 
Motivated by the Feynman Technique, we aim to equip the model with systematic knowledge learning abilities from three perspectives: memorization, comprehension, and self-reflection, as shown in the lower part of Figure \ref{fig:framework}. Specifically, we devise a self-supervised \selfteach{} learning strategy that presents the raw documents $D_{train}^{Doc}$ as plain texts for memorization and as a series of knowledge-intensive tasks in a question-answering format related to their content for comprehension and self-reflection (Table \ref{tab:multi_task_list}). This method \textit{does not require any specific mining patterns, making it applicable to any raw texts}.
\vspace{-2mm}
\paragraph{Memorization.}
To allow the model $M$ to learn to memorize and capitalize on the factual information embedded in the raw texts, we execute the \textit{\textbf{next-token prediction}} task on plain document texts.
\vspace{-2mm}
\paragraph{Comprehension.}
Our goal is to facilitate the model's ability to comprehend the factual knowledge within the document in a top-down manner. To achieve this, we conduct the following tasks:

$(\RN{1})$ \textit{\textbf{Summarization}} allows the model to learn to grasp the topic by using the prompt \texttt{Write a title:} to encourage the model to summarize the raw text, with the document title serving as the ground truth.

$(\RN{2})$  \textit{\textbf{Gist identification}} improves the model's ability to pinpoint the key elements (\ie entities) within the atomic facts. Specifically, we prompt the model with \texttt{Highlight the key information within the article:}, and use the entities within the document as gold answers, identified using Spacy\footnote{\url{https://spacy.io/usage}}.

$(\RN{3})$ \textit{\textbf{Natural language inference}} provides the model with the capability to determine whether a statement can be inferred from specific document contents (\ie ``Yes,'' ``No,'' or ``It's impossible to say''), thus avoiding misconceptions that may arise during knowledge acquisition. Specifically, we use a randomly sampled sentence (identified using NLTK\footnote{A natural language toolkit. \url{https://www.nltk.org/}}) within the document content as the true statement, and a corrupted version where one entity is replaced by an irrelevant entity from another sentence as the false statement. Then, we prompt the model with \texttt{Based on the article above can we conclude that} and the sampled sentence (either initial or corrupted), with the three relations as options and corresponding answers.
\vspace{-2mm}
\paragraph{Self-Reflection.}
Our objective is to improve the model's ability to memorize and recall acquired knowledge by ``identifying and filling in the knowledge gaps.'' To this end, we devise the following closed-book generation tasks:

$(\RN{1})$ \textit{\textbf{``Teaching''}} fosters the model's ability to recall its acquired knowledge on a particular topic by "pretending to teach" others, using the prompt \texttt{Tell me about \{topic\}:} with the document content serving as the answer.

$(\RN{2})$ \textit{\textbf{``Flashcards''}} imparts the model with the ability to recall its learned information based on the topic and associated keywords, using the prompt \texttt{Generate a concrete description about \{topic\} based on the following keywords:}, with the document text as the answer.

$(\RN{3})$ \textit{\textbf{Fill-in-the-Blank}} equips the model with the ability to conduct a detailed check on the acquired factual information. Specifically, we randomly replace one entity with a ``--'' symbol to form a cloze question, with the replaced entity serving as the corresponding answer.

$(\RN{4})$ \textit{\textbf{Multi-choice QA}} helps the model learn to differentiate the correct answer from the available options and prevents confusion with irrelevant content. Specifically, we randomly replace one entity with a ``--'' symbol to form a cloze question, with the replaced entity and three other entities randomly sampled from the document forming the options, and the replaced entity serving as the correct choice.

$(\RN{5})$ \textit{\textbf{Sentence completion}} allows the model to develop its ability to focus on factual data found towards the end of a sentence. This is crucial since our initial observations indicate that the model frequently encounters difficulties when attempting to extract knowledge from later positions. Additionally, the model is anticipated to learn to emphasize not only entities but also phrase-level factual information. To achieve this, we first employ Spacy to pinpoint prepositions in a randomly chosen sentence from the document. Then, we store the phrase that follows the final preposition as the correct answer and the portion of the sentence preceding the phrase as the question. Comprehensive templates for each task can be found in Table \ref{tab:multi_task_list}.

\section{Experiments}
\begin{table*}[!t]
\setlength\tabcolsep{1pt}
  \centering
  \begin{threeparttable}
  \fontsize{9}{10}
  \selectfont
    \begin{tabular}{lccc}
    \toprule
    \multirow{2}{*}{\textbf{Method}}&
    \multicolumn{3}{c}{\textbf{Training Data in Each Stage}}
     \cr\cmidrule(lr){2-4} 
     &\textbf{Stage 1}&\textbf{Stage 2} & \textbf{Stage 3} \cr
    \midrule
    Continued Pre-Training& & & \circled{1} \colorbox{mypink1!90}{test doc} \\ 
    Standard Instruction-Tuning &\multicolumn{2}{c}{  \circled{1} \colorbox{mygrey1}{train doc}\&\colorbox{mypink1!90}{test doc} } & \circled{2} \colorbox{myblue1}{train QA}  \\
    PIT & \multicolumn{2}{c}{ \circled{1} \colorbox{myblue1}{train QA}\colorbox{mygrey1}{train doc} } &  \circled{2} \colorbox{mypink1!90}{test doc}  \\
    \selfpt{} & \circled{1}  \colorbox{myblue1}{train QA}\&\colorbox{mygrey1}{train doc w/ self-teaching tasks}  &  \circled{2} \colorbox{myblue1}{train QA}\&\colorbox{mypink1!90}{test doc} &  \circled{3}  \colorbox{mypink1!90}{test doc} \\
    \midrule
    \multicolumn{4}{l}{\textit{\textbf{Variants of \selfpt{}}}} \\
    \midrule
    \selfpt{} w/o Review&  \multicolumn{2}{c}{ \circled{1} \colorbox{myblue1}{train QA}\&\colorbox{mygrey1}{train doc w/ self-teaching tasks} } &  \circled{2} \colorbox{mypink1!90}{test doc} \\
    \selfpt{} via Read. &\multicolumn{2}{c}{ \circled{1} \colorbox{myblue1}{train QA}\&\colorbox{mygrey1}{train doc (reading-comprehension format~\cite{cheng2024adapting})} } &  \circled{2} \colorbox{mypink1!90}{test doc}  \\
    \selfpt{} w/ Pre-Review &  \circled{1} \colorbox{myblue1}{train QA}\&\colorbox{mygrey1}{train doc w/ self-teaching tasks}  &  \circled{2} \colorbox{myblue1}{train QA}\&\colorbox{mygrey1}{train doc}  &  \circled{3}  \colorbox{mypink1!90}{test doc} \\
    \bottomrule  
    \end{tabular}
  \end{threeparttable}
  \caption{Depiction of the training stages and datasets used in the compared methods. All approaches train on test documents for the same number of epochs. See Table~\ref{tab:method_traing_stages_full} for the complete version.}
  \label{tab:method_traing_stages}
\end{table*}

\begin{table*}[!t]
\setlength\tabcolsep{2.5pt}
  \centering
  \begin{threeparttable}
  \fontsize{9}{10}
  \selectfont
    \begin{tabular}{lcccccccccccc}
    \toprule
    \multirow{4}{*}{\textbf{Method}}&
    \multicolumn{7}{c}{\textbf{Wiki-Newpages-2023-QA (Acquisition)}} 
     &\multicolumn{2}{c}{\textbf{NQ (Reten.)}} &\textbf{CSQA (Reten.)}\cr\cmidrule(lr){2-8} \cmidrule(lr){9-10} \cmidrule(lr){11-11}
     & \textbf{Memorization}
      &\multicolumn{5}{c}{\textbf{Extraction}}&\textbf{Reason.} &\multicolumn{2}{c}{\textbf{Extraction}}&\textbf{Reasoning} \cr\cmidrule(lr){2-2} \cmidrule(lr){3-7} \cmidrule(lr){8-8} \cmidrule(lr){9-10} \cmidrule(lr){11-11}
     & $\mathtt{PPL}$ ($\downarrow$)&\%  $\mathtt{Acc.}$ &\%  $\mathtt{EM}$ &\%  $\mathtt{F1}$&\%  $\mathtt{Rec.}$&\%  $\mathtt{Rouge}$ &\% $\mathtt{Acc.}$ &\%  $\mathtt{EM}$ &\%  $\mathtt{F1}$ &\%  $\mathtt{Acc.}$\cr
    \midrule
    \multicolumn{11}{c}{\textbf{Knowledge Acquisition on Wiki-Newpages-2023-10-Bio (Single-Domain Scenario)}}\cr
    \midrule
    \multicolumn{8}{l}{\textcolor{lightgray!99}{\textit{w/o Knowledge Injection}}}\cr
    Open-book w/ test doc &8.41&55.20& 31.83&64.48&75.55&62.10& 7.96&-&-&-\\
    Closed-book &8.41&4.68&2.87&14.63&16.98&15.07 &7.96& 16.05&24.67&53.40\\
    \midrule
   \multicolumn{8}{l}{\textcolor{lightgray!99}{\textit{w/ Knowledge Injection}}}\cr 
    Cont. Pre-training & 7.28&6.33&3.62&15.96&18.72&16.11 &15.09 & 16.00&24.11&53.40\\
    
    Standard Ins.-tuning & 6.83& 6.94 & 5.13& 19.15& 19.05& 19.48&39.09& 15.72& 23.67 & 51.84  \\ 
     PIT & 2.08& 14.03 & 11.61& 27.15 & 28.86& 27.11 &11.93& 15.72& \textbf{26.31} & 57.58\\
    \selfpt{} & \textbf{1.11}& \textbf{37.25}& \textbf{31.52} & \textbf{50.83}& \textbf{52.62} & \textbf{50.61} &\textbf{44.31}& \textbf{16.45} & 25.67 & \textbf{66.01}\\

    \midrule
        \multicolumn{11}{c}{\textbf{Knowledge Acquisition on Wiki-Newpages-2023-10-Multi (Multi-Domain Scenario)}}\cr
    \midrule
    Open-book w/ test doc & 7.84&48.93&26.63&60.37&71.71&58.54&6.33&-&-&-\\
    Closed-book & 7.84&4.53&2.73&16.19&18.63&16.38&6.33&16.05&24.67 & 53.40\\
    \midrule
    Cont. Pre-training & 3.32&5.86&3.40&18.04&20.59&18.42&14.51&\textbf{17.02}&25.05&53.56\\
    
    Standard Ins.-tuning & 2.73& 8.66 & 5.73& 24.94& 25.64& 25.31 & 34.91& 15.60& 26.26 & 52.74 \\ 
    
     
     PIT & 1.96& 14.31 & 8.72& 30.26 & 33.97& 30.22 &10.69&15.55 & \textbf{27.02} & 55.12 \\ 
     
      \selfpt{} & \textbf{1.13} & \textbf{22.30}& \textbf{16.51} & \textbf{39.94}& \textbf{41.02} & \textbf{39.89}  & \textbf{50.65}& 16.34 & 25.85 & \textbf{69.29}\\ 
      

        \midrule
    \multicolumn{11}{c}{\textbf{Knowledge Acquisition on Wiki-Newpages-2023-(9)10-Film (Cross-Domain Scenario)}}\cr
    \midrule
    Open-book w/ film doc &8.30&57.38 & 34.45&68.64&78.92&66.31&7.35&-&-&-\\
    Closed-book &8.30&3.35&1.88&11.27&12.97&11.49&7.35&16.05&24.67&53.40\\
    \midrule
    Cont. Pre-training &5.52 &3.46&2.30&11.83&14.30&11.98&12.04&\textbf{16.79}& 25.35 & 56.02\\
    Standard Ins.-tuning &2.83& 5.23 & 3.77& 16.15& 17.45& 16.45& \textbf{51.69}& 14.41& 25.54 & 49.80 \\ 

     
     PIT&1.52& 6.39 & 4.50& 16.97 & 18.92& 17.10 &3.03&13.06 & 23.42 & 54.38 \\

     \selfpt{} & \textbf{1.10} & \textbf{22.51}& \textbf{16.44} & \textbf{35.58}& \textbf{36.60} & \textbf{35.43} & 44.92& 16.77& \textbf{26.44} & \textbf{66.34}\\ 
    \bottomrule  
    \end{tabular}
  \end{threeparttable}
  \caption{Five-shot evaluation results on \llamaa{} for knowledge acquisition and retention are presented across single-domain, multi-domain, and cross-domain scenarios. Following \cite{jiang2024instructiontuned}, we also report results for: $(\RN{1})$ closed-book, where base LLMs are prompted with open-ended questions related to new knowledge in the test documents, and $(\RN{2})$ open-book w/ test doc, where base LLMs are prompted with questions along with relevant gold knowledge snippets from the test documents. For the complete results, refer to Table~\ref{tab:results_addition} (Appendix \ref{sec:results_addition}).}
  \label{tab:hard-bio-domain}
\end{table*}

\subsection{Setup}

\vspace{-1mm}
\noindent \textbf{Datasets and Evaluation Metrics.}
We validate \selfpt{} in both knowledge acquisition and retention for a well-rounded analysis.

We perform assessments on three \textbf{knowledge acquisition} tasks: $(\RN{1})$ \textit{Memorization}: We use test document datasets and report perplexity (PPL)~\cite{Jelinek1977PerplexityaMO}. $(\RN{2})$ \textit{Extraction}: We use test QA datasets for open-ended generation tasks and evaluate factual accuracy using exact match (EM), Recall, F1~\cite{kwiatkowski-etal-2019-natural}, Rouge-L~\cite{lin-2004-rouge}, and accuracy. $(\RN{3})$ \textit{Reasoning}: We use test QA datasets for NLI tasks and report accuracy.

We evaluate two aspects of \textbf{knowledge retention}: $(\RN{1})$ \textit{Extraction}: We assess the model's performance in retaining general factual knowledge using Natural Questions (NQ)~\cite{kwiatkowski-etal-2019-natural}, with EM and F1. $(\RN{2})$ \textit{Reasoning}: We evaluate the capability in retaining commonsense knowledge using CommonsenseQA (CSQA)~\cite{talmor-etal-2019-commonsenseqa} and report accuracy. All evaluations are conducted in a closed-book setting (see Appendix \ref{sec:setup_details}). 

\noindent \textbf{Compared Methods.} We compare \selfpt{} with three representative approaches (Table \ref{tab:method_traing_stages}): (1) Continued Pre-training \cite{ovadia2024finetuning}, (2) Standard Instruction-tuning \cite{saito2024answer}, and (3) PIT \cite{jiang2024instructiontuned}, which trains on $D_{train}^{QA}$ and $D_{train}^{Doc}$ with QA pairs positioned before their corresponding document texts. We also evaluate their variants (Table \ref{tab:method_traing_stages_full}). Results, averaged over three runs, show significant differences in means ($p$ < 0.001), with details in Appendix \ref{sec:implement_details}.


\vspace{-2.5mm}

\subsection{Main Results}

Table \ref{tab:hard-bio-domain} (top) presents the evaluation results on \llamaa{} in relation to knowledge acquisition and retention in the single-domain scenario using the Wiki-Bio dataset. 


\noindent\textbf{The curated dataset exhibits minimal overlap with the pre-training data of the LLMs.} The extremely low performance in the closed-book setting (\eg with EM around 2\% for knowledge extraction) indicates that the dataset has little in common with the pre-training data, thus ensuring the reliability of the evaluation results. The non-zero EM values might be due to a small number of collected Wikipedia articles that were initially published but underwent revisions after the cut-off time.

\noindent\textbf{\selfpt{} substantially improves the LLM's knowledge acquisition ability.} \selfpt{} greatly enhances the performance of \llamaa{} across three dimensions: $(\RN{1})$ reducing PPL to nearly 1, signifying effective memorization of the new documents; $(\RN{2})$ increasing EM by roughly 11.5\% on the knowledge extraction task, attaining performance comparable to the open-book setting; $(\RN{3})$ achieving high accuracy among the compared methods for the reasoning task, demonstrating excellent understanding of the newly acquired knowledge. These results underscore the importance of first training the model to develop knowledge absorption capabilities before exposing it to test documents, aligning with findings from~\citet{jiang2024instructiontuned}. More importantly, \selfpt{} significantly outperforms \pit{}, highlighting the role of comprehension and self-reflection beyond mere memorization, further validating the effectiveness of \selfteach{}.



\noindent\textbf{\selfpt{} excels in knowledge retention.} Unlike other methods that display fluctuating performance, \selfpt{} shows a strong ability to maintain previously acquired knowledge in terms of both knowledge extraction and reasoning. The slight improvements in evaluation metrics, such as F1 (roughly 1\% on extracting learned world knowledge) and accuracy (around 13\% on commonsense reasoning), compared to the closed-book performance without knowledge injection, suggest that systematically learning new knowledge doesn't necessarily lead to catastrophic forgetting.

We further validate the efficacy of \selfpt{} through additional analyses presented in the following appendices:  
Appendix \ref{sec:results_addition} (evaluation of three additional methods to validate the effectiveness of our three-stage design),  
\ref{sec:evaluate_compare_testqa_based} (comparison of training on test documents with accompanying QA),  
\ref{sec:fine_grained_compare} (fine-grained evaluation of the open-ended generation task),  
\ref{sec:qua_ana_case} (qualitative comparison of responses from \pit{} and \selfpt{}),  
\ref{sec:ablation} (ablation study investigating the impact of removing comprehension and self-reflection tasks from self-teaching on knowledge memorization and acquisition),  
\ref{sec:error_analysis} (error analysis), and  
\ref{sec:training_efficiency} (assessment of training efficiency).





\subsection{Results in the Multi-Domain and Cross-Domain Scenarios}
To explore the potential of \selfpt{} for enhancing LLM's knowledge acquisition and retention in real-world scenarios, we evaluate its performance in two challenging settings (Table \ref{tab:hard-bio-domain}): $(\RN{1})$ the multi-domain scenario (in the middle part); $(\RN{2})$ the cross-domain scenario (in the bottom part), where the training data is from Wiki-Bio, while the test data is from Wiki-Film.


\noindent \textbf{\selfpt{} shows strong potential in enhancing knowledge acquisition and retention across documents containing diverse new knowledge.} In Table \ref{tab:hard-bio-domain}, \selfpt{} consistently achieves the best performance in both settings.




\noindent\textbf{The capacity to systematically absorb knowledge improves generalization ability.} The substantial improvements over all compared methods in the cross-domain setting, \eg exceeding EM by 13\% on the knowledge extraction task, highlight the value of equipping the model with the ability to effectively absorb knowledge from raw documents using the \selfteach{} strategy, rather than solely teaching it how to answer questions.

\subsection{Training Dynamics}
\begin{figure}[!t]
\centering
\includegraphics[width=0.95\linewidth]{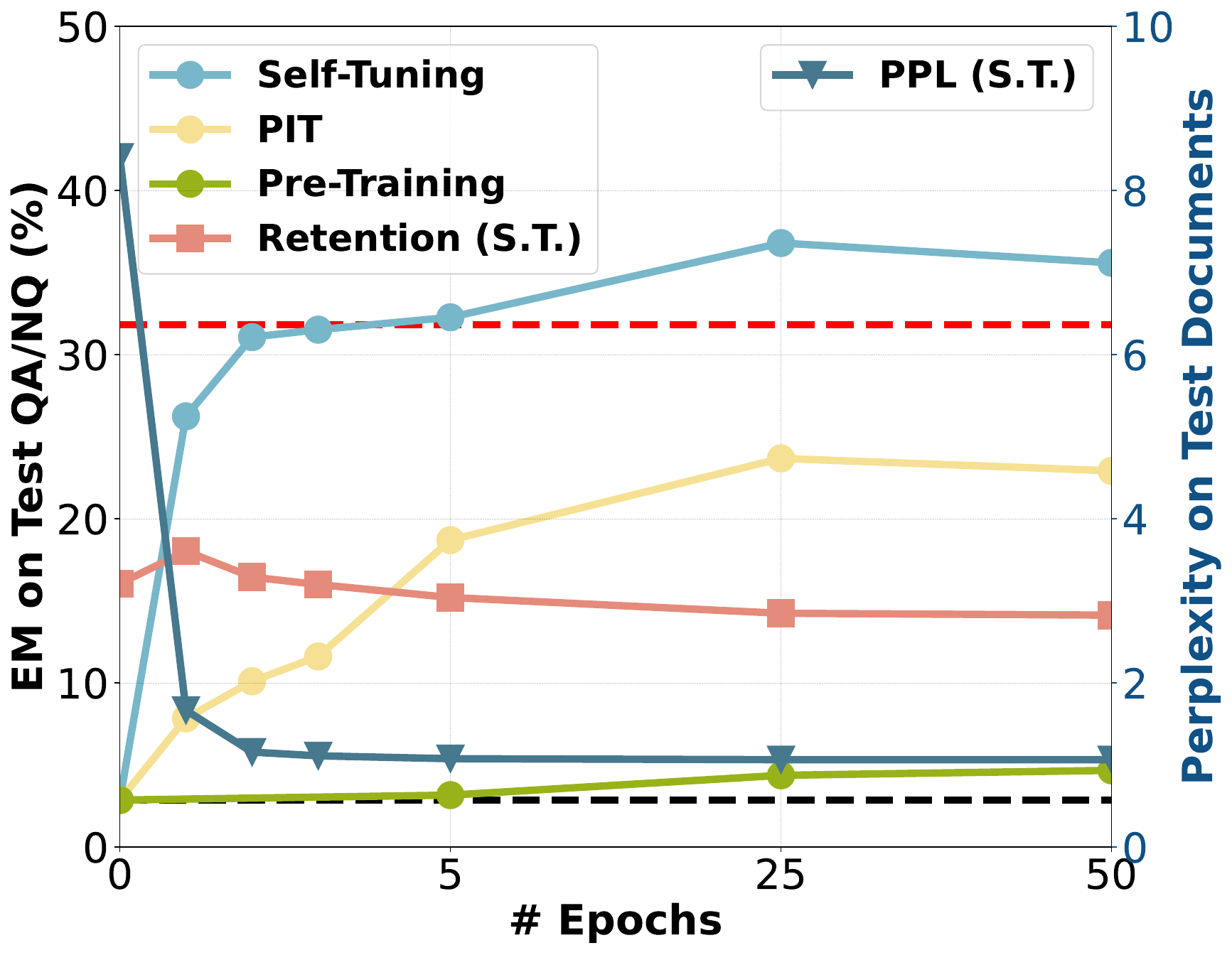}
\caption{Training dynamics on \llamaa{} \wrt knowledge memorization, extraction, and retention across different numbers of training epochs. We present the EM scores on NQ datasets to evaluate knowledge retention. The black and red dashed lines represent the baseline closed-book and open-book performances for the knowledge extraction task, respectively.}
\label{fig:train_dynamics_mr_epoch50}
\end{figure}

We analyze the training dynamics of \selfpt{} during continued pre-training (beginning from Stage 2 in Figure \ref{fig:framework}) on the test documents by varying the number of training epochs for two main reasons: $(\RN{1})$ to eliminate the possibility that the exceptional performance of \selfpt{} in enhancing knowledge acquisition is merely a result of early fitting on the test documents, and $(\RN{2})$ to conduct an in-depth assessment of its long-term knowledge retention capability. Furthermore, we integrate the results of PIT and continued pre-training to offer a well-rounded evaluation.

\noindent\textbf{The remarkable performance of \selfpt{} in enhancing knowledge acquisition does not stem from early-fitting.} In Figure \ref{fig:train_dynamics_mr_epoch50}, we observe that \selfpt{} not only memorizes new knowledge more rapidly than the compared methods, lowering PPL to almost 1 within 3 epochs, but also consistently achieves the best performance during long-term training. Remarkably, \selfpt{} begins to outperform the open-book performance from the 5th epoch and reaches its peak at the 25th epoch with a 5\% higher EM score on the knowledge extraction task.


\selfpt{} performs well in preserving previously acquired knowledge, with only a small decline in EM of roughly 2-3\% over the course of 50 training epochs. This suggests that \selfpt{} has great potential for real-world applications.

\subsection{Variants of \selfpt{}}

\begin{table}[!t]
\setlength\tabcolsep{1pt}
  \centering
  \begin{threeparttable}
  \fontsize{8.7}{9}
  \selectfont
    \begin{tabular}{lcccccc}
    \toprule
    \multirow{3}{*}{\textbf{Method}} &
    \multicolumn{5}{c}{\textbf{Wiki-Bio (Acquisition)}} \cr\cmidrule(lr){2-6}
     & \textbf{Mem.} &\multicolumn{3}{c}{\textbf{Extraction}} & \textbf{Reason.} \cr\cmidrule(lr){2-2} \cmidrule(lr){3-5} \cmidrule(lr){6-6}
     & $\mathtt{PPL}$ ($\downarrow$) & \% $\mathtt{Acc.}$ & \% $\mathtt{EM}$ & \% $\mathtt{Rouge}$ & \% $\mathtt{Acc.}$\cr
    \midrule
    Cont. Pre-training & 7.28 & 4.68 & 2.87 & 15.07 & 7.96 \\
    \hdashline
    \noalign{\vskip 0.07cm} 
     S.T. w/o Review & 1.26 & 28.36 & 23.68 & 41.11 & \textbf{50.40} \\
     S.T. via Read. & 1.46 & 20.97 & 17.65 & 34.55 & 39.37 \\
     S.T. w/ Pre-Review & 1.28 & 29.86 & 25.94 & 43.31 & 46.91 \\
    \selfpt{} & \textbf{1.11} & \textbf{37.25} & \textbf{31.52} & \textbf{50.61} & 44.31 \\
    \bottomrule  
    \end{tabular}
  \end{threeparttable}
  \caption{Results of the \selfpt{} variants on \llamaa{} on Wiki-Bio (Appendix~\ref{sec:selfpt_variants}).}
  \label{tab:variants-bio-domain}
\end{table}

\noindent \textbf{Setup.} We investigate three variants of \selfpt{} (in the lower part of Table \ref{tab:method_traing_stages}): $(\RN{1})$ \textbf{\selfpt{} w/o Review}, where we continue training on test documents without the reviewing capability; $(\RN{2})$ \textbf{\selfpt{} via Read.}, which displays the training documents in a reading-comprehension format \cite{cheng2024adapting} (see Table \ref{tab:train_doc_read_compre_format}); $(\RN{3})$ \textbf{\selfpt{} w/ Pre-Review}, which trains on a mix of training documents and training QA in the second stage, before training on test documents.


\noindent \textbf{Results.} In Table \ref{tab:variants-bio-domain}, despite having lower performance than \selfpt{}, all variants significantly enhance the model's ability for knowledge acquisition compared to continued pre-training.

\noindent \textbf{Reviewing QA ability aids in knowledge acquisition.} Compared to \selfpt{}, \selfpt{} w/o Review exhibits inferior performance. Moreover, we suspect that the lower performance of \selfpt{} w/ Pre-Review is because reviewing QA ability \textit{during}, rather than before, the continuous learning of new knowledge is more effective in reducing distribution shift, thereby stabilizing the training process.



\noindent\textbf{Decoupling the knowledge acquisition process into three perspectives is more effective than solely focusing on comprehension.} The comparison between \selfpt{} w/o Review and \selfpt{} w/ Read. demonstrates that presenting the test document text from three distinct perspectives contributes more to knowledge memorization (1.26\% vs. 1.46\% on PPL), extraction (23.68\% vs. 17.65\% on EM), and reasoning (50.40\% vs. 39.37\% on accuracy) than presenting the test document text with all constructed tasks as a whole.

\subsection{Results of Varying Models and Corpora}
\begin{table}[!t]
\setlength\tabcolsep{1pt}
  \centering
  \begin{threeparttable}
  \fontsize{8.6}{9}
  \selectfont
    \begin{tabular}{lcccccccccccc}
    \toprule
    \multirow{2}{*}{\textbf{Method}} &
    \multicolumn{6}{c}{\textbf{Acquisition}}  \cr\cmidrule(lr){2-7} 
     & $\mathtt{PPL}$ ($\downarrow$) & \% $\mathtt{Acc.}$ & \% $\mathtt{EM}$ & \% $\mathtt{Recall}$ & \% $\mathtt{Rouge}$ \cr
    \midrule
    \multicolumn{7}{c}{\textbf{Varying Model} (Qwen2-7B on WikiBio-2024)}\cr
    \midrule
    Closed-book & 12.41 & 4.16 & 2.55 & 15.01  & 13.17 \\
    Stand. Ins.-tuning & 2.77 & 11.29 & 9.36 & 25.45 & 24.83 \\
    PIT & 1.97 & 11.41 & 9.53 & 25.98 & 25.64 \\
    \selfpt{} & \textbf{1.14} & \textbf{31.79} & \textbf{28.51} & \textbf{44.91}  & \textbf{43.33} \\
    \midrule
    \multicolumn{7}{c}{\textbf{Varying Model} (Mistral-7B-v0.1 on WikiBio-2023)}\cr
    \midrule
    Closed-book & 8.45 & 6.64 & 4.37 &  19.51 & 17.25 \\
    Stand. Ins.-tuning & 2.84 & 16.44 & 13.88 &  29.54 & 29.13 \\
    PIT & 1.42 & 26.85 & 23.08 &  40.36 & 39.52 \\
    \selfpt{} & \textbf{1.08} & \textbf{41.63} & \textbf{36.50} & \textbf{55.32} & \textbf{52.87} \\
    \midrule
    \midrule
    \multicolumn{7}{c}{\textbf{Varying Corpora} (\llamaa{} on WebNews-2023)}\cr
    \midrule
    Closed-book & 11.20 & 9.04 & 6.30 & 24.22 & 17.99 \\
    Stand. Ins.-tuning & 3.27 & 21.48 & 13.38 & 37.66 & 31.31 \\
    PIT & 1.67 & 30.37 & 18.96 & 51.17 & 40.53 \\
    \selfpt{} & \textbf{1.10} & \textbf{37.48} & \textbf{28.74} & \textbf{56.26} & \textbf{48.21} \\

    \bottomrule  
    \end{tabular}
  \end{threeparttable}
  \caption{Results of Varying Models and Corpora.}
  \label{tab:eval_new_model}
  \vspace{-6mm}
\end{table}




\noindent \textbf{Setup.}  
We evaluate \selfpt{} using diverse models, including Qwen2-7B~\cite{yang2024qwen2technicalreport}, Mistral-7B-v0.1~\cite{jiang2023mistral7b}, and Gemma-7B~\cite{team2024gemma} (see Appendix~\ref{sec:evaluate_diverse_models}), as well as different corpora, such as WebNews-2023~\cite{tang2024multihopragbenchmarkingretrievalaugmentedgeneration}, which consists of worldwide news articles from diverse websites (see Appendix~\ref{sec:evaluate_varied} for further details).

\noindent \textbf{Results.}  
The results in Table~\ref{tab:eval_new_model} demonstrate that \selfpt{} consistently achieves the best performance, highlighting its strong generalizability across both models and corpora. Further evaluation results for \llamaathirteen{} and \llamaachat{} are available in Appendices~\ref{sec:evaluate_13b} and~\ref{sec:evaluate_chat}, respectively.

\label{sec:experiments}
\section{Conclusion}
In this study, we introduce \selfpt{} to enhance an LLM's ability to effectively learn from raw documents through self-teaching. Specifically, we develop \selfteach{}, a self-supervised learning strategy that presents documents as plain texts along with various knowledge-intensive tasks derived directly from the documents. Additionally, we present three Wikipedia-Newpages-2023-QA datasets to enable a comprehensive evaluation of an LLM's knowledge acquisition capabilities across three distinct scenarios. Our findings show that \selfpt{} consistently yields superior performance on the knowledge acquisition tasks while showing impressive knowledge retention performance. These results suggest the potential for broader applications of \selfpt{}.


\section*{Limitations}
While our experimental results show promise, we consider these findings to be preliminary, as there are still many unexplored aspects in this field.

\paragraph{Combining with Continual Learning Approaches.}

Our study primarily focuses on enhancing a language model's ability to effectively learn new knowledge from previously unseen raw corpora. Although experimental results on MCQA and NQ demonstrate that our \selfpt{} method well preserves previously acquired knowledge, future research could explore integrating \selfpt{} with continual learning approaches~\cite{wang2024comprehensivesurveycontinuallearning}. For instance, regularization-based methods such as EWC~\cite{Kirkpatrick_2017} and replay-based methods, like incorporating segments from general domain datasets (\eg Wiki data~\cite{zhang-etal-2024-self}), could improve the model's capacity to retain learned knowledge and skills while mitigating the risk of overfitting to new information. 

In this study, we intentionally avoided using continual learning approaches to ensure a fair comparison of knowledge injection with previous methods. However, we present preliminary results of combining \selfpt{} with a replay-based approach in Appendix~\ref{sec:evaluate_continual_learning}. These results confirm the strong potential of integrating \selfpt{} with continual learning techniques to improve both knowledge acquisition and retention.


\paragraph{Performing More Comprehensive Evaluations of LLMs' Knowledge Acquisition Capabilities.} 
In this study, we evaluate the knowledge acquisition capabilities of LLMs from three important perspectives: knowledge memorization, extraction, and reasoning. Future work could consider additional evaluation aspects, such as integrating factual knowledge with mathematical reasoning, to explore the model's ability to utilize the learned factual knowledge in solving more complex real-world problems \cite{zheng2024take}.

\paragraph{Regarding Resource Demands.} 

To verify the efficacy of \textsc{Self-Tuning}, we provide a detailed analysis of training efficiency on the Wiki-Newpages-2023-Bio dataset, conducted using 8 Tesla V100 GPUs (32GB) with \textsc{Llama-7B}, in Appendix~\ref{sec:training_efficiency}. This analysis demonstrates that our \textsc{Self-Tuning} framework not only significantly outperforms the strongest baseline method but also achieves this with reduced training time. Furthermore, the effectiveness of \textsc{Self-Tuning} across three distinct scenarios highlights its ability to directly assimilate new knowledge from incoming test documents without requiring retraining on the original training corpus (\ie omitting the first stage). Notably, \textsc{Self-Tuning} eliminates the need for any additional annotation costs. All experiments were conducted on 8 Tesla V100 GPUs (32GB), with training completing in just a few hours. Consequently, we anticipate minimal barriers to the adoption of \textsc{Self-Tuning}, even for teams with limited computational resources.




\newpage
\bibliography{custom}

\begin{thebibliography}{60}
\providecommand{\natexlab}[1]{#1}

\bibitem[{Allen-Zhu and Li(2023)}]{allenzhu2023physics}
Zeyuan Allen-Zhu and Yuanzhi Li. 2023.
\newblock \href {https://arxiv.org/abs/2309.14316} {Physics of language models: Part 3.1, knowledge storage and extraction}.
\newblock \emph{Preprint}, arXiv:2309.14316.

\bibitem[{Ambion et~al.(2020)Ambion, De~Leon, Mendoza, and Navarro}]{ambion2020utilization}
Ronnel Ian~A Ambion, Rainier Santi~C De~Leon, Alfonso Pio Angelo~R Mendoza, and Reinier~M Navarro. 2020.
\newblock The utilization of the feynman technique in paired team teaching towards enhancing grade 10 anhs students’ academic achievement in science.
\newblock In \emph{2020 IEEE Integrated STEM Education Conference (ISEC)}, pages 1--3. IEEE.

\bibitem[{Cheng et~al.(2024)Cheng, Huang, and Wei}]{cheng2024adapting}
Daixuan Cheng, Shaohan Huang, and Furu Wei. 2024.
\newblock \href {https://openreview.net/forum?id=y886UXPEZ0} {Adapting large language models via reading comprehension}.
\newblock In \emph{The Twelfth International Conference on Learning Representations}.

\bibitem[{Cohen et~al.(2023)Cohen, Geva, Berant, and Globerson}]{cohen-etal-2023-crawling}
Roi Cohen, Mor Geva, Jonathan Berant, and Amir Globerson. 2023.
\newblock \href {https://doi.org/10.18653/v1/2023.findings-eacl.139} {Crawling the internal knowledge-base of language models}.
\newblock In \emph{Findings of the Association for Computational Linguistics: EACL 2023}, pages 1856--1869, Dubrovnik, Croatia. Association for Computational Linguistics.

\bibitem[{ContextualAI(2024)}]{contextualai2024rag}
ContextualAI. 2024.
\newblock \href {https://contextual.ai/introducing-rag2/} {Introducing rag 2.0}.

\bibitem[{Gekhman et~al.(2024)Gekhman, Yona, Aharoni, Eyal, Feder, Reichart, and Herzig}]{gekhman2024does}
Zorik Gekhman, Gal Yona, Roee Aharoni, Matan Eyal, Amir Feder, Roi Reichart, and Jonathan Herzig. 2024.
\newblock \href {https://arxiv.org/abs/2405.05904} {Does fine-tuning llms on new knowledge encourage hallucinations?}
\newblock \emph{Preprint}, arXiv:2405.05904.

\bibitem[{He et~al.(2021)He, Liu, Gao, and Chen}]{he2021deberta}
Pengcheng He, Xiaodong Liu, Jianfeng Gao, and Weizhu Chen. 2021.
\newblock \href {https://openreview.net/forum?id=XPZIaotutsD} {Deberta: Decoding-enhanced bert with disentangled attention}.
\newblock In \emph{International Conference on Learning Representations}.

\bibitem[{Hendrycks et~al.(2021)Hendrycks, Burns, Basart, Zou, Mazeika, Song, and Steinhardt}]{hendrycks2021measuringmassivemultitasklanguage}
Dan Hendrycks, Collin Burns, Steven Basart, Andy Zou, Mantas Mazeika, Dawn Song, and Jacob Steinhardt. 2021.
\newblock \href {https://arxiv.org/abs/2009.03300} {Measuring massive multitask language understanding}.
\newblock \emph{Preprint}, arXiv:2009.03300.

\bibitem[{Huang et~al.(2023)Huang, Yu, Ma, Zhong, Feng, Wang, Chen, Peng, Feng, Qin, and Liu}]{huang2023survey}
Lei Huang, Weijiang Yu, Weitao Ma, Weihong Zhong, Zhangyin Feng, Haotian Wang, Qianglong Chen, Weihua Peng, Xiaocheng Feng, Bing Qin, and Ting Liu. 2023.
\newblock \href {https://arxiv.org/abs/2311.05232} {A survey on hallucination in large language models: Principles, taxonomy, challenges, and open questions}.
\newblock \emph{Preprint}, arXiv:2311.05232.

\bibitem[{Ibrahim et~al.(2024)Ibrahim, Thérien, Gupta, Richter, Anthony, Lesort, Belilovsky, and Rish}]{ibrahim2024simple}
Adam Ibrahim, Benjamin Thérien, Kshitij Gupta, Mats~L. Richter, Quentin Anthony, Timothée Lesort, Eugene Belilovsky, and Irina Rish. 2024.
\newblock \href {https://arxiv.org/abs/2403.08763} {Simple and scalable strategies to continually pre-train large language models}.
\newblock \emph{Preprint}, arXiv:2403.08763.

\bibitem[{Jang et~al.(2022)Jang, Ye, Yang, Shin, Han, KIM, Choi, and Seo}]{jang2022towards}
Joel Jang, Seonghyeon Ye, Sohee Yang, Joongbo Shin, Janghoon Han, Gyeonghun KIM, Stanley~Jungkyu Choi, and Minjoon Seo. 2022.
\newblock \href {https://openreview.net/forum?id=vfsRB5MImo9} {Towards continual knowledge learning of language models}.
\newblock In \emph{International Conference on Learning Representations}.

\bibitem[{Jelinek et~al.(1977)Jelinek, Mercer, Bahl, and Baker}]{Jelinek1977PerplexityaMO}
Frederick Jelinek, Robert~L. Mercer, Lalit~R. Bahl, and Janet~M. Baker. 1977.
\newblock \href {https://api.semanticscholar.org/CorpusID:121680873} {Perplexity—a measure of the difficulty of speech recognition tasks}.
\newblock \emph{Journal of the Acoustical Society of America}, 62.

\bibitem[{Jeong et~al.(2024)Jeong, Baek, Cho, Hwang, and Park}]{jeong2024adaptiverag}
Soyeong Jeong, Jinheon Baek, Sukmin Cho, Sung~Ju Hwang, and Jong~C. Park. 2024.
\newblock \href {https://arxiv.org/abs/2403.14403} {Adaptive-rag: Learning to adapt retrieval-augmented large language models through question complexity}.
\newblock \emph{Preprint}, arXiv:2403.14403.

\bibitem[{Jiang et~al.(2023)Jiang, Sablayrolles, Mensch, Bamford, Chaplot, de~las Casas, Bressand, Lengyel, Lample, Saulnier, Lavaud, Lachaux, Stock, Scao, Lavril, Wang, Lacroix, and Sayed}]{jiang2023mistral7b}
Albert~Q. Jiang, Alexandre Sablayrolles, Arthur Mensch, Chris Bamford, Devendra~Singh Chaplot, Diego de~las Casas, Florian Bressand, Gianna Lengyel, Guillaume Lample, Lucile Saulnier, Lélio~Renard Lavaud, Marie-Anne Lachaux, Pierre Stock, Teven~Le Scao, Thibaut Lavril, Thomas Wang, Timothée Lacroix, and William~El Sayed. 2023.
\newblock \href {https://arxiv.org/abs/2310.06825} {Mistral 7b}.
\newblock \emph{Preprint}, arXiv:2310.06825.

\bibitem[{Jiang et~al.(2024{\natexlab{a}})Jiang, Li, Zhao, Song, Zhang, and Wen}]{jiang2024mix}
Jinhao Jiang, Junyi Li, Wayne~Xin Zhao, Yang Song, Tao Zhang, and Ji-Rong Wen. 2024{\natexlab{a}}.
\newblock Mix-cpt: A domain adaptation framework via decoupling knowledge learning and format alignment.
\newblock \emph{arXiv preprint arXiv:2407.10804}.

\bibitem[{Jiang et~al.(2024{\natexlab{b}})Jiang, Wang, Wu, Zhong, Zeng, Gao, Li, Jiang, Shang, Tang, Liu, and Wang}]{jiang2024learning}
Yuxin Jiang, Yufei Wang, Chuhan Wu, Wanjun Zhong, Xingshan Zeng, Jiahui Gao, Liangyou Li, Xin Jiang, Lifeng Shang, Ruiming Tang, Qun Liu, and Wei Wang. 2024{\natexlab{b}}.
\newblock \href {https://arxiv.org/abs/2402.11905} {Learning to edit: Aligning llms with knowledge editing}.
\newblock \emph{Preprint}, arXiv:2402.11905.

\bibitem[{Jiang et~al.(2024{\natexlab{c}})Jiang, Sun, Shi, Rodriguez, Zhou, Neubig, Lin, tau Yih, and Iyer}]{jiang2024instructiontuned}
Zhengbao Jiang, Zhiqing Sun, Weijia Shi, Pedro Rodriguez, Chunting Zhou, Graham Neubig, Xi~Victoria Lin, Wen tau Yih, and Srinivasan Iyer. 2024{\natexlab{c}}.
\newblock \href {https://arxiv.org/abs/2402.12847} {Instruction-tuned language models are better knowledge learners}.
\newblock \emph{Preprint}, arXiv:2402.12847.

\bibitem[{Kirkpatrick et~al.(2017)Kirkpatrick, Pascanu, Rabinowitz, Veness, Desjardins, Rusu, Milan, Quan, Ramalho, Grabska-Barwinska, Hassabis, Clopath, Kumaran, and Hadsell}]{Kirkpatrick_2017}
James Kirkpatrick, Razvan Pascanu, Neil Rabinowitz, Joel Veness, Guillaume Desjardins, Andrei~A. Rusu, Kieran Milan, John Quan, Tiago Ramalho, Agnieszka Grabska-Barwinska, Demis Hassabis, Claudia Clopath, Dharshan Kumaran, and Raia Hadsell. 2017.
\newblock \href {https://doi.org/10.1073/pnas.1611835114} {Overcoming catastrophic forgetting in neural networks}.
\newblock \emph{Proceedings of the National Academy of Sciences}, 114(13):3521–3526.

\bibitem[{Kwiatkowski et~al.(2019)Kwiatkowski, Palomaki, Redfield, Collins, Parikh, Alberti, Epstein, Polosukhin, Devlin, Lee, Toutanova, Jones, Kelcey, Chang, Dai, Uszkoreit, Le, and Petrov}]{kwiatkowski-etal-2019-natural}
Tom Kwiatkowski, Jennimaria Palomaki, Olivia Redfield, Michael Collins, Ankur Parikh, Chris Alberti, Danielle Epstein, Illia Polosukhin, Jacob Devlin, Kenton Lee, Kristina Toutanova, Llion Jones, Matthew Kelcey, Ming-Wei Chang, Andrew~M. Dai, Jakob Uszkoreit, Quoc Le, and Slav Petrov. 2019.
\newblock \href {https://doi.org/10.1162/tacl_a_00276} {Natural questions: A benchmark for question answering research}.
\newblock \emph{Transactions of the Association for Computational Linguistics}, 7:452--466.

\bibitem[{Lewis et~al.(2021)Lewis, Perez, Piktus, Petroni, Karpukhin, Goyal, Küttler, Lewis, tau Yih, Rocktäschel, Riedel, and Kiela}]{lewis2021retrievalaugmented}
Patrick Lewis, Ethan Perez, Aleksandra Piktus, Fabio Petroni, Vladimir Karpukhin, Naman Goyal, Heinrich Küttler, Mike Lewis, Wen tau Yih, Tim Rocktäschel, Sebastian Riedel, and Douwe Kiela. 2021.
\newblock \href {https://arxiv.org/abs/2005.11401} {Retrieval-augmented generation for knowledge-intensive nlp tasks}.
\newblock \emph{Preprint}, arXiv:2005.11401.

\bibitem[{Lin(2004)}]{lin-2004-rouge}
Chin-Yew Lin. 2004.
\newblock \href {https://aclanthology.org/W04-1013} {{ROUGE}: A package for automatic evaluation of summaries}.
\newblock In \emph{Text Summarization Branches Out}, pages 74--81, Barcelona, Spain. Association for Computational Linguistics.

\bibitem[{Liu et~al.(2023)Liu, Yao, Ton, Zhang, Guo, Cheng, Klochkov, Taufiq, and Li}]{liu2023trustworthy}
Yang Liu, Yuanshun Yao, Jean-Francois Ton, Xiaoying Zhang, Ruocheng Guo, Hao Cheng, Yegor Klochkov, Muhammad~Faaiz Taufiq, and Hang Li. 2023.
\newblock \href {https://arxiv.org/abs/2308.05374} {Trustworthy llms: a survey and guideline for evaluating large language models' alignment}.
\newblock \emph{Preprint}, arXiv:2308.05374.

\bibitem[{Mecklenburg et~al.(2024)Mecklenburg, Lin, Li, Holstein, Nunes, Malvar, Silva, Chandra, Aski, Yannam, Aktas, and Hendry}]{mecklenburg2024injecting}
Nick Mecklenburg, Yiyou Lin, Xiaoxiao Li, Daniel Holstein, Leonardo Nunes, Sara Malvar, Bruno Silva, Ranveer Chandra, Vijay Aski, Pavan Kumar~Reddy Yannam, Tolga Aktas, and Todd Hendry. 2024.
\newblock \href {https://arxiv.org/abs/2404.00213} {Injecting new knowledge into large language models via supervised fine-tuning}.
\newblock \emph{Preprint}, arXiv:2404.00213.

\bibitem[{Min et~al.(2021)Min, Boyd-Graber, Alberti, Chen, Choi, Collins, Guu, Hajishirzi, Lee, Palomaki, Raffel, Roberts, Kwiatkowski, Lewis, Wu, Küttler, Liu, Minervini, Stenetorp, Riedel, Yang, Seo, Izacard, Petroni, Hosseini, Cao, Grave, Yamada, Shimaoka, Suzuki, Miyawaki, Sato, Takahashi, Suzuki, Fajcik, Docekal, Ondrej, Smrz, Cheng, Shen, Liu, He, Chen, Gao, Oguz, Chen, Karpukhin, Peshterliev, Okhonko, Schlichtkrull, Gupta, Mehdad, and tau Yih}]{min2021neurips}
Sewon Min, Jordan Boyd-Graber, Chris Alberti, Danqi Chen, Eunsol Choi, Michael Collins, Kelvin Guu, Hannaneh Hajishirzi, Kenton Lee, Jennimaria Palomaki, Colin Raffel, Adam Roberts, Tom Kwiatkowski, Patrick Lewis, Yuxiang Wu, Heinrich Küttler, Linqing Liu, Pasquale Minervini, Pontus Stenetorp, Sebastian Riedel, Sohee Yang, Minjoon Seo, Gautier Izacard, Fabio Petroni, Lucas Hosseini, Nicola~De Cao, Edouard Grave, Ikuya Yamada, Sonse Shimaoka, Masatoshi Suzuki, Shumpei Miyawaki, Shun Sato, Ryo Takahashi, Jun Suzuki, Martin Fajcik, Martin Docekal, Karel Ondrej, Pavel Smrz, Hao Cheng, Yelong Shen, Xiaodong Liu, Pengcheng He, Weizhu Chen, Jianfeng Gao, Barlas Oguz, Xilun Chen, Vladimir Karpukhin, Stan Peshterliev, Dmytro Okhonko, Michael Schlichtkrull, Sonal Gupta, Yashar Mehdad, and Wen tau Yih. 2021.
\newblock \href {https://arxiv.org/abs/2101.00133} {Neurips 2020 efficientqa competition: Systems, analyses and lessons learned}.
\newblock \emph{Preprint}, arXiv:2101.00133.

\bibitem[{Mitchell et~al.(2022)Mitchell, Lin, Bosselut, Finn, and Manning}]{DBLP:conf/iclr/MitchellLBFM22}
Eric Mitchell, Charles Lin, Antoine Bosselut, Chelsea Finn, and Christopher~D. Manning. 2022.
\newblock \href {https://openreview.net/forum?id=0DcZxeWfOPt} {Fast model editing at scale}.
\newblock In \emph{The Tenth International Conference on Learning Representations, {ICLR} 2022, Virtual Event, April 25-29, 2022}. OpenReview.net.

\bibitem[{OpenAI(2023)}]{openai2023gpt4}
OpenAI. 2023.
\newblock \href {https://arxiv.org/abs/2303.08774} {Gpt-4 technical report}.
\newblock \emph{Preprint}, arXiv:2303.08774.

\bibitem[{OpenAI(2024)}]{gpt4o}
OpenAI. 2024.
\newblock \href {https://openai.com/index/hello-gpt-4o/} {Hello gpt-4o}.
\newblock \emph{OpenAI blog}.

\bibitem[{Ouyang et~al.(2022{\natexlab{a}})Ouyang, Wu, Jiang, Almeida, Wainwright, Mishkin, Zhang, Agarwal, Slama, Ray et~al.}]{ouyang2022training}
Long Ouyang, Jeffrey Wu, Xu~Jiang, Diogo Almeida, Carroll Wainwright, Pamela Mishkin, Chong Zhang, Sandhini Agarwal, Katarina Slama, Alex Ray, et~al. 2022{\natexlab{a}}.
\newblock Training language models to follow instructions with human feedback.
\newblock \emph{Advances in Neural Information Processing Systems}, 35:27730--27744.

\bibitem[{Ouyang et~al.(2022{\natexlab{b}})Ouyang, Wu, Jiang, Almeida, Wainwright, Mishkin, Zhang, Agarwal, Slama, Ray, Schulman, Hilton, Kelton, Miller, Simens, Askell, Welinder, Christiano, Leike, and Lowe}]{DBLP:conf/nips/Ouyang0JAWMZASR22}
Long Ouyang, Jeffrey Wu, Xu~Jiang, Diogo Almeida, Carroll~L. Wainwright, Pamela Mishkin, Chong Zhang, Sandhini Agarwal, Katarina Slama, Alex Ray, John Schulman, Jacob Hilton, Fraser Kelton, Luke Miller, Maddie Simens, Amanda Askell, Peter Welinder, Paul~F. Christiano, Jan Leike, and Ryan Lowe. 2022{\natexlab{b}}.
\newblock \href {http://papers.nips.cc/paper\_files/paper/2022/hash/b1efde53be364a73914f58805a001731-Abstract-Conference.html} {Training language models to follow instructions with human feedback}.
\newblock In \emph{Advances in Neural Information Processing Systems 35: Annual Conference on Neural Information Processing Systems 2022, NeurIPS 2022, New Orleans, LA, USA, November 28 - December 9, 2022}.

\bibitem[{Ovadia et~al.(2024)Ovadia, Brief, Mishaeli, and Elisha}]{ovadia2024finetuning}
Oded Ovadia, Menachem Brief, Moshik Mishaeli, and Oren Elisha. 2024.
\newblock \href {https://arxiv.org/abs/2312.05934} {Fine-tuning or retrieval? comparing knowledge injection in llms}.
\newblock \emph{Preprint}, arXiv:2312.05934.

\bibitem[{Reyes et~al.(2021)Reyes, Blanco, Doroon, Limana, and Torcende}]{article}
Englevert Reyes, Ron Blanco, Defanee Doroon, Jay Limana, and Ana Torcende. 2021.
\newblock \href {https://doi.org/10.32871/rmrj2109.02.06} {Feynman technique as a heutagogical learning strategy for independent and remote learning}.
\newblock \emph{Recoletos Multidisciplinary Research Journal}, 9:1--13.

\bibitem[{Saito et~al.(2024)Saito, Sohn, Lee, and Ushiku}]{saito2024answer}
Kuniaki Saito, Kihyuk Sohn, Chen-Yu Lee, and Yoshitaka Ushiku. 2024.
\newblock \href {https://arxiv.org/abs/2402.12170} {Where is the answer? investigating positional bias in language model knowledge extraction}.
\newblock \emph{Preprint}, arXiv:2402.12170.

\bibitem[{Talmor et~al.(2019)Talmor, Herzig, Lourie, and Berant}]{talmor-etal-2019-commonsenseqa}
Alon Talmor, Jonathan Herzig, Nicholas Lourie, and Jonathan Berant. 2019.
\newblock \href {https://doi.org/10.18653/v1/N19-1421} {{C}ommonsense{QA}: A question answering challenge targeting commonsense knowledge}.
\newblock In \emph{Proceedings of the 2019 Conference of the North {A}merican Chapter of the Association for Computational Linguistics: Human Language Technologies, Volume 1 (Long and Short Papers)}, pages 4149--4158, Minneapolis, Minnesota. Association for Computational Linguistics.

\bibitem[{Tang and Yang(2024)}]{tang2024multihopragbenchmarkingretrievalaugmentedgeneration}
Yixuan Tang and Yi~Yang. 2024.
\newblock \href {https://arxiv.org/abs/2401.15391} {Multihop-rag: Benchmarking retrieval-augmented generation for multi-hop queries}.
\newblock \emph{Preprint}, arXiv:2401.15391.

\bibitem[{Team et~al.(2024)Team, Mesnard, Hardin, Dadashi, Bhupatiraju, Pathak, Sifre, Rivi{\`e}re, Kale, Love et~al.}]{team2024gemma}
Gemma Team, Thomas Mesnard, Cassidy Hardin, Robert Dadashi, Surya Bhupatiraju, Shreya Pathak, Laurent Sifre, Morgane Rivi{\`e}re, Mihir~Sanjay Kale, Juliette Love, et~al. 2024.
\newblock Gemma: Open models based on gemini research and technology.
\newblock \emph{arXiv preprint arXiv:2403.08295}.

\bibitem[{Touvron et~al.(2023{\natexlab{a}})Touvron, Martin, Stone, Albert, Almahairi, Babaei, Bashlykov, Batra, Bhargava, Bhosale, Bikel, Blecher, Ferrer, Chen, Cucurull, Esiobu, Fernandes, Fu, Fu, Fuller, Gao, Goswami, Goyal, Hartshorn, Hosseini, Hou, Inan, Kardas, Kerkez, Khabsa, Kloumann, Korenev, Koura, Lachaux, Lavril, Lee, Liskovich, Lu, Mao, Martinet, Mihaylov, Mishra, Molybog, Nie, Poulton, Reizenstein, Rungta, Saladi, Schelten, Silva, Smith, Subramanian, Tan, Tang, Taylor, Williams, Kuan, Xu, Yan, Zarov, Zhang, Fan, Kambadur, Narang, Rodriguez, Stojnic, Edunov, and Scialom}]{touvron2023llama}
Hugo Touvron, Louis Martin, Kevin Stone, Peter Albert, Amjad Almahairi, Yasmine Babaei, Nikolay Bashlykov, Soumya Batra, Prajjwal Bhargava, Shruti Bhosale, Dan Bikel, Lukas Blecher, Cristian~Canton Ferrer, Moya Chen, Guillem Cucurull, David Esiobu, Jude Fernandes, Jeremy Fu, Wenyin Fu, Brian Fuller, Cynthia Gao, Vedanuj Goswami, Naman Goyal, Anthony Hartshorn, Saghar Hosseini, Rui Hou, Hakan Inan, Marcin Kardas, Viktor Kerkez, Madian Khabsa, Isabel Kloumann, Artem Korenev, Punit~Singh Koura, Marie-Anne Lachaux, Thibaut Lavril, Jenya Lee, Diana Liskovich, Yinghai Lu, Yuning Mao, Xavier Martinet, Todor Mihaylov, Pushkar Mishra, Igor Molybog, Yixin Nie, Andrew Poulton, Jeremy Reizenstein, Rashi Rungta, Kalyan Saladi, Alan Schelten, Ruan Silva, Eric~Michael Smith, Ranjan Subramanian, Xiaoqing~Ellen Tan, Binh Tang, Ross Taylor, Adina Williams, Jian~Xiang Kuan, Puxin Xu, Zheng Yan, Iliyan Zarov, Yuchen Zhang, Angela Fan, Melanie Kambadur, Sharan Narang, Aurelien Rodriguez, Robert Stojnic, Sergey Edunov, and Thomas
  Scialom. 2023{\natexlab{a}}.
\newblock \href {https://arxiv.org/abs/2307.09288} {Llama 2: Open foundation and fine-tuned chat models}.
\newblock \emph{Preprint}, arXiv:2307.09288.

\bibitem[{Touvron et~al.(2023{\natexlab{b}})Touvron, Martin, Stone, Albert, Almahairi, Babaei, Bashlykov, Batra, Bhargava, Bhosale, Bikel, Blecher, Ferrer, Chen, Cucurull, Esiobu, Fernandes, Fu, Fu, Fuller, Gao, Goswami, Goyal, Hartshorn, Hosseini, Hou, Inan, Kardas, Kerkez, Khabsa, Kloumann, Korenev, Koura, Lachaux, Lavril, Lee, Liskovich, Lu, Mao, Martinet, Mihaylov, Mishra, Molybog, Nie, Poulton, Reizenstein, Rungta, Saladi, Schelten, Silva, Smith, Subramanian, Tan, Tang, Taylor, Williams, Kuan, Xu, Yan, Zarov, Zhang, Fan, Kambadur, Narang, Rodriguez, Stojnic, Edunov, and Scialom}]{touvron2023llama2openfoundation}
Hugo Touvron, Louis Martin, Kevin Stone, Peter Albert, Amjad Almahairi, Yasmine Babaei, Nikolay Bashlykov, Soumya Batra, Prajjwal Bhargava, Shruti Bhosale, Dan Bikel, Lukas Blecher, Cristian~Canton Ferrer, Moya Chen, Guillem Cucurull, David Esiobu, Jude Fernandes, Jeremy Fu, Wenyin Fu, Brian Fuller, Cynthia Gao, Vedanuj Goswami, Naman Goyal, Anthony Hartshorn, Saghar Hosseini, Rui Hou, Hakan Inan, Marcin Kardas, Viktor Kerkez, Madian Khabsa, Isabel Kloumann, Artem Korenev, Punit~Singh Koura, Marie-Anne Lachaux, Thibaut Lavril, Jenya Lee, Diana Liskovich, Yinghai Lu, Yuning Mao, Xavier Martinet, Todor Mihaylov, Pushkar Mishra, Igor Molybog, Yixin Nie, Andrew Poulton, Jeremy Reizenstein, Rashi Rungta, Kalyan Saladi, Alan Schelten, Ruan Silva, Eric~Michael Smith, Ranjan Subramanian, Xiaoqing~Ellen Tan, Binh Tang, Ross Taylor, Adina Williams, Jian~Xiang Kuan, Puxin Xu, Zheng Yan, Iliyan Zarov, Yuchen Zhang, Angela Fan, Melanie Kambadur, Sharan Narang, Aurelien Rodriguez, Robert Stojnic, Sergey Edunov, and Thomas
  Scialom. 2023{\natexlab{b}}.
\newblock \href {https://arxiv.org/abs/2307.09288} {Llama 2: Open foundation and fine-tuned chat models}.
\newblock \emph{Preprint}, arXiv:2307.09288.

\bibitem[{van~de Kar et~al.(2022)van~de Kar, Xia, Chen, and Artetxe}]{van-de-kar-etal-2022-dont}
Mozes van~de Kar, Mengzhou Xia, Danqi Chen, and Mikel Artetxe. 2022.
\newblock \href {https://doi.org/10.18653/v1/2022.emnlp-main.509} {Don{'}t prompt, search! mining-based zero-shot learning with language models}.
\newblock In \emph{Proceedings of the 2022 Conference on Empirical Methods in Natural Language Processing}, pages 7508--7520, Abu Dhabi, United Arab Emirates. Association for Computational Linguistics.

\bibitem[{Vu et~al.(2023)Vu, Iyyer, Wang, Constant, Wei, Wei, Tar, Sung, Zhou, Le, and Luong}]{vu2023freshllms}
Tu~Vu, Mohit Iyyer, Xuezhi Wang, Noah Constant, Jerry Wei, Jason Wei, Chris Tar, Yun-Hsuan Sung, Denny Zhou, Quoc Le, and Thang Luong. 2023.
\newblock \href {https://arxiv.org/abs/2310.03214} {Freshllms: Refreshing large language models with search engine augmentation}.
\newblock \emph{Preprint}, arXiv:2310.03214.

\bibitem[{Wang et~al.(2023)Wang, Liu, Yue, Tang, Zhang, Jiayang, Yao, Gao, Hu, Qi, Wang, Yang, Wang, Xie, Zhang, and Zhang}]{wang2023survey}
Cunxiang Wang, Xiaoze Liu, Yuanhao Yue, Xiangru Tang, Tianhang Zhang, Cheng Jiayang, Yunzhi Yao, Wenyang Gao, Xuming Hu, Zehan Qi, Yidong Wang, Linyi Yang, Jindong Wang, Xing Xie, Zheng Zhang, and Yue Zhang. 2023.
\newblock \href {https://arxiv.org/abs/2310.07521} {Survey on factuality in large language models: Knowledge, retrieval and domain-specificity}.
\newblock \emph{Preprint}, arXiv:2310.07521.

\bibitem[{Wang et~al.(2024)Wang, Zhang, Su, and Zhu}]{wang2024comprehensivesurveycontinuallearning}
Liyuan Wang, Xingxing Zhang, Hang Su, and Jun Zhu. 2024.
\newblock \href {https://arxiv.org/abs/2302.00487} {A comprehensive survey of continual learning: Theory, method and application}.
\newblock \emph{Preprint}, arXiv:2302.00487.

\bibitem[{Wei et~al.(2022)Wei, Bosma, Zhao, Guu, Yu, Lester, Du, Dai, and Le}]{wei2022finetuned}
Jason Wei, Maarten Bosma, Vincent Zhao, Kelvin Guu, Adams~Wei Yu, Brian Lester, Nan Du, Andrew~M. Dai, and Quoc~V Le. 2022.
\newblock \href {https://openreview.net/forum?id=gEZrGCozdqR} {Finetuned language models are zero-shot learners}.
\newblock In \emph{International Conference on Learning Representations}.

\bibitem[{Wu et~al.(2024{\natexlab{a}})Wu, Wu, and Zou}]{wu2024faithful}
Kevin Wu, Eric Wu, and James Zou. 2024{\natexlab{a}}.
\newblock \href {https://arxiv.org/abs/2404.10198} {How faithful are rag models? quantifying the tug-of-war between rag and llms' internal prior}.
\newblock \emph{Preprint}, arXiv:2404.10198.

\bibitem[{Wu et~al.(2024{\natexlab{b}})Wu, Xie, Chen, Zhu, Zhang, and Xiao}]{wu2024easily}
Siye Wu, Jian Xie, Jiangjie Chen, Tinghui Zhu, Kai Zhang, and Yanghua Xiao. 2024{\natexlab{b}}.
\newblock \href {https://arxiv.org/abs/2404.03302} {How easily do irrelevant inputs skew the responses of large language models?}
\newblock \emph{Preprint}, arXiv:2404.03302.

\bibitem[{Xiang et~al.(2024)Xiang, Wu, Zhong, Wagner, Chen, and Mittal}]{xiang2024certifiably}
Chong Xiang, Tong Wu, Zexuan Zhong, David Wagner, Danqi Chen, and Prateek Mittal. 2024.
\newblock \href {https://arxiv.org/abs/2405.15556} {Certifiably robust rag against retrieval corruption}.
\newblock \emph{Preprint}, arXiv:2405.15556.

\bibitem[{Xiaofei et~al.(2017)Xiaofei, Qing, Yanyan, Weifeng, and Wenzhi}]{xiaofei2017application}
Wang Xiaofei, Chen Qing, Sun Yanyan, Tong Weifeng, and Niu Wenzhi. 2017.
\newblock The application of the feynman technique for practical teaching of prosthodontics.
\newblock \emph{Chinese Journal of Medical Education}, 41(9):822.

\bibitem[{Xu et~al.(2023)Xu, Namazifar, Hazarika, Padmakumar, Liu, and Hakkani{-}T{\"{u}}r}]{DBLP:journals/corr/abs-2302-09170}
Yan Xu, Mahdi Namazifar, Devamanyu Hazarika, Aishwarya Padmakumar, Yang Liu, and Dilek Hakkani{-}T{\"{u}}r. 2023.
\newblock \href {https://doi.org/10.48550/ARXIV.2302.09170} {{KILM:} knowledge injection into encoder-decoder language models}.
\newblock \emph{CoRR}, abs/2302.09170.

\bibitem[{Yang et~al.(2024)Yang, Yang, Hui, Zheng, Yu, Zhou, Li, Li, Liu, Huang, Dong, Wei, Lin, Tang, Wang, Yang, Tu, Zhang, Ma, Yang, Xu, Zhou, Bai, He, Lin, Dang, Lu, Chen, Yang, Li, Xue, Ni, Zhang, Wang, Peng, Men, Gao, Lin, Wang, Bai, Tan, Zhu, Li, Liu, Ge, Deng, Zhou, Ren, Zhang, Wei, Ren, Liu, Fan, Yao, Zhang, Wan, Chu, Liu, Cui, Zhang, Guo, and Fan}]{yang2024qwen2technicalreport}
An~Yang, Baosong Yang, Binyuan Hui, Bo~Zheng, Bowen Yu, Chang Zhou, Chengpeng Li, Chengyuan Li, Dayiheng Liu, Fei Huang, Guanting Dong, Haoran Wei, Huan Lin, Jialong Tang, Jialin Wang, Jian Yang, Jianhong Tu, Jianwei Zhang, Jianxin Ma, Jianxin Yang, Jin Xu, Jingren Zhou, Jinze Bai, Jinzheng He, Junyang Lin, Kai Dang, Keming Lu, Keqin Chen, Kexin Yang, Mei Li, Mingfeng Xue, Na~Ni, Pei Zhang, Peng Wang, Ru~Peng, Rui Men, Ruize Gao, Runji Lin, Shijie Wang, Shuai Bai, Sinan Tan, Tianhang Zhu, Tianhao Li, Tianyu Liu, Wenbin Ge, Xiaodong Deng, Xiaohuan Zhou, Xingzhang Ren, Xinyu Zhang, Xipin Wei, Xuancheng Ren, Xuejing Liu, Yang Fan, Yang Yao, Yichang Zhang, Yu~Wan, Yunfei Chu, Yuqiong Liu, Zeyu Cui, Zhenru Zhang, Zhifang Guo, and Zhihao Fan. 2024.
\newblock \href {https://arxiv.org/abs/2407.10671} {Qwen2 technical report}.
\newblock \emph{Preprint}, arXiv:2407.10671.

\bibitem[{Yao et~al.(2023)Yao, Wang, Tian, Cheng, Li, Deng, Chen, and Zhang}]{yao-etal-2023-editing}
Yunzhi Yao, Peng Wang, Bozhong Tian, Siyuan Cheng, Zhoubo Li, Shumin Deng, Huajun Chen, and Ningyu Zhang. 2023.
\newblock \href {https://doi.org/10.18653/v1/2023.emnlp-main.632} {Editing large language models: Problems, methods, and opportunities}.
\newblock In \emph{Proceedings of the 2023 Conference on Empirical Methods in Natural Language Processing}, pages 10222--10240, Singapore. Association for Computational Linguistics.

\bibitem[{Zhang et~al.(2024{\natexlab{a}})Zhang, Yao, Tian, Wang, Deng, Wang, Xi, Mao, Zhang, Ni, Cheng, Xu, Xu, Gu, Jiang, Xie, Huang, Liang, Zhang, Zhu, Zhou, and Chen}]{zhang2024comprehensive}
Ningyu Zhang, Yunzhi Yao, Bozhong Tian, Peng Wang, Shumin Deng, Mengru Wang, Zekun Xi, Shengyu Mao, Jintian Zhang, Yuansheng Ni, Siyuan Cheng, Ziwen Xu, Xin Xu, Jia-Chen Gu, Yong Jiang, Pengjun Xie, Fei Huang, Lei Liang, Zhiqiang Zhang, Xiaowei Zhu, Jun Zhou, and Huajun Chen. 2024{\natexlab{a}}.
\newblock \href {https://arxiv.org/abs/2401.01286} {A comprehensive study of knowledge editing for large language models}.
\newblock \emph{Preprint}, arXiv:2401.01286.

\bibitem[{Zhang et~al.(2024{\natexlab{b}})Zhang, Patil, Jain, Shen, Zaharia, Stoica, and Gonzalez}]{zhang2024raft}
Tianjun Zhang, Shishir~G. Patil, Naman Jain, Sheng Shen, Matei Zaharia, Ion Stoica, and Joseph~E. Gonzalez. 2024{\natexlab{b}}.
\newblock \href {https://arxiv.org/abs/2403.10131} {Raft: Adapting language model to domain specific rag}.
\newblock \emph{Preprint}, arXiv:2403.10131.

\bibitem[{Zhang et~al.(2022)Zhang, Peng, Gao, and Meng}]{zhang-etal-2022-toward-self}
Xiaoying Zhang, Baolin Peng, Jianfeng Gao, and Helen Meng. 2022.
\newblock \href {https://doi.org/10.18653/v1/2022.sigdial-1.49} {Toward self-learning end-to-end task-oriented dialog systems}.
\newblock In \emph{Proceedings of the 23rd Annual Meeting of the Special Interest Group on Discourse and Dialogue}, pages 516--530, Edinburgh, UK. Association for Computational Linguistics.

\bibitem[{Zhang et~al.(2023)Zhang, Peng, Li, Zhou, and Meng}]{zhang-etal-2023-sgp}
Xiaoying Zhang, Baolin Peng, Kun Li, Jingyan Zhou, and Helen Meng. 2023.
\newblock \href {https://doi.org/10.18653/v1/2023.findings-emnlp.891} {{SGP}-{TOD}: Building task bots effortlessly via schema-guided {LLM} prompting}.
\newblock In \emph{Findings of the Association for Computational Linguistics: EMNLP 2023}, pages 13348--13369, Singapore. Association for Computational Linguistics.

\bibitem[{Zhang et~al.(2024{\natexlab{c}})Zhang, Peng, Tian, Zhou, Jin, Song, Mi, and Meng}]{zhang2024self}
Xiaoying Zhang, Baolin Peng, Ye~Tian, Jingyan Zhou, Lifeng Jin, Linfeng Song, Haitao Mi, and Helen Meng. 2024{\natexlab{c}}.
\newblock Self-alignment for factuality: Mitigating hallucinations in llms via self-evaluation.
\newblock \emph{arXiv preprint arXiv:2402.09267}.

\bibitem[{Zhang et~al.(2024{\natexlab{d}})Zhang, Peng, Tian, Zhou, Jin, Song, Mi, and Meng}]{zhang-etal-2024-self}
Xiaoying Zhang, Baolin Peng, Ye~Tian, Jingyan Zhou, Lifeng Jin, Linfeng Song, Haitao Mi, and Helen Meng. 2024{\natexlab{d}}.
\newblock \href {https://doi.org/10.18653/v1/2024.acl-long.107} {Self-alignment for factuality: Mitigating hallucinations in {LLM}s via self-evaluation}.
\newblock In \emph{Proceedings of the 62nd Annual Meeting of the Association for Computational Linguistics (Volume 1: Long Papers)}, pages 1946--1965, Bangkok, Thailand. Association for Computational Linguistics.

\bibitem[{Zhang et~al.(2025)Zhang, Peng, Zhang, Guo, Wu, Chen, Ke, Meng, and Sun}]{zhang2025will}
Xiaoying Zhang, Da~Peng, Yipeng Zhang, Zonghao Guo, Chengyue Wu, Chi Chen, Wei Ke, Helen Meng, and Maosong Sun. 2025.
\newblock Will pre-training ever end? a first step toward next-generation foundation mllms via self-improving systematic cognition.
\newblock \emph{arXiv preprint arXiv:2503.12303}.

\bibitem[{Zhang et~al.(2024{\natexlab{e}})Zhang, Chen, Bai, Kang, Guo, and Zhang}]{zhang-etal-2024-question}
Yu~Zhang, Kehai Chen, Xuefeng Bai, Zhao Kang, Quanjiang Guo, and Min Zhang. 2024{\natexlab{e}}.
\newblock \href {https://doi.org/10.18653/v1/2024.findings-emnlp.524} {Question-guided knowledge graph re-scoring and injection for knowledge graph question answering}.
\newblock In \emph{Findings of the Association for Computational Linguistics: EMNLP 2024}, pages 8972--8985, Miami, Florida, USA. Association for Computational Linguistics.

\bibitem[{Zheng et~al.(2023)Zheng, Li, Dong, Fan, Wu, Xu, and Chang}]{DBLP:conf/emnlp/ZhengLDFWXC23}
Ce~Zheng, Lei Li, Qingxiu Dong, Yuxuan Fan, Zhiyong Wu, Jingjing Xu, and Baobao Chang. 2023.
\newblock \href {https://aclanthology.org/2023.emnlp-main.296} {Can we edit factual knowledge by in-context learning?}
\newblock In \emph{Proceedings of the 2023 Conference on Empirical Methods in Natural Language Processing, {EMNLP} 2023, Singapore, December 6-10, 2023}, pages 4862--4876. Association for Computational Linguistics.

\bibitem[{Zheng et~al.(2024)Zheng, Mishra, Chen, Cheng, Chi, Le, and Zhou}]{zheng2024take}
Huaixiu~Steven Zheng, Swaroop Mishra, Xinyun Chen, Heng-Tze Cheng, Ed~H. Chi, Quoc~V Le, and Denny Zhou. 2024.
\newblock \href {https://openreview.net/forum?id=3bq3jsvcQ1} {Take a step back: Evoking reasoning via abstraction in large language models}.
\newblock In \emph{The Twelfth International Conference on Learning Representations}.

\bibitem[{Zhou et~al.(2023)Zhou, Liu, Xu, Iyer, Sun, Mao, Ma, Efrat, Yu, Yu, Zhang, Ghosh, Lewis, Zettlemoyer, and Levy}]{zhou2023lima}
Chunting Zhou, Pengfei Liu, Puxin Xu, Srini Iyer, Jiao Sun, Yuning Mao, Xuezhe Ma, Avia Efrat, Ping Yu, Lili Yu, Susan Zhang, Gargi Ghosh, Mike Lewis, Luke Zettlemoyer, and Omer Levy. 2023.
\newblock \href {https://arxiv.org/abs/2305.11206} {Lima: Less is more for alignment}.
\newblock \emph{Preprint}, arXiv:2305.11206.

\end{thebibliography}

\appendix

\clearpage
\tableofcontents
\addtocontents{toc}{\protect\setcounter{tocdepth}{2}}
\clearpage

\section{Additional Efforts for Knowledge Injection}
\label{sec:additional_related_work}
Knowledge editing~\cite{DBLP:conf/emnlp/ZhengLDFWXC23, yao-etal-2023-editing, jiang2024learning, zhang2024comprehensive} and retrieval-augmented generation (RAG)~\cite{lewis2021retrievalaugmented, ovadia2024finetuning, jeong2024adaptiverag} are recognized as two related research initiatives in the field of knowledge injection.

$(\RN{1})$ \textit{Knowledge editing} \cite{DBLP:conf/iclr/MitchellLBFM22, DBLP:conf/emnlp/ZhengLDFWXC23, yao-etal-2023-editing, jiang2024learning, zhang2024comprehensive} concentrates on rectifying outdated or inaccurate factual knowledge stored in the model, without affecting other facts. In contrast, our focus lies in enabling LLMs to efficiently acquire knowledge from raw corpora.

$(\RN{2})$ \textit{Retrieval-augmented generation (RAG)} \cite{lewis2021retrievalaugmented,vu2023freshllms, ovadia2024finetuning, jeong2024adaptiverag} equips LLMs with new knowledge by augmenting off-the-shelf LLMs with retrieved knowledge from external sources. However, its performance is vulnerable to irrelevant or malicious information in the retrieval results \cite{contextualai2024rag}, potentially leading to inaccurate responses \cite{zhang2024raft, wu2024easily, xiang2024certifiably}. Moreover, recent findings \cite{wu2024faithful} emphasize an underlying tension between a model's prior knowledge and the information presented in retrieved documents. Consequently, this paper primarily focuses on exploring the injection of knowledge into the parameters of LLMs.

\section{Wiki-Newpages-2023-QA: Datasets for Studying LLM Knowledge Acquisition} 
\label{sec:details_data_construction}
To explore the knowledge acquisition capabilities of LLMs from new documents, \wrt memorization, extraction and reasoning, we introduce the Wiki-Newpages-2023-QA datasets, which are carefully designed to minimize overlap with the initial pre-training corpus. These datasets comprise new document corpora for studying knowledge memorization and associated QA datasets for two vital knowledge-intensive tasks: open-ended generation and NLI for examining extraction and reasoning, respectively. We provide the details on dataset construction in the following subsections.

\subsection{Document Collection} 
\label{sec:doc_corpus} 
Given the well-structured nature of Wikipedia articles, which encompass extensive factual information and cover a wide range of topics across various domains, we gather documents from Wikipedia NewPages\footnote{\url{https://en.wikipedia.org/wiki/Special:NewPages}}. This collection includes new articles from diverse domains published after the pre-training cut-off time of the LLMs being evaluated, allowing us to largely ensure that the models have not been exposed to these facts. To construct the document corpus, we specifically gather articles from September to October 2023, resulting in a total of 4,257 articles.\footnote{The pre-training cut-off for the \textsc{Llama2} family models used in this study is 2022.} Following \citet{jiang2024instructiontuned}, we only utilize the first paragraph of each article, which provides a comprehensive summary and sufficient factual information.

\subsection{QA Pair Generation} 
\label{sec:qa_generation}

To gather QA pairs, we utilize GPT-4 \cite{openai2023gpt4} along with our manually curated prompts to generate a variety of questions and their corresponding answers, aiming to cover all factual information within the given document. Note that GPT-4 is solely used to convert document knowledge into QA pairs \textit{without introducing any additional information}. It can be replaced by any other model.

Specifically, we construct QA datasets for the open-ended generation and NLI tasks by employing the prompts shown in Table \ref{tab:wikiqa_gen_prompt} and Table \ref{tab:wikiqa_nli_prompt}, respectively. 
A simplified example document with associated QA pairs is provided in Table \ref{tab:qa_gen_nli_simple_example}. More detailed examples can be found in Appendix \ref{sec:detailed_qa_example}.
\begin{table}[!t]
\centering
\fontsize{9.5}{10}\selectfont
\begin{tabular}{p{7cm}}
\toprule
\textbf{Document:} <Sawyer Gipson-Long - Wikipedia> Alec Sawyer Gipson-Long \textcolor{blue}{(born December 12, 1997)} is an American professional baseball pitcher for ...\\
\midrule
\textbf{QA Pair Example for Generation Task} \\
\midrule
\textbf{Question:} When was Sawyer Gipson-Long \textcolor{blue}{born}?\\ 
\textbf{Answer:} \textcolor{blue}{December 12, 1997}.\\ 
\midrule
\textbf{QA Pair Example for NLI Task} \\
\midrule
\textbf{Question:} Based on the paragraph above can we conclude that  <Alec Sawyer Gipson-Long> Sawyer Gipson-Long was \textcolor{blue}{born in December 1997}. \\
Options: -Yes; -It's impossible to say; -No\\ 
\textbf{Answer:} \textcolor{blue}{Yes}\\ 
\bottomrule
\end{tabular}
\caption{A simplified example of a document and its associated QA pair for the open-ended generation task. Factual information related to the QA pairs is denoted in \textcolor{blue}{blue}.}
\label{tab:qa_gen_nli_simple_example}
\end{table}


\subsection{Splitting} 
To enable a comprehensive analysis in single-domain, multi-domain, and cross-domain situations, we develop three datasets and divide them into training and testing subsets, \textit{ensuring zero knowledge overlap}.

\paragraph{Dataset Splitting.} 
We create three datasets: Wiki-Newpages-2023-10-Bio (Wiki-Bio), Wiki-Newpages-2023-10-Multi (Wiki-Multi), and Wiki-Newpages-2023-(9)10-Film (Wiki-Film). Specifically, we randomly select 1,263 biographical documents to curate Wiki-Bio, choose 2,104 documents covering various topics for constructing Wiki-Multi, and compile 955 film documents for producing Wiki-Film, using the assembled document corpus along with their associated QA pairs.

\paragraph{Train-test Splitting.} 
We partition the Wiki-Bio and Wiki-Multi datasets, comprising the document corpus and the derived QA datasets, into training and testing subsets for conducting evaluations in single-domain and multi-domain contexts. We directly utilize the Wiki-Film dataset as the test set for the cross-domain scenario. It is crucial to note that the training QA datasets only contain the QA pairs from open-ended generation tasks, ensuring a fair comparison with existing knowledge injection approaches. We provide extensive statistical information for the three datasets in Table \ref{tab:wiki_data} and a thorough analysis of the QA types in Appendix \ref{sec:analysis_qa_type}.

\begin{table*}[!t]
\setlength\tabcolsep{1pt}
  \centering
  \begin{threeparttable}
  \fontsize{9}{9}
  \selectfont
    \begin{tabular}{lccc}
    \toprule
    \multirow{2}{*}{\textbf{Method}}&
    \multicolumn{3}{c}{\textbf{Training Data in Each Stage}}
     \cr\cmidrule(lr){2-4} 
     &\textbf{Stage 1}&\textbf{Stage 2} & \textbf{Stage 3} \cr
    \midrule
    Continued Pre-training& \multicolumn{3}{c}{ \circled{1} \colorbox{mypink1!90}{test doc} }\\ 
    Standard Ins.-tuning &  \circled{1} \colorbox{mygrey1}{train doc}\&\colorbox{mypink1!90}{test doc}  & \multicolumn{2}{c}{\circled{2} \colorbox{myblue1}{train QA}  }\\
    PIT &  \circled{1} \colorbox{myblue1}{train QA}\colorbox{mygrey1}{train doc}  &  \multicolumn{2}{c}{\circled{2} \colorbox{mypink1!90}{test doc}  }\\
    \selfpt{} & \circled{1}  \colorbox{myblue1}{train QA}\&\colorbox{mygrey1}{train doc w/ self-teaching tasks}  &  \circled{2} \colorbox{myblue1}{train QA}\&\colorbox{mypink1!90}{test doc} &  \circled{3}  \colorbox{mypink1!90}{test doc} \\
    \midrule
    \multicolumn{4}{l}{\textit{\textbf{Variants of \selfpt{}}}} \\
    \midrule
    \selfpt{} w/o Review& \circled{1}  \colorbox{myblue1}{train QA}\&\colorbox{mygrey1}{train doc w/ self-teaching tasks}  &  \multicolumn{2}{c}{ \circled{2} \colorbox{mypink1!90}{test doc} }\\
    \selfpt{} via Read. & \circled{1} \colorbox{myblue1}{train QA}\&\colorbox{mygrey1}{train doc (reading-comp. format)}  & \multicolumn{2}{c}{ \circled{2} \colorbox{mypink1!90}{test doc}  }\\
    \selfpt{} w/ Pre-Review &  \circled{1} \colorbox{myblue1}{train QA}\&\colorbox{mygrey1}{train doc w/ self-teaching tasks}  &  \circled{2} \colorbox{myblue1}{train QA}\&\colorbox{mygrey1}{train doc}  &  \circled{3}  \colorbox{mypink1!90}{test doc} \\
    \midrule
    \multicolumn{4}{l}{\textit{\textbf{Additional Compared Methods}}} \\
    \midrule
    Standard Ins.-Tuning w/o Forget. &  \circled{1} \colorbox{mygrey1}{train doc}\&\colorbox{mypink1!90}{test doc}  & \multicolumn{2}{c}{\circled{2} \colorbox{myblue1}{train QA}\&\colorbox{mypink1!90}{test doc} } \\

    PIT$^{++}$ &  \circled{1} \colorbox{myblue1}{train QA}  & \circled{2} \colorbox{myblue1}{train QA}\colorbox{mygrey1}{train doc} &  \circled{3} \colorbox{mypink1!90}{test doc} \\
    Mixed Training & \multicolumn{3}{c}{ \circled{1} \colorbox{mygrey1}{train doc}\&\colorbox{myblue1}{train QA}\&\colorbox{mypink1!90}{test doc} } \\

    \bottomrule  
    \end{tabular}
  \end{threeparttable}
  
  \caption{Depiction of the training stages and associated datasets employed in the compared methods. ``Train doc w/ self-teaching tasks'': the training documents presented together with the self-teaching tasks. ``Reading-comp. format'': reading-comprehension format. ``Forget.'': ``Forgetting''.\protect\footnotemark}
  \label{tab:method_traing_stages_full}
  \vspace{-2mm}
\end{table*}

\footnotetext{To ensure a fair comparison, all compared approaches train on the test documents for 3 epochs in total, regardless of the number of training stages. For continued pre-training, which is observed to struggle in grasping new knowledge, we train the models for 5 epochs.}

\begin{table*}[!t]
\setlength\tabcolsep{1.5pt}
  \centering
  \begin{threeparttable}
  \fontsize{9}{9}
  \selectfont
    \begin{tabular}{lcccccccccccc}
    \toprule
    \multirow{4}{*}{\textbf{Method}}&
    \multicolumn{7}{c}{\textbf{Wiki-Newpages-2023-QA (Acquisition)}} 
     &\multicolumn{2}{c}{\textbf{NQ (Reten.)}} &\textbf{CSQA (Reten.)}\cr\cmidrule(lr){2-8} \cmidrule(lr){9-10} \cmidrule(lr){11-11}
     & \textbf{Mem.}
      &\multicolumn{5}{c}{\textbf{Extraction}}&\textbf{Reason.} &\multicolumn{2}{c}{\textbf{Extraction}}&\textbf{Reasoning} \cr\cmidrule(lr){2-2} \cmidrule(lr){3-7} \cmidrule(lr){8-8} \cmidrule(lr){9-10} \cmidrule(lr){11-11}
     & $\mathtt{PPL}$ ($\downarrow$)&\%  $\mathtt{Acc.}$ &\%  $\mathtt{EM}$ &\%  $\mathtt{F1}$&\%  $\mathtt{Rec.}$&\%  $\mathtt{Rouge}$ &\% $\mathtt{Acc.}$ &\%  $\mathtt{EM}$ &\%  $\mathtt{F1}$ &\%  $\mathtt{Acc.}$\cr
    \midrule
    \multicolumn{11}{c}{\textbf{Knowledge Acquisition on Wiki-Newpages-2023-10-Bio (Single-Domain Scenario)}}\cr
    \midrule
    \multicolumn{8}{l}{\textcolor{lightgray!99}{\textit{w/o Knowledge Injection}}}\cr
    Open-book w/ test doc &8.41&55.20& 31.83&64.48&75.55&62.10& 7.96&-&-&-\\
    Closed-book &8.41&4.68&2.87&14.63&16.98&15.07 &7.96& 16.05&24.67&53.40\\
    \midrule
   \multicolumn{8}{l}{\textcolor{lightgray!99}{\textit{w/ Knowledge Injection}}}\cr 
    Cont. Pre-training & 7.28&6.33&3.62&15.96&18.72&16.11 &15.09 & \textcolor{red}{16.00}&\textcolor{red}{24.11}&53.40\\
    
    Standard Ins.-tuning & 6.83& 6.94 & 5.13& 19.15& 19.05& 19.48&39.09& \textcolor{red}{15.72}& \textcolor{red}{23.67} & \textcolor{red}{51.84}  \\ 
      Standard Ins.-Tuning w/o Forget. & 2.82 & 9.35 & 7.09 & 21.25 & 21.72 & 21.51 & 36.08 & 16.05 & 24.88 & 54.30\\
     PIT & 2.08& 14.03 & 11.61& 27.15 & 28.86& 27.11 &11.93& \textcolor{red}{15.72}& 26.31 & 57.58\\
     PIT$^{++}$ & 1.78 & 22.78 & 20.06 & 37.11 & 37.62 & 37.06 & 42.25 & 16.39 & 25.67 & 57.00 \\  
      Mixed Training &1.42&  24.13 & 20.67 & 38.82 & 39.95 & 38.66 & \textbf{55.69} & \textbf{19.33} & \textbf{28.40} & 58.97 & \\ 
    \selfpt{} & \textbf{1.11}& \textbf{37.25}& \textbf{31.52} & \textbf{50.83}& \textbf{52.62} & \textbf{50.61} &44.31& 16.45 & 25.67 & \textbf{66.01}\\

    \midrule
        \multicolumn{11}{c}{\textbf{Knowledge Acquisition on Wiki-Newpages-2023-10-Multi (Multi-Domain Scenario)}}\cr
    \midrule
    \multicolumn{8}{l}{\textcolor{lightgray!99}{\textit{w/o Knowledge Injection}}}\cr
    Open-book w/ test doc & 7.84&48.93&26.63&60.37&71.71&58.54&6.33&-&-&-\\
    Closed-book & 7.84&4.53&2.73&16.19&18.63&16.38&6.33&16.05&24.67 & 53.40\\
    \midrule
   \multicolumn{8}{l}{\textcolor{lightgray!99}{\textit{w/ Knowledge Injection}}}\cr 
    Cont. Pre-training & 3.32&5.86&3.40&18.04&20.59&18.42&14.51&\textbf{17.02}&25.05&53.56\\
    
    Standard Ins.-tuning & 2.73& 8.66 & 5.73& 24.94& 25.64& 25.31 & 34.91& \textcolor{red}{15.60}& 26.26 & \textcolor{red}{52.74} \\ 
    
     
     PIT & 1.96& 14.31 & 8.72& 30.26 & 33.97& 30.22 &10.69&\textcolor{red}{15.55} & \textbf{27.02} & 55.12 \\ 
     
      \selfpt{} & \textbf{1.13} & \textbf{22.30}& \textbf{16.51} & \textbf{39.94}& \textbf{41.02} & \textbf{39.89}  & \textbf{50.65}& 16.34 & 25.85 & \textbf{69.29}\\ 
      

        \midrule
    \multicolumn{11}{c}{\textbf{Knowledge Acquisition on Wiki-Newpages-2023-(9)10-Film (Cross-Domain Scenario)}}\cr
    \midrule
    \multicolumn{8}{l}{\textcolor{lightgray!99}{\textit{w/o Knowledge Injection}}}\cr
    Open-book w/ film doc &8.30&57.38 & 34.45&68.64&78.92&66.31&7.35&-&-&-\\
    Closed-book &8.30&3.35&1.88&11.27&12.97&11.49&7.35&16.05&24.67&53.40\\
    \midrule
   \multicolumn{8}{l}{\textcolor{lightgray!99}{\textit{w/ Knowledge Injection}}}\cr 
    Cont. Pre-training &5.52 &3.46&2.30&11.83&14.30&11.98&12.04&\textbf{16.79}& 25.35 & 56.02\\
    Standard Ins.-tuning &2.83& 5.23 & 3.77& 16.15& 17.45& 16.45& \textbf{51.69}& \textcolor{red}{14.41}& 25.54 & \textcolor{red}{49.80} \\ 

     
     PIT&1.52& 6.39 & 4.50& 16.97 & 18.92& 17.10 &3.03&\textcolor{red}{13.06} & \textcolor{red}{23.42} & 54.38 \\

     \selfpt{} & \textbf{1.10} & \textbf{22.51}& \textbf{16.44} & \textbf{35.58}& \textbf{36.60} & \textbf{35.43} & 44.92& 16.77& \textbf{26.44} & \textbf{66.34}\\ 
    \bottomrule  
    \end{tabular}
  \end{threeparttable}
  \caption{Five-shot evaluation results on \llamaa{} for knowledge acquisition and retention in three scenarios: single-domain (top), multi-domain (middle), and cross-domain (bottom). Following \cite{jiang2024instructiontuned}, we also report results for: $(\RN{1})$ closed-book, where base LLMs are prompted with open-ended questions related to new knowledge in the test documents, and $(\RN{2})$ open-book w/ test doc, where base LLMs are prompted with questions along with relevant gold knowledge snippets from the test documents. Results that fall below the baseline performance are highlighted in \textcolor{red}{red}.}
  \label{tab:results_addition}
  \vspace{-3.5mm}
\end{table*}


\section{Evaluation Results on \llamaa{}} 
\label{sec:results_addition}
To thoroughly assess our proposed \selfpt{} method, we compare its efficiency against three other notable approaches: standard instruction-tuning without forgetting, PIT$^{++}$, and mixed training, as shown in Table \ref{tab:method_traing_stages_full}. The evaluation results, presented in Table \ref{tab:results_addition}, demonstrate that \selfpt{} consistently outperforms the alternatives. For instance, it improves EM by 11\% in the knowledge extraction task. 

Combining the results from Table~\ref{tab:results_addition} with the training strategies in Table~\ref{tab:method_traing_stages_full}, we emphasize the importance of first training the model to develop the ability to absorb new knowledge before training on test documents. This finding aligns with the conclusions of \citet{jiang2024instructiontuned}.

Notably, \selfpt{} enables the model to absorb new knowledge from incoming test documents more efficiently. Unlike mixed training—which requires retraining on both training documents, training QA, and test documents—\selfpt{} leverages the capabilities acquired in the initial training stage to directly learn from test documents, needing only a review of QA ability. This makes it significantly more training-efficient over time.


\section{Fine-grained Comparison}
\label{sec:fine_grained_compare} 
\begin{table}[!t]
\setlength\tabcolsep{2.5pt}
  \centering
  \begin{threeparttable}
  \fontsize{9}{10}
  \selectfont
    \begin{tabular}{lcccccc}
    \toprule
    \multirow{4}{*}{\textbf{Method}}&
    \multicolumn{4}{c}{\textbf{Q\&A Types (\% $\mathtt{EM}$)}}
     \cr\cmidrule(lr){2-5} 
     & \multirow{2}{*}{\makecell[c]{\textbf{Total}}} & \multirow{2}{*}{\makecell[c]{\textbf{Top-5}\\ \textbf{(37\%)}}} &\multirow{2}{*}{\makecell[c]{\textbf{Time-Related}\\ \textbf{(27\%)}}}&\multirow{2}{*}{\makecell[c]{\textbf{Multiple}\\ \textbf{(10\%)}}}\cr \cr
    \midrule
     PIT &7.00 & 10.81& 3.70  & 0\\
     \selfpt{} & \textbf{32.00} & \textbf{37.84} &\textbf{40.74} & \textbf{20.00} \\
    \bottomrule  
    \end{tabular}
  \end{threeparttable}
  
  \caption{Fine-grained evaluation results on the open-ended generation task, using the Wiki-Bio test dataset concerning the fact types of QA pairs.}
  \label{tab:fine_grained_comparsions}
\end{table}

\noindent \textbf{Setup.}  
To fully understand how the ability to systematically acquire knowledge aids in the knowledge extraction task, we conduct fine-grained comparisons of PIT and \selfpt{} on generated answers for 100 randomly sampled questions from the Wiki-bio dataset. This subset includes 56 QA types in total. Furthermore, we categorize the questions based on the fact types they contain: $(\RN{1})$ the top-5 most common (accounting for 37\%), which includes birthdate, affiliation, nationality, profession, and position/sport; $(\RN{2})$ time-related (accounting for 27\%), such as birthdate, event date, and time period; $(\RN{3})$ multiple-facts (accounting for 10\%), which ask about more than one fact, for example, inquiring both birth date and place; and we report the evaluation results separately. We assess the factual accuracy using exact match.

\noindent \textbf{Results.} As shown in Table \ref{tab:fine_grained_comparsions}, we observe that \selfpt{} consistently outperforms PIT in the overall evaluation and the fine-grained evaluations related to different QA types. These findings underscore the importance of equipping the model with the ability to systematically acquire new knowledge. Furthermore, we present a qualitative comparison between the answers generated by PIT and \selfpt{} in Appendix \ref{sec:qua_ana_case}. To gain insights into potential enhancements for \selfpt{}, we also conduct a detailed error analysis on the types of factual errors that remain challenging after implementing \selfpt{} in Appendix \ref{sec:error_analysis}.

\section{Qualitative Analysis}
\label{sec:qua_ana_case} 

\begin{table*}[!t]
\centering
\fontsize{10}{10}\selectfont
\begin{tabular}{p{15.5cm}}
\toprule
\textbf{Case study 1: Questions requesting information on multiple facts.} \\
\midrule
\textbf{Document:} <Helmut Moritz - Wikipedia> Helmut Moritz (1 November 1933 - 21 October 2022) was an Austrian physical geodesist. He was a member of the Austrian Academy of Sciences and of many other international academies and societies. He became internationally known with a fundamental work on Error propagation in Geodesy. From 1991 to 1995, he was president of the International Union of Geodesy and Geophysics (IUGG). \\
\textbf{Question:} When was Helmut Moritz born and when did he pass away?\\ 
\textbf{Gold Answer:} Born on November 1, 1933, passed away on October 21, 2022.\\ 
\textcolor{gray}{\textit{Model Answers}}\cr 
\hdashline
\noalign{\vskip 0.07cm} 
\textbf{PIT's Answer:} \colorbox{red!15}{Information not provided.}\\
\textbf{\selfpt{}'s Answer:} \colorbox{myblue1}{Born on november 1, 1933, passed away on october 21, 2022.}\\
\midrule
\textbf{Case study 2: Questions inquiring about time-related details.} \\
\midrule
\textbf{Document:} <Brad Smiley - Wikipedia> Brad Smiley (born June 19, 1973) is an American college football coach. He is the head football coach for Southern Arkansas University; a position he has held since 2022. He also was the head coach for Trinity Valley Community College from 2007 to 2017. He also coached for Baylor, Northwestern State, and Tulane.\\
\textbf{Question:} Since when has Brad Smiley been the head football coach for Southern Arkansas University?\\ 
\textbf{Gold Answer:} Since 2022.\\ 
\textcolor{gray}{\textit{Model Answers}}\cr 
\hdashline
\noalign{\vskip 0.07cm} 
\textbf{PIT's Answer:} \colorbox{red!15}{Since 2016.}\\
\textbf{\selfpt{}'s Answer:} \colorbox{myblue1}{Since 2022.}\\
\midrule
\textbf{Case study 3: Questions inquiring about facts encoded in the end of the document, \ie ``positional bias''.} \\
\midrule
\textbf{Document:} <Nathan Saliba  - Wikipedia> Nathan-Dylan Saliba (born February 7, 2004) is a Canadian professional soccer player who plays for Major League Soccer club CF Montréal.\\\
\textbf{Question:} Which Major League Soccer club does Nathan Saliba play for?\\ 
\textbf{Gold Answer:} CF Montréal.\\ 
\textcolor{gray}{\textit{Model Answers}}\cr 
\hdashline
\noalign{\vskip 0.07cm} 
\textbf{PIT's Answer:} \colorbox{red!15}{San jose earthquakes.}\\
\textbf{\selfpt{}'s Answer:} \colorbox{myblue1}{CF Montréal.}\\
\bottomrule
\end{tabular}
\caption{Qualitative analyses comparing the answers produced by PIT and \selfpt{} on the open-ended generation task using the Wiki-Newpages-2023-10-Bio test dataset. The false answers and correct answers are highlighted in \colorbox{red!15}{red} and \colorbox{myblue1}{blue}, respectively.}
\label{tab:quali_analysis}
\end{table*}


In Table \ref{tab:quali_analysis}, we provide a qualitative comparison between the answers generated by PIT and \selfpt{} on the Wiki-Bio test set. We observe that \selfpt{} performs better in answering questions that inquire about multiple facts and time-related facts, as indicated in the top part of Table \ref{tab:quali_analysis}. Furthermore, as shown in the lower part, PIT tend to fail to recall and extract facts at the end of the documents, \ie suffering from ``positional bias''. This observation is consistent with the findings in \citet{saito2024answer}. Encouragingly, our proposed \selfpt{} aids in recalling and extracting factual knowledge embedded at the end of the documents. These findings align with the automatic evaluation results, underscoring the effectiveness of \selfpt{} in enhancing the LLM's knowledge acquisition capability, particularly in knowledge extraction.

\section{Ablation Study}
\label{sec:ablation}
\begin{figure}[!t]
\centering
\includegraphics[width=0.95\linewidth]{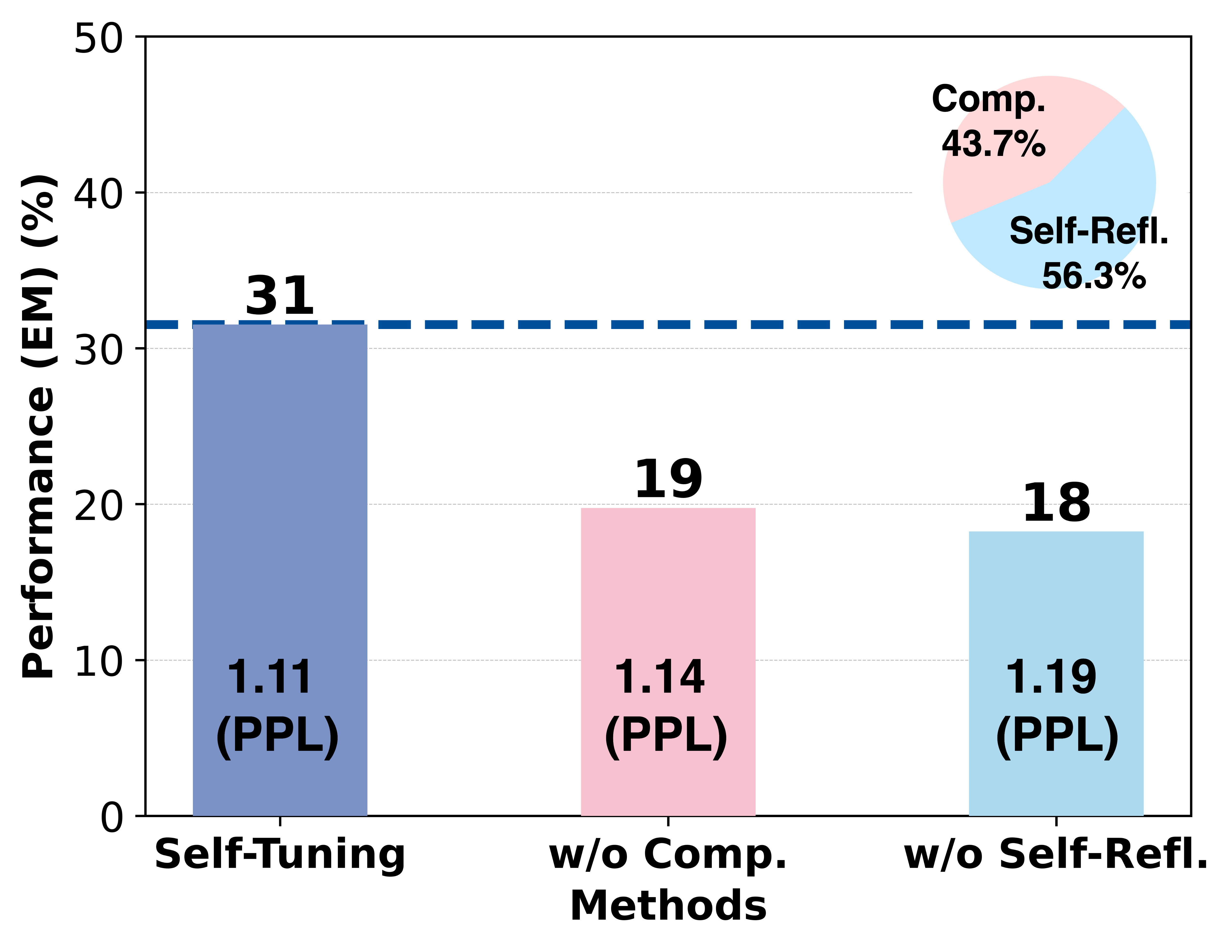}
\caption{Ablation analysis exploring the impact of removing comprehension and self-reflection tasks from the self-teaching tasks for knowledge memorization and acquisition. The proportion of each task type among the self-teaching tasks in the training documents is shown in the upper right corner.}
\label{fig:ablation_study}
\end{figure}



\noindent \textbf{Setup.} We conduct a comprehensive analysis of how comprehension and self-reflection tasks within the self-teaching tasks contribute to enhancing the LLM's knowledge acquisition ability. We focus on two vital aspects: knowledge memorization and extraction. Specifically, we calculate the percentage of the constructed examples for each task type and systematically remove certain tasks to study their impacts.

\noindent \textbf{Results.} In Figure \ref{fig:ablation_study}, we observe the following: $(\RN{1})$ The examples of self-reflection tasks account for a slightly higher ratio than comprehension tasks among the self-teaching tasks. $(\RN{2})$  Both comprehension and self-reflection tasks benefit overall performance on the knowledge acquisition tasks. Notably, removing the examples of self-reflection tasks results in a more significant drop in performance, aligning with its higher percentage over comprehension tasks. These findings confirm the efficacy of the developed \selfteach{} strategy, underscoring the crucial role of comprehension and self-reflection in learning new knowledge for LLMs.

\section{Error Analysis}
\label{sec:error_analysis}
\begin{table*}[ht]
\centering
\fontsize{9}{10}\selectfont
\begin{tabular}{p{1.5cm} p{1.1cm} p{4.5cm} p{2.25cm} p{2cm} p{2cm}}
\toprule
\multirow{2}{*}{\textbf{Type}} & \multirow{2}{*}{\textbf{Fraction}} & \multicolumn{4}{c}{\textbf{Example}} \cr\cmidrule(lr){3-6} && \textbf{Document} & \textbf{Question} & \textbf{Gold Answer} & \textbf{Model Answer} \cr 
\midrule
Wrong answer & 76.47\% & <Jalen Mack  - Wikipedia> Jalen Mack (born August 5, 2005) is an American professional stock car racing driver who competes part-time in the ARCA Menards Series and ARCA Menards Series East, driving the No. 43 Chevrolet for \colorbox{myblue1}{Tamayo Cosentino Racing}. He also competes part time in the ARCA Menards Series West, driving the No. 83 Chevrolet for Mack Motorsports in conjunction with Bill McAnally Racing. & Which team does Jalen Mack drive for in the ARCA Menards Series and ARCA Menards Series East? & Tamayo Cosentino Racing. & \textcolor{red}{Venturini motorsports.} \\
\midrule
Higher granularity & 7.35\% & <Andriyko Olha Fedorivna  - Wikipedia> Andriyko Olha Fedorivna (born January 28, 1945, Voronkiv, Kyiv region) is a \colorbox{myblue1}{Doctor of Law, Professor}, Head of the Department of Constitutional, Administrative and Financial Law of the Kyiv University of Law of the National Academy of Sciences of Ukraine, and Deputy Head of the Department of State and Legal Problems of Management of the V. M. Koretsky Institute of State and Law of the National Academy of Sciences of Ukraine.& What are Andriyko Olha Fedorivna's academic and professional titles? & Doctor of Law, Professor. & Doctor of law, professor, \textcolor{red}{head of the department of constitutional, administrative, and financial law of the kyiv university of law of the national academy of sciences of ukraine.} \\
\midrule
Lower granularity & 5.88\% & <Mike Babcock (American football)  - Wikipedia> Michael Babcock (born February 13, 1979) is an American college football coach. He is the head football coach for McKendree University; a position he has held \colorbox{myblue1}{since 2013}. He also coached for UCLA, Colorado, San Diego, and CSU Pueblo. He played college football for UCLA as a linebacker. & Since when has Mike Babcock (American football) held the head coach position at McKendree University? & Since 2013. & 2013. \\
\midrule
Paraphrase & 10.29\% & 	<Lil Tay  - Wikipedia> Tay Tian (born July 29, 2009), known professionally as Lil Tay, is an \colorbox{myblue1}{American-born Canadian} internet personality and singer. In 2018, she gained prominence online for a period of three months, proclaiming herself to be the ``youngest flexer of the century''. During her brief career, she posted rap videos on YouTube and Instagram which garnered tens of millions of views. Her career ended in mid-2018, after her father applied to the superior court of Canada for full custody and control of her career. According to court documents, he was abusive and largely an absentee. & What is Lil Tay's nationality? & American-born Canadian. & \textcolor{red}{Canadian-American}. \\
\bottomrule

\end{tabular}
\caption{Analysis on the types of factual errors that remain challenging after applying \selfpt{}.}
\label{tab:error_analysis}
\end{table*}

In order to gain insights into potential enhancements for \selfpt{}, we outline four common errors that persist as challenges after implementing \selfpt{}. We offer an in-depth analysis of these errors in Table \ref{tab:error_analysis}, using EM as the evaluation metric.

\section{Evaluation Results on \selfpt{} Variants}
\label{sec:selfpt_variants}

\begin{table*}[!t]
\setlength\tabcolsep{2.5pt}
  \centering
  \begin{threeparttable}
  \fontsize{9}{10}
  \selectfont
    \begin{tabular}{lcccccccccccc}
    \toprule
    \multirow{4}{*}{\textbf{Method}}&
    \multicolumn{7}{c}{\textbf{Wiki-Newpages-2023-10-Bio (Acquisition)}} 
     &\multicolumn{2}{c}{\textbf{NQ (Reten.)}} &\textbf{CSQA (Reten.)}\cr\cmidrule(lr){2-8} \cmidrule(lr){9-10} \cmidrule(lr){11-11}
     & \textbf{Mem.}
      &\multicolumn{5}{c}{\textbf{Extraction}}&\textbf{Reason.} &\multicolumn{2}{c}{\textbf{Extraction}}&\textbf{Reasoning} \cr\cmidrule(lr){2-2} \cmidrule(lr){3-7} \cmidrule(lr){8-8} \cmidrule(lr){9-10} \cmidrule(lr){11-11}
     & $\mathtt{PPL}$ ($\downarrow$)&\%  $\mathtt{Acc.}$ &\%  $\mathtt{EM}$ &\%  $\mathtt{F1}$&\%  $\mathtt{Rec.}$&\%  $\mathtt{Rouge}$ &\% $\mathtt{Acc.}$ &\%  $\mathtt{EM}$ &\%  $\mathtt{F1}$ &\%  $\mathtt{Acc.}$\cr
    \midrule
    Continued Pre-training &7.28&4.68&2.87&14.63&16.98&15.07 &7.96& 16.05&24.67&53.40\\
    \hdashline
    \noalign{\vskip 0.07cm} 
    \selfpt{} w/o Review & 1.26& 28.36 & 23.68& 41.29 &41.93& 41.11 &\textbf{50.40}& \textcolor{red}{15.55}& \textcolor{red}{24.20} & 65.11\\
    \selfpt{} via Read. & 1.46 & 20.97 & 17.65 & 34.54 & 39.19 & 34.55 & 39.37 & \textbf{18.43} & \textbf{27.99} & 62.74 \\
    \selfpt{} w/ Pre-Review & 1.28& 29.86& 25.94 & 43.46& 44.96 & 43.31 &46.91 & 16.28 & 24.80 & 65.11\\
    \selfpt{} & \textbf{1.11}& \textbf{37.25}& \textbf{31.52} & \textbf{50.83}& \textbf{52.62} & \textbf{50.61} &44.31& 16.45 & 25.67 & \textbf{66.01}\\
    \bottomrule  
    \end{tabular}
  \end{threeparttable}
  \caption{Five-shot evaluation results of the \selfpt{} variants on \llamaa{} in the single-domain scenario. Results that fall below the baseline closed-book performance (previously shown in Table \ref{tab:hard-bio-domain}) are highlighted in \textcolor{red}{red}.}
  \label{tab:variants-bio-domain_full}
  \vspace{-3.5mm}
\end{table*}

\noindent \textbf{Setup.} To further investigate the effectiveness of \selfpt{}, we present three variants, as depicted in Table \ref{tab:method_traing_stages_full}: (1) \textbf{\selfpt{} w/o Review}, where we continue training on test documents without the reviewing capability; (2) \textbf{\selfpt{} via Read.}, which displays the training documents in a reading-comprehension format \cite{cheng2024adapting} (an example is shown in Table \ref{tab:train_doc_read_compre_format}); (3) \textbf{\selfpt{} w/ Pre-Review}, which trains on a combination of training documents and training QA in the second stage, before training on test documents.

\noindent \textbf{Results.} In Table \ref{tab:variants-bio-domain_full}, despite having lower performance than \selfpt{}, all variations significantly enhance the model's ability for knowledge acquisition compared to continued pre-training, which further validates the effectiveness of \selfpt{} in improving knowledge acquisition. 

\noindent \textbf{Reviewing QA ability aids in both knowledge acquisition and retention.} Compared to \selfpt{}, \selfpt{} w/o Review also displays inferior performance on the knowledge retention task.

\section{Evaluation Results on \llamaathirteen{} in the Single-domain Scenario} 
\label{sec:evaluate_13b} 

\begin{table*}[!t]
\setlength\tabcolsep{2pt}
  \centering
  \begin{threeparttable}
  \fontsize{9}{10}
  \selectfont
    \begin{tabular}{lcccccccccccc}
    \toprule
    \multirow{4}{*}{\textbf{Method}}&
    \multicolumn{7}{c}{\textbf{Wiki-Newpages-2023-10-Bio (Acquisition)}} 
     &\multicolumn{2}{c}{\textbf{NQ (Reten.)}} &\textbf{CSQA (Reten.)}\cr\cmidrule(lr){2-8} \cmidrule(lr){9-10} \cmidrule(lr){11-11}
     & \textbf{Memorization}
      &\multicolumn{5}{c}{\textbf{Extraction}}&\textbf{Reason.} &\multicolumn{2}{c}{\textbf{Extraction}}&\textbf{Reasoning} \cr\cmidrule(lr){2-2} \cmidrule(lr){3-7} \cmidrule(lr){8-8} \cmidrule(lr){9-10} \cmidrule(lr){11-11}
     & $\mathtt{PPL}$ ($\downarrow$)&\%  $\mathtt{Acc.}$ &\%  $\mathtt{EM}$ &\%  $\mathtt{F1}$&\%  $\mathtt{Rec.}$&\%  $\mathtt{Rouge}$ &\% $\mathtt{Acc.}$ &\%  $\mathtt{EM}$ &\%  $\mathtt{F1}$ &\%  $\mathtt{Acc.}$\cr
    
    
    
      
         \midrule
         \multicolumn{11}{c}{\textbf{\textit{\llamaathirteen{}}}}\cr
    \midrule
    \multicolumn{8}{l}{\textcolor{lightgray!99}{\textit{w/o Knowledge Injection}}}\cr
    Open-book w/ test doc &8.27& 58.97 & 37.41 & 70.38 & 78.64 & 68.09 & 3.57 & -& - & -\\
    Closed-book &8.27& 6.33 & 4.68 & 17.45 & 19.37 & 17.58 & 3.57 & 19.84 & 28.71 & 66.34\\
    \midrule
   \multicolumn{8}{l}{\textcolor{lightgray!99}{\textit{w/ Knowledge Injection}}}\cr 
    Con. Pre-training &  6.35 & \textcolor{red}{4.98} & \textcolor{red}{3.77} & \textcolor{red}{17.12} & \textcolor{red}{18.95} & \textcolor{red}{17.04} & 5.49 & \textbf{21.25} & 30.35&66.34 \\
    
    Standard Ins.-tuning & 3.00& 12.67 & 10.11 & 26.79 & 27.42 & 27.00 & 52.43 & 19.95 & 30.95 & \textcolor{red}{65.77}  \\ 
    
     
     PIT & 1.70 & 22.93 & 19.61 & 36.50 & 36.99 & 36.25 & \textbf{59.40} & \textcolor{red}{19.05} & 31.02 & 70.93\\
    \selfpt{} & \textbf{1.09} & \textbf{44.19} & \textbf{39.37} & \textbf{58.31} & \textbf{60.47} & \textbf{57.90} & 54.18 & 20.69 & \textbf{31.62} & \textbf{71.50} \\
    
    \bottomrule  
    \end{tabular}
  \end{threeparttable}
  \caption{Five-shot evaluation results on \llamaathirteen{} for knowledge acquisition and retention in the single-domain scenario. Results that are inferior to closed-book performance without knowledge injection are indicated in \textcolor{red}{red}.}
  \label{tab:bio-domain-13b}
  \vspace{-1mm}
\end{table*}

Table \ref{tab:bio-domain-13b} presents the evaluation results on \llamaathirteen{} concerning knowledge acquisition and retention in the single-domain scenario using the Wiki-Bio dataset. We make the following observations:

\noindent\textbf{\selfpt{} consistently demonstrates superior performance in enhancing the model's knowledge acquisition and retention abilities as the model size scales.} As the model size scales, \selfpt{} continues to achieve the highest performance across all evaluation metrics on memorization and acquisition tasks, consistently outperforming the compared methods by a significant margin (\eg improving EM score by 20\% on the extraction task). On the reasoning task, \selfpt{} consistently attains high accuracy. Additionally, \selfpt{} consistently exhibits strong performance on knowledge retention tasks. These findings confirm the effectiveness of \selfpt{}, suggesting the potential and robustness of \selfpt{} for applications on larger-scale models.

\noindent\textbf{Continued pre-training for knowledge acquisition proves challenging across all three dimensions.} We find that continuing pre-training on new documents may result in a decline in knowledge extraction performance on \llamaathirteen{}, compared to the baseline performance. This could be due to the fact that merely continuing pre-training might adversely affect its question-answering capability, even when equipped with new knowledge, as demonstrated by the lowered PPL. This observation is consistent with the findings in \citet{cheng2024adapting}. Moreover, the marginal improvements in memorization (reducing PPL by 2\%) and reasoning (increasing accuracy by 2\%) suggest that such a naive approach fails to help the model memorize and capitalize on new knowledge. This highlights the importance of evaluating the model's knowledge acquisition ability comprehensively across multiple dimensions.

\section{Evaluation Results on \llamaachat{} in the Single-domain Scenario} \label{sec:evaluate_chat} 

\begin{table*}[!t]
\setlength\tabcolsep{2pt}
  \centering
  \begin{threeparttable}
  \fontsize{9}{10}
  \selectfont
    \begin{tabular}{lcccccccccccc}
    \toprule
    \multirow{4}{*}{\textbf{Method}}&
    \multicolumn{7}{c}{\textbf{Wiki-Newpages-2023-10-Bio (Acquisition)}} 
     &\multicolumn{2}{c}{\textbf{NQ (Reten.)}} &\textbf{CSQA (Reten.)}\cr\cmidrule(lr){2-8} \cmidrule(lr){9-10} \cmidrule(lr){11-11}
     & \textbf{Memorization}
      &\multicolumn{5}{c}{\textbf{Extraction}}&\textbf{Reason.} &\multicolumn{2}{c}{\textbf{Extraction}}&\textbf{Reasoning} \cr\cmidrule(lr){2-2} \cmidrule(lr){3-7} \cmidrule(lr){8-8} \cmidrule(lr){9-10} \cmidrule(lr){11-11}
     & $\mathtt{PPL}$ ($\downarrow$)&\%  $\mathtt{Acc.}$ &\%  $\mathtt{EM}$ &\%  $\mathtt{F1}$&\%  $\mathtt{Rec.}$&\%  $\mathtt{Rouge}$ &\% $\mathtt{Acc.}$ &\%  $\mathtt{EM}$ &\%  $\mathtt{F1}$ &\%  $\mathtt{Acc.}$\cr
    
    
     
     
    
      
    \midrule
         \multicolumn{11}{c}{\textbf{\textit{\llamaachat{}}}}\cr
    \midrule
    \multicolumn{8}{l}{\textcolor{lightgray!99}{\textit{w/o Knowledge Injection}}}\cr
    Open-book w/ test doc &12.36&71.34& 43.74&75.11&88.38&73.74& 31.14&-&-&-\\
    Closed-book &12.36&5.58&4.07&16.05&17.63&16.19 &31.14& 18.20&26.84&67.16\\
    \midrule
   \multicolumn{8}{l}{\textcolor{lightgray!99}{\textit{w/ Knowledge Injection}}}\cr 
    Con. Pre-training & 8.12& 5.73&\textcolor{red}{3.32}&\textcolor{red}{15.89}&18.60&\textcolor{red}{15.81}&\textcolor{red}{24.83} &\textbf{18.32} & \textbf{27.01}&\textcolor{red}{65.19}\\
    
    Standard Ins.-tuning & 2.99& 12.67 & 10.56& 25.13& 25.41& 25.38&67.76& \textcolor{red}{14.81}& \textcolor{red}{23.72} & \textcolor{red}{58.07}  \\ 
    
     
     PIT & 1.85& 15.54 & 13.12& 29.03 & 29.47& 29.45 &39.51& \textcolor{red}{14.92}& \textcolor{red}{23.38} & \textcolor{red}{62.33}\\
     
    \selfpt{} & \textbf{1.10} & \textbf{33.03}& \textbf{29.41} & \textbf{46.94}& \textbf{47.90} & \textbf{47.00} &\textbf{72.29}& \textcolor{red}{13.57} & \textcolor{red}{22.28} & \textcolor{red}{64.21}\\
    

    \bottomrule  
    \end{tabular}
  \end{threeparttable}
  \caption{Five-shot evaluation results on \llamaachat{} for knowledge acquisition and retention in the single-domain scenario. Results that are inferior to closed-book performance without knowledge injection are indicated in \textcolor{red}{red}.}
  \label{tab:hard-bio-chat}
  \vspace{-1mm}
\end{table*}
 
In this section, we showcase the evaluation outcomes for \llamaachat{} in Table \ref{tab:hard-bio-chat}. We find that even after extensive instruction-following training \cite{ouyang2022training}, \llamaachat{} faces difficulty in extracting newly acquired knowledge after simply continuing pre-training on test documents. Almost all high-performing approaches struggle with knowledge retention, indicating that to incorporate new knowledge, it is preferable to train a base model rather than the version fine-tuned via RLHF (reinforcement learning from human feedback) \cite{ouyang2022training}, despite its remarkable instruction-following capability. More significantly, \selfpt{} consistently surpasses all other compared methods by a considerable margin on knowledge acquisition tasks. These promising outcomes further validate the effectiveness of \selfpt{}. The results imply a potential foundation for exploring the domain of enhancing knowledge acquisition for various models.

\section{Training Efficiency Analysis}
\label{sec:training_efficiency}

To ensure a fair comparison, all methods for knowledge injection presented in Table~\ref{tab:hard-bio-domain} were trained on raw test documents for 3 epochs, as detailed in Appendix~\ref{sec:setup_details}. Additionally, we conducted a detailed analysis of training efficiency on the Wiki-Newpages-2023-Bio dataset using 8 Tesla V100 GPUs (32G) with \llamaa{}:
\begin{itemize}
    \item Continued pre-training: 112.91 seconds
    \item Standard instruction-tuning: 1661.06 seconds
    \item PIT: 6205.52 seconds
    \item \selfpt{}: 5220.50 seconds
\end{itemize}

Our \selfpt{} significantly outperforms the most competitive baseline, PIT, on the knowledge acquisition task and is more time-efficient. 

\section{Evaluation Results on Gemma-7B in the Single-Domain Scenario} \label{sec:evaluate_diverse_models} 

\begin{table*}[!t]
\setlength\tabcolsep{4pt}
  \centering
  \begin{threeparttable}
  \fontsize{9}{10}
  \selectfont
    \begin{tabular}{lcccccccccccc}
    \toprule
    \multirow{2}{*}{\textbf{Method}} &
    \multicolumn{6}{c}{\textbf{Knowledge Acquisition}}  \cr\cmidrule(lr){2-7} 
     & $\mathtt{PPL}$ ($\downarrow$) & \% $\mathtt{Acc.}$ & \% $\mathtt{EM}$ & \% $\mathtt{F1}$ & \% $\mathtt{Recall}$ & \% $\mathtt{Rouge}$ \cr
    \midrule
    \multicolumn{7}{l}{\textcolor{lightgray!99}{\textit{w/o Knowledge Injection}}}\cr
    Closed-book & 12.41 & 7.09 & 4.68 & 17.60 & 18.10 & 17.65 \\
    \midrule
    \multicolumn{7}{l}{\textcolor{lightgray!99}{\textit{w/ Knowledge Injection}}}\cr
    Continued Pre-training & 3.99 & 8.14 & 6.33 & 19.91 & 20.97 & 19.82 \\
    Standard Instruction-tuning & 10.13 & 10.41 & 8.60 & 24.06 & 23.86 & 24.16 \\
    PIT & 4.19 & 8.14 & 5.88 & 20.87 & 20.78 & 20.68 \\
    Self-Tuning & \textbf{1.09} & \textbf{41.93} & \textbf{36.80} & \textbf{56.95} & \textbf{57.41} & \textbf{56.30} \\

    \bottomrule  
    \end{tabular}
  \end{threeparttable}
  \caption{Evaluation results of different methods applied to Gemma-7B on the Wiki-Newpages-2023-Bio dataset.}
  \label{tab:eval_new_model_full}
\end{table*}

We present the evaluation results for Gemma-7B in Table~\ref{tab:eval_new_model_full}. Our \selfpt{} method consistently achieves the best performance, significantly outperforming the baseline methods by a substantial margin. This observation aligns with the results reported across all other evaluation scenarios in Section~\ref{sec:experiments}.

\section{Evaluation Results with Continual Learning Techniques}
\label{sec:evaluate_continual_learning}

\begin{table*}[!t]
\setlength\tabcolsep{2.5pt}
  \centering
  \begin{threeparttable}
  \fontsize{9}{10}
  \selectfont
    \begin{tabular}{lcccccccccccc}
    \toprule
    \multirow{4}{*}{\textbf{Method}}&
    \multicolumn{7}{c}{\textbf{Wiki-Newpages-2023-QA (Acquisition)}} 
     &\multicolumn{2}{c}{\textbf{NQ (Reten.)}} &\textbf{CSQA (Reten.)}\cr\cmidrule(lr){2-8} \cmidrule(lr){9-10} \cmidrule(lr){11-11}
     & \textbf{Memorization}
      &\multicolumn{5}{c}{\textbf{Extraction}}&\textbf{Reason.} &\multicolumn{2}{c}{\textbf{Extraction}}&\textbf{Reasoning} \cr\cmidrule(lr){2-2} \cmidrule(lr){3-7} \cmidrule(lr){8-8} \cmidrule(lr){9-10} \cmidrule(lr){11-11}
     & $\mathtt{PPL}$ ($\downarrow$)&\%  $\mathtt{Acc.}$ &\%  $\mathtt{EM}$ &\%  $\mathtt{F1}$&\%  $\mathtt{Rec.}$&\%  $\mathtt{Rouge}$ &\% $\mathtt{Acc.}$ &\%  $\mathtt{EM}$ &\%  $\mathtt{F1}$ &\%  $\mathtt{Acc.}$\cr
    \midrule
    \multicolumn{11}{c}{\textbf{Knowledge Acquisition on Wiki-Newpages-2023-10-Bio (Single-Domain Scenario)}}\cr
    \midrule
    \multicolumn{8}{l}{\textcolor{lightgray!99}{\textit{w/o Knowledge Injection}}}\cr
    Open-book w/ test doc &8.41&55.20& 31.83&64.48&75.55&62.10& 7.96&-&-&-\\
    Closed-book &8.41&4.68&2.87&14.63&16.98&15.07 &7.96& 16.05&24.67&53.40\\
    \midrule
   \multicolumn{8}{l}{\textcolor{lightgray!99}{\textit{w/ Knowledge Injection}}}\cr 
 PIT & 2.08& 14.03 & 11.61& 27.15 & 28.86& 27.11 &11.93& 15.72& 26.31 & 57.58\\ 
    \selfpt{} & 1.11& 37.25& 31.52 & 50.83& 52.62 & 50.61 &44.31& 16.45 & 25.67 & 66.01\\
    \selfpt{}+Replay &\textbf{1.03} & \textbf{44.49}& \textbf{39.82} & \textbf{58.44}& \textbf{60.58} & \textbf{58.00}  & \textbf{56.24} & \textbf{22.67} & \textbf{33.86} & \textbf{73.55}\\
    \bottomrule  
    \end{tabular}
  \end{threeparttable}
  \caption{Five-shot evaluation results of \llamaa{} combined with continual learning techniques for knowledge acquisition and retention in the single-domain scenario.}
  \label{tab:hard-bio-domain_continual}
\end{table*}

This section explores the potential of integrating \selfpt{} with continual learning techniques. Table~\ref{tab:hard-bio-domain_continual} presents the evaluation results of combining \selfpt{} with a representative continual learning strategy: the replay-based method. In this approach, 500 training QA pairs were randomly sampled from the Wiki QA datasets, curated prior to the pre-training cutoff date of \llamaa{}~\cite{zhang-etal-2024-self}. These QA pairs were included throughout the training process to reinforce general domain knowledge.

The evaluation results on MCQA and NQ confirm that \selfpt{} effectively preserves previously acquired knowledge. Moreover, integrating \selfpt{} with replay-based continual learning (\selfpt{}+Replay) further enhances model performance, demonstrating the following benefits:
\begin{enumerate}
    \item \textbf{Effective knowledge acquisition:} The model successfully learns new knowledge from previously unseen raw documents. This aligns with findings in Appendix~\ref{sec:selfpt_variants}, which highlight that the reviewing QA ability facilitates knowledge acquisition.
    \item \textbf{Efficient knowledge retention:} The replay-based approach ensures that previously learned knowledge is preserved, mitigating catastrophic forgetting during the learning of new tasks.
\end{enumerate}

These findings underscore the significant potential of integrating \selfpt{} with continual learning techniques. By combining \selfpt{}'s strengths with replay-based strategies, the model not only excels in acquiring new knowledge but also maintains strong retention of existing information, making this approach an effective solution for long-term knowledge management.

\section{Evaluation Results Comparing with a Baseline Utilizing Document-Based QA Generation on Test Corpora}
\label{sec:evaluate_compare_testqa_based}

\begin{table*}[!t]
\setlength\tabcolsep{4pt}
  \centering
  \begin{threeparttable}
  \fontsize{9}{10}
  \selectfont
    \begin{tabular}{lcccccccccccc}
    \toprule
    \multirow{2}{*}{\textbf{Method}} &
    \multicolumn{6}{c}{\textbf{Knowledge Acquisition (Wiki-Newpages-2023-Bio)}}  \cr\cmidrule(lr){2-7} 
     & $\mathtt{PPL}$ ($\downarrow$) & \% $\mathtt{Acc.}$ & \% $\mathtt{EM}$ & \% $\mathtt{F1}$ & \% $\mathtt{Recall}$ & \% $\mathtt{Rouge}$ \cr
    \midrule
    \multicolumn{7}{l}{\textcolor{lightgray!99}{\textit{w/o Knowledge Injection}}}\cr
    Open-book w/ test doc & 8.41 & 55.20 & 31.83 & 64.48 & 75.55 & 62.10 \\
    Closed-book & 8.41 & 4.68 & 2.87 & 14.63 & 16.98 & 15.07 \\
    \midrule
    \multicolumn{7}{l}{\textcolor{lightgray!99}{\textit{w/ Knowledge Injection}}}\cr
    Continued Pre-training & 7.28 & 6.33 & 3.62 & 15.96 & 18.72 & 16.11 \\
    Training on test doc w/ QA pairs & \textbf{1.08} & 15.84 & 12.07 & 28.58 & 31.06 & 28.07 \\
    \selfpt{} & 1.11 & \textbf{37.25} & \textbf{31.52} & \textbf{50.83} & \textbf{52.62} & \textbf{50.61} \\
    \bottomrule  
    \end{tabular}
  \end{threeparttable}
  \caption{Evaluation results comparing \selfpt{} to training on test documents with constructed QA pairs using \llamaa{} for knowledge acquisition on the Wiki-Newpages-2023-Bio dataset.}
  \label{tab:testqa_baseline}
\end{table*}

Constructing QA data for every newly introduced raw corpus is infeasible in real-world scenarios due to cost and scalability constraints. Our goal is to enable the model to autonomously acquire new knowledge from previously unseen raw test documents. Therefore, approaches that rely on constructing and training on QA pairs specifically tailored to test documents, as explored in prior studies~\cite{mecklenburg2024injecting, jiang2024mix}, are not aligned with the objectives of this work.

Nevertheless, for completeness and to provide additional context, we include the results of a baseline that simultaneously trains on test documents and QA pairs generated using our proposed \selfteach{} strategy.

As presented in Table~\ref{tab:testqa_baseline}, \selfpt{} consistently demonstrates superior performance. This comparison highlights the significant advantages of our approach in fostering autonomous knowledge acquisition without depending on pre-constructed QA pairs tailored to the test data.

\section{In-depth Sample Documents and Corresponding QA Pairs for Open-Ended Generation and Natural Language Inference Tasks} 
\label{sec:detailed_qa_example}
We present detailed sample documents along with their corresponding QA pairs for open-ended generation and natural language inference tasks in Table \ref{tab:open_end_example} and Table \ref{tab:nli_example}, respectively.
\begin{table*}[!t]
\centering
\fontsize{10}{10}\selectfont
\begin{tabular}{p{15.5cm}}
\toprule
\textbf{Sample document and associated QA pairs for open-ended generation tasks} \\
\midrule
\textbf{\textit{Dataset: Wiki-Newpages-2023-10-Bio}} \\
\midrule
\textbf{Document:} <Helmut Moritz - Wikipedia> Helmut Moritz (1 November 1933 - 21 October 2022) was an Austrian physical geodesist. He was a member of the Austrian Academy of Sciences and of many other international academies and societies. He became internationally known with a fundamental work on Error propagation in Geodesy. From 1991 to 1995, he was president of the International Union of Geodesy and Geophysics (IUGG). \\
\midrule
\textbf{Question:} When was Helmut Moritz born and when did he pass away?\\ 
\textbf{Answer:} Born on November 1, 1933, passed away on October 21, 2022.\\ 
\textbf{Question:} What was Helmut Moritz's profession?\\ 
\textbf{Answer:} Austrian physical geodesist.\\ 
\textbf{Question:} Which academies and societies was Helmut Moritz a member of?\\ 
\textbf{Answer:} Austrian Academy of Sciences, many other international academies, and societies.\\ 
\textbf{Question:} What work made Helmut Moritz internationally known?\\ 
\textbf{Answer:} A fundamental work on Error propagation in Geodesy.\\ 
\textbf{Question:} What position did Helmut Moritz hold from 1991 to 1995?\\ 
\textbf{Answer:} President of the International Union of Geodesy and Geophysics (IUGG).\\ 
\midrule
\textbf{\textit{Dataset: Wiki-Newpages-2023-10-Multi}} \\
\midrule
\textbf{Document:} <2018 California Proposition 71 - Wikipedia> Proposition 71, also known as Prop 71, was a California ballot proposition and proposed state constitution amendment to change the effective date of passed ballot measures from the day after the election to the fifth day after the Secretary of State certified the results.\textbackslash n\textbackslash n Stated goals of the measure was to ensure results were official before new measures were implemented. Opposers fearing a delay in urgent measures. Kevin Mullin supported the amendment. The California Democratic Party endorsed the amendment. Rural County Representatives of California also endorsed the amendment. \\
\midrule
\textbf{Question:} What was the 2018 California Proposition 71, also known as Prop 71?\\ 
\textbf{Answer:} A California ballot proposition, proposed state constitution amendment, change effective date of passed ballot measures.\\ 
\textbf{Question:} What was the proposed change in the effective date of passed ballot measures in the 2018 California Proposition 71?\\ 
\textbf{Answer:} From the day after the election, to the fifth day after the Secretary of State certified the results.\\ 
\textbf{Question:} What were the stated goals of the 2018 California Proposition 71?\\ 
\textbf{Answer:} To ensure results were official before new measures were implemented.\\ 
\textbf{Question:} What concern did opposers of the 2018 California Proposition 71 have?\\ 
\textbf{Answer:} A delay in urgent measures.\\ 
\textbf{Question:} Who supported the 2018 California Proposition 71 amendment?\\ 
\textbf{Answer:} Kevin Mullin.\\ 
\textbf{Question:} Which organizations endorsed the 2018 California Proposition 71 amendment?\\ 
\textbf{Answer:} The California Democratic Party, Rural County Representatives of California.\\ 
\midrule

\textbf{\textit{Dataset: Wiki-Newpages-2023-(9)10-Film}} \\
\midrule
\textbf{Document:} <Krazy House (film) - Wikipedia> Krazy House is an upcoming Dutch comedy film. It is written, directed, and co-produced by Steffen Haars and Flip van der Kuil in their English-language feature debut. Shot on location in Amsterdam, the film stars Nick Frost, Kevin Connolly and Alicia Silverstone. Maarten Swart is producer for Kaap Holland Films. \\
\midrule
\textbf{Question:} What is Krazy House (film)?\\ 
\textbf{Answer:} An upcoming Dutch comedy film.\\ 
\textbf{Question:} Who are the writers, directors, and co-producers of Krazy House (film)?\\ 
\textbf{Answer:} Steffen Haars, Flip van der Kuil.\\ 
\textbf{Question:} What is significant about Steffen Haars and Flip van der Kuil's involvement in Krazy House (film)?\\ 
\textbf{Answer:} It is their English-language feature debut.\\ 
\textbf{Question:} Where was Krazy House (film) shot?\\ 
\textbf{Answer:} On location in Amsterdam.\\ 
\textbf{Question:} Who is the producer of Krazy House (film) and which production company is involved?\\ 
\textbf{Answer:} Maarten Swart, Kaap Holland Films.\\ 
\bottomrule
\end{tabular}
\caption{Sample document and associated QA pairs for open-ended generation tasks in Wiki-Newpages-2023-10-Bio, Wiki-Newpages-2023-10-Multi, and Wiki-Newpages-2023-(9)10-Film datasets.}
\label{tab:open_end_example}
\end{table*}

\begin{table*}[!t]
\centering
\fontsize{10}{10}\selectfont
\begin{tabular}{p{15.5cm}}
\toprule
\textbf{Sample document and associated QA pairs for natural language inference tasks} \\
\midrule
\textbf{\textit{Dataset: Wiki-Newpages-2023-10-Bio}} \\
\midrule
\textbf{Document:} <Sawyer Gipson-Long  - Wikipedia> Alec Sawyer Gipson-Long (born December 12, 1997) is an American professional baseball pitcher for the Detroit Tigers of Major League Baseball (MLB). He made his MLB debut in 2023.\\
\midrule
\textbf{Question:} Based on the paragraph above can we conclude that  <Alec Sawyer Gipson-Long> Sawyer Gipson-Long was born in December 1997. Options: -Yes; -It's impossible to say; -No\\ 
\textbf{Answer:} Yes\\ 
\textbf{Question:} Based on the paragraph above can we conclude that  <Alec Sawyer Gipson-Long> Sawyer Gipson-Long is a professional football player. Options: -Yes; -It's impossible to say; -No\\ 
\textbf{Answer:} No\\ 
\textbf{Question:} Based on the paragraph above can we conclude that  <Alec Sawyer Gipson-Long> Sawyer Gipson-Long plays for the Detroit Tigers in Major League Baseball. Options: -Yes; -It's impossible to say; -No\\ 
\textbf{Answer:} Yes\\ 
\textbf{Question:} Based on the paragraph above can we conclude that  <Alec Sawyer Gipson-Long> Sawyer Gipson-Long made his MLB debut in 2020. Options: -Yes; -It's impossible to say; -No\\ 
\textbf{Answer:} No\\ 
\midrule
\textbf{\textit{Dataset: Wiki-Newpages-2023-10-Multi}} \\
\midrule
\textbf{Document:} <2023 Astana Open \u2013 Singles  - Wikipedia> Novak Djokovic was the reigning champion, but chose not to compete this year.Seeds. \\
\midrule
\textbf{Question:} Based on the paragraph above can we conclude that <2023 Astana Open \u2013 Singles> Novak Djokovic won the previous Astana Open singles tournament.Options: -Yes; -It's impossible to say; -No\\ 
\textbf{Answer:} Yes\\ 
\textbf{Question:} Based on the paragraph above can we conclude that <2023 Astana Open \u2013 Singles> Novak Djokovic is participating in the 2023 Astana Open singles tournament.Options: -Yes; -It's impossible to say; -No\\ 
\textbf{Answer:} No\\ 
\textbf{Question:} Based on the paragraph above can we conclude that <2023 Astana Open \u2013 Singles> The 2023 Astana Open is a tennis tournament.Options: -Yes; -It's impossible to say; -No\\ 
\textbf{Answer:} It's impossible to say\\ 
\textbf{Question:} Based on the paragraph above can we conclude that <2023 Astana Open \u2013 Singles> Novak Djokovic was injured and could not compete in the 2023 Astana Open singles tournament.Options: -Yes; -It's impossible to say; -No\\ 
\textbf{Answer:} It's impossible to say\\ 
\midrule

\textbf{\textit{Dataset: Wiki-Newpages-2023-(9)10-Film}} \\
\midrule
\textbf{Document:} <Unstoppable (2023 film)  - Wikipedia> Unstoppable is a 2023 comedy-drama film directed by Diamond Ratnababu and produced by Rajith Rao under AB2 Productions. The film was released theatrically worldwide on 9 June 2023. \\
\midrule
\textbf{Question:} Based on the paragraph above can we conclude that<Unstoppable (2023 film)> Unstoppable is a film that combines elements of comedy and drama.Options: -Yes; -It's impossible to say; -No\\ 
\textbf{Answer:} Yes\\ 
\textbf{Question:} Based on the paragraph above can we conclude that<Unstoppable (2023 film)> Diamond Ratnababu is the producer of the film Unstoppable.Options: -Yes; -It's impossible to say; -No\\ 
\textbf{Answer:} No\\ 
\textbf{Question:} Based on the paragraph above can we conclude that<Unstoppable (2023 film)> Unstoppable was released in theaters worldwide.Options: -Yes; -It's impossible to say; -No\\ 
\textbf{Answer:} Yes\\ 
\textbf{Question:} Based on the paragraph above can we conclude that<Unstoppable (2023 film)> The film Unstoppable was released before June 2023.Options: -Yes; -It's impossible to say; -No\\ 
\textbf{Answer:} No\\ 
\textbf{Question:} Based on the paragraph above can we conclude that<Unstoppable (2023 film)> The film Unstoppable was distributed by Diamond Ratnababu.Options: -Yes; -It's impossible to say; -No\\ 
\textbf{Answer:} It's impossible to say\\ 
\bottomrule
\end{tabular}
\caption{Sample document and associated QA pairs for natural language inference tasks in Wiki-Newpages-2023-10-Bio, Wiki-Newpages-2023-10-Multi, and Wiki-Newpages-2023-(9)10-Film test datasets.}
\label{tab:nli_example}
\end{table*}

\section{Token Count Distribution for the Open-ended Generation Task Across the Three Datasets} 
\label{sec:token_distribution}
\begin{figure*}[!t]
\centering
\subfigure[Document length.]{
\begin{minipage}[t]{0.3\linewidth}
  \centering
  \includegraphics[width=1.8in]{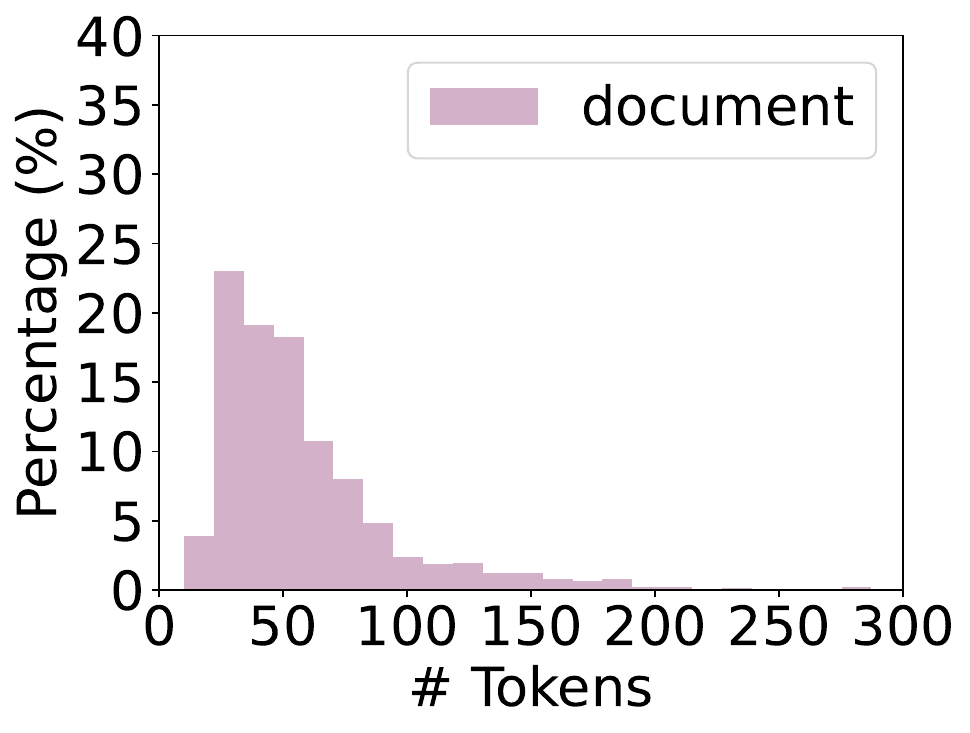}
  \label{fig:bio_doc_hist}
\end{minipage}%
}%
\subfigure[Question length.]{
\begin{minipage}[t]{0.3\linewidth}
  \centering
  \includegraphics[width=1.8in]{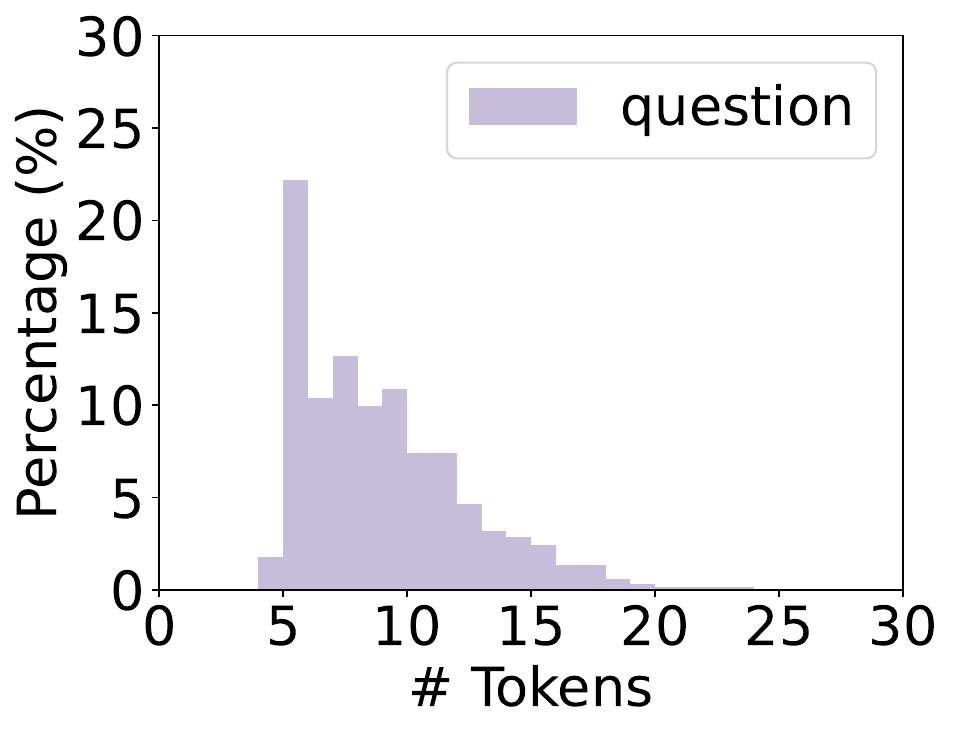}
  \label{fig:bio_q_hist}
\end{minipage}%
}%
\subfigure[Answer length.]{
\begin{minipage}[t]{0.3\linewidth}
  \centering
  \includegraphics[width=1.8in]{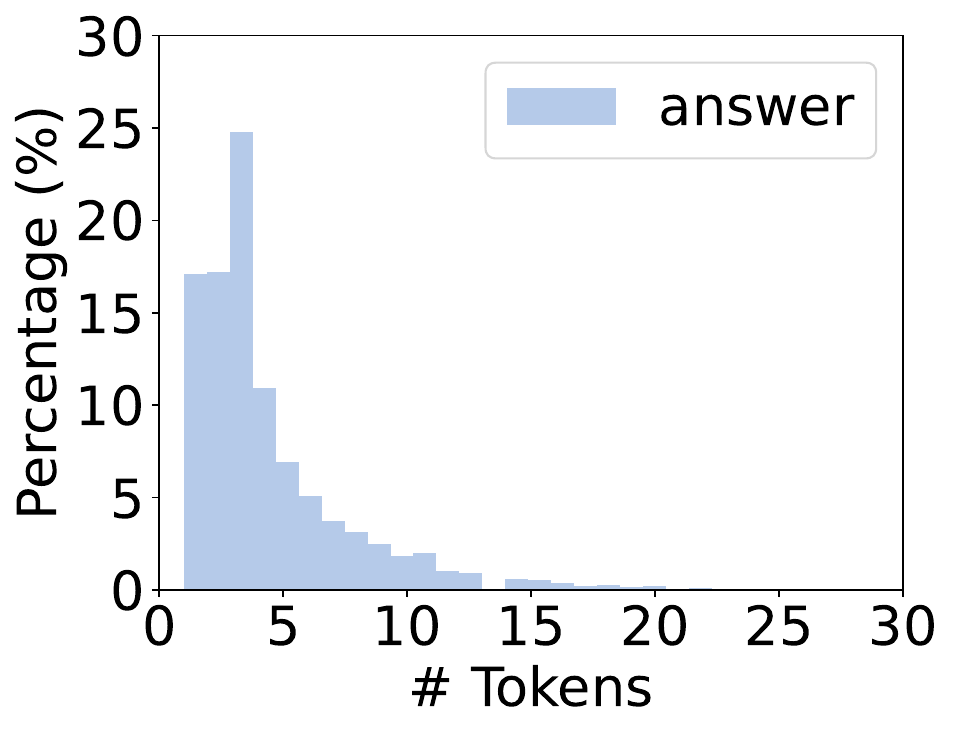}
  \label{fig:bio_ans_hist}
\end{minipage}%
}%
\quad
\caption{Distribution histogram of the token count in a document, a question, and an answer for the open-ended generation task from the Wiki-Newpages-2023-10-Bio dataset, respectively.}
\label{fig:bio_hist}
\end{figure*}

\begin{figure*}[!t]
\centering
\subfigure[Document length.]{
\begin{minipage}[t]{0.3\linewidth}
  \centering
  \includegraphics[width=1.8in]{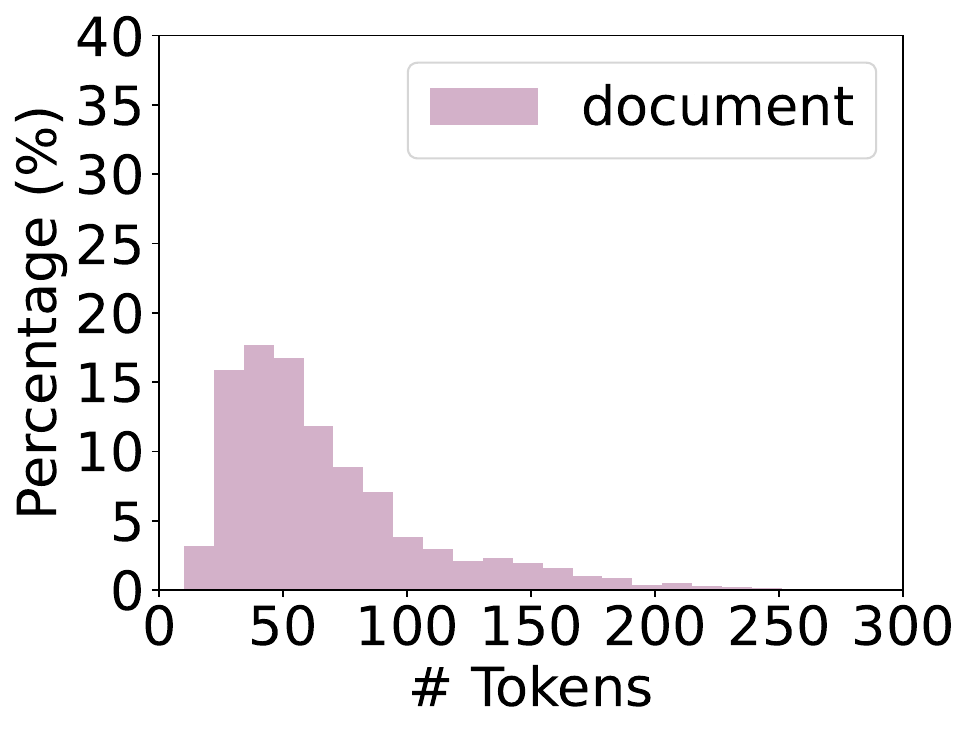}
  \label{fig:multi_doc_hist}
\end{minipage}%
}%
\subfigure[Question length.]{
\begin{minipage}[t]{0.3\linewidth}
  \centering
  \includegraphics[width=1.8in]{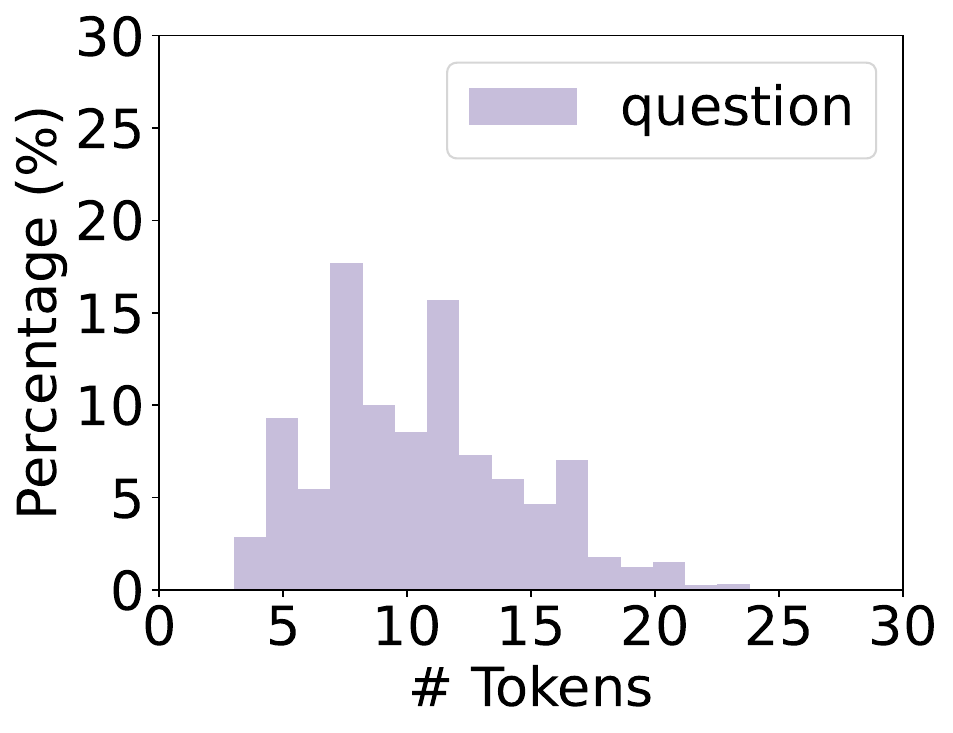}
  \label{fig:multi_q_hist}
\end{minipage}%
}%
\subfigure[Answer length.]{
\begin{minipage}[t]{0.3\linewidth}
  \centering
  \includegraphics[width=1.8in]{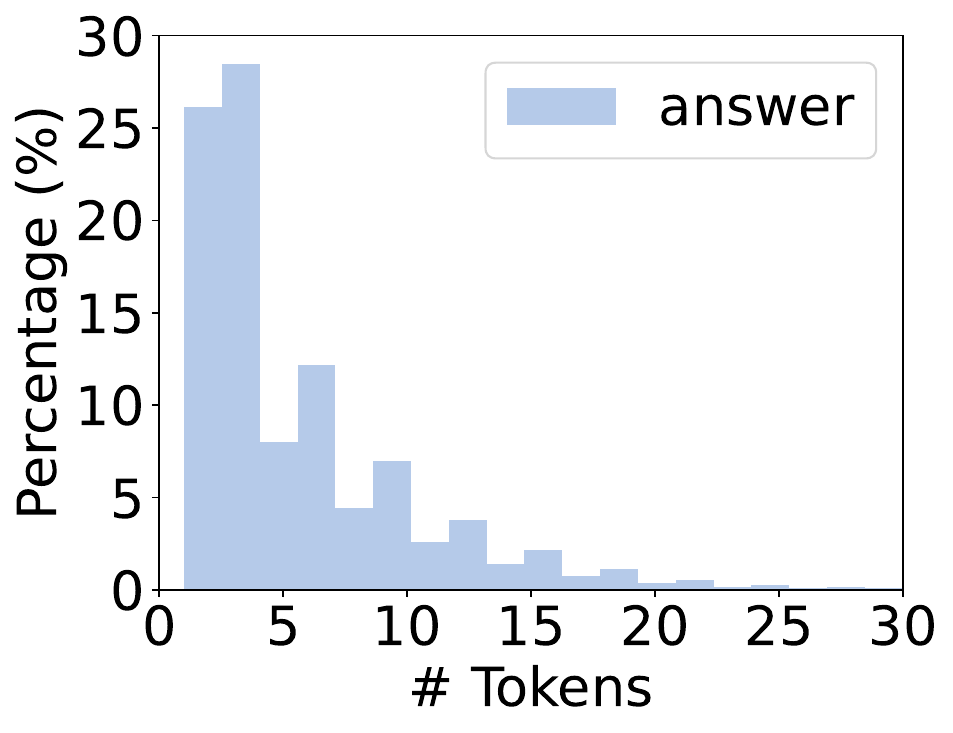}
  \label{fig:multi_ans_hist}
\end{minipage}%
}%
\quad
\caption{Distribution histogram of the token count in a document, a question, and an answer for the open-ended generation task from the Wiki-Newpages-2023-10-Multi dataset, respectively.}
\label{fig:multi_hist}
\end{figure*}

\begin{figure*}[!t]
\centering
\subfigure[Document length.]{
\begin{minipage}[t]{0.3\linewidth}
  \centering
  \includegraphics[width=1.8in]{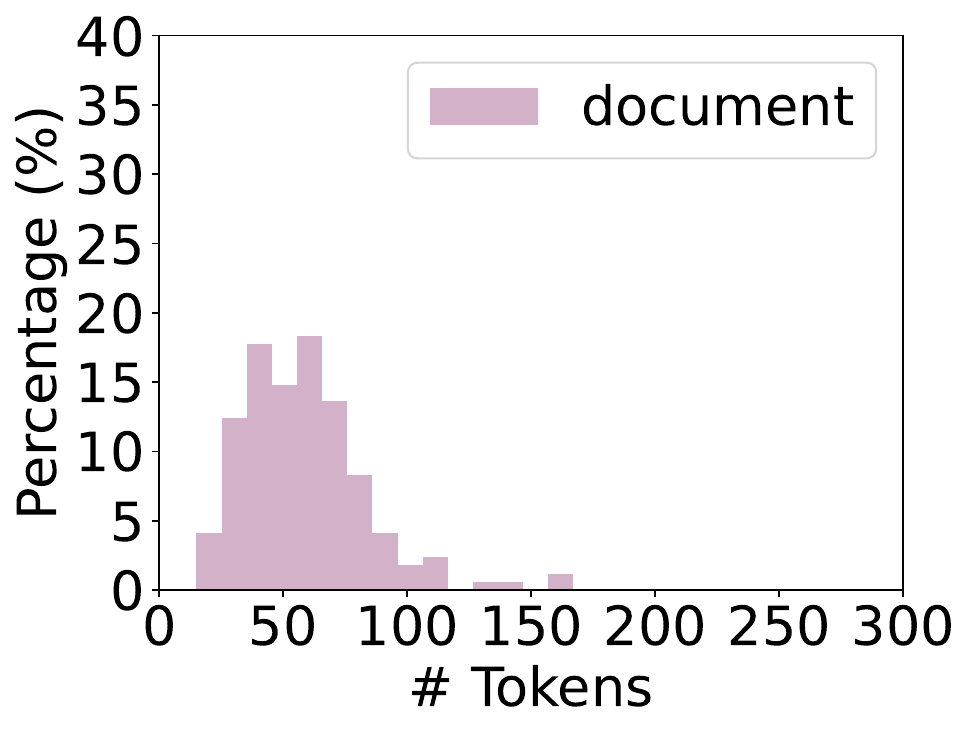}
  \label{fig:film_doc_hist}
\end{minipage}%
}%
\subfigure[Question length.]{
\begin{minipage}[t]{0.3\linewidth}
  \centering
  \includegraphics[width=1.8in]{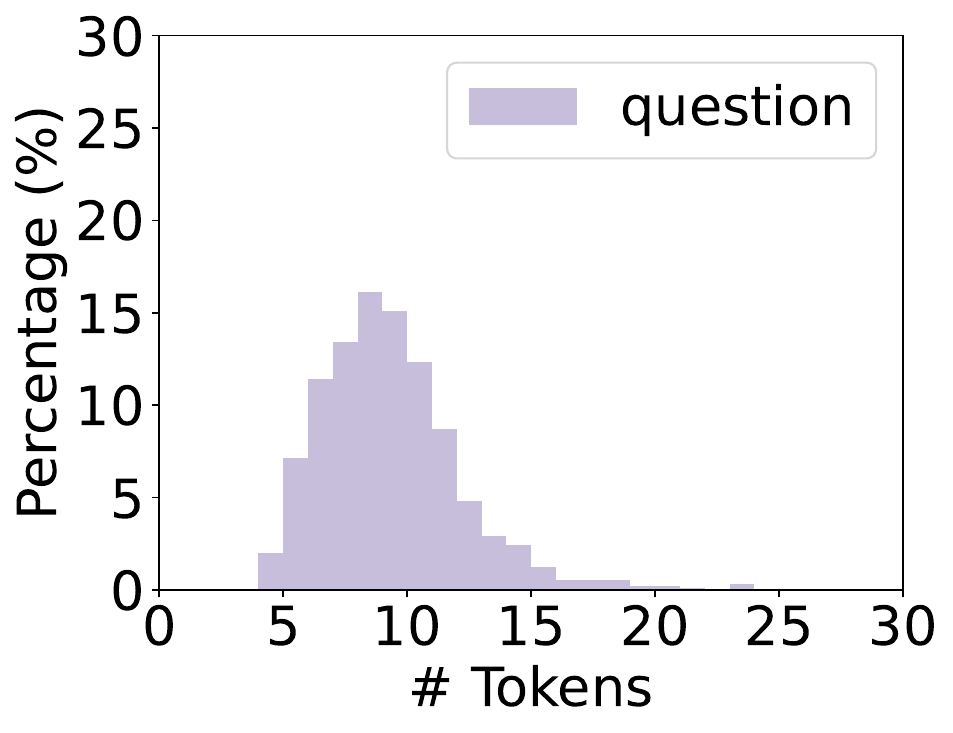}
  \label{fig:film_qa_hist}
\end{minipage}%
}%
\subfigure[Answer length.]{
\begin{minipage}[t]{0.3\linewidth}
  \centering
  \includegraphics[width=1.8in]{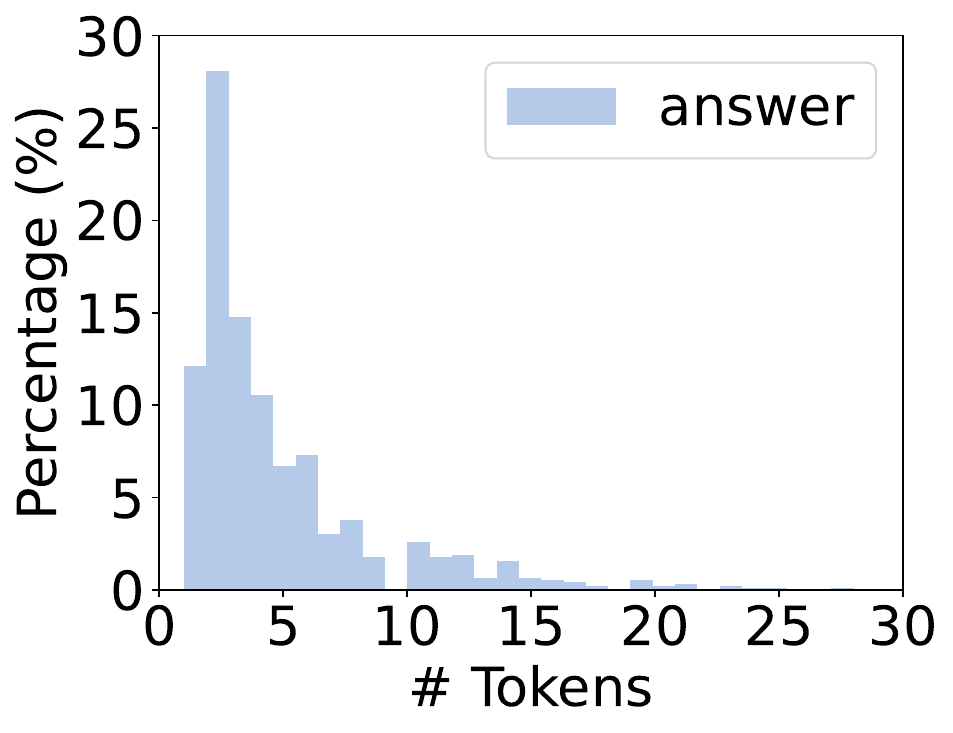}
  \label{fig:film_ans_hist}
\end{minipage}%
}%
\quad
\caption{Distribution histogram of the token count in a document, a question, and an answer for the open-ended generation task from the Wiki-Newpages-2023-(9)10-Film dataset, respectively.}
\label{fig:film_hist}
\end{figure*}

The distribution of token counts for the open-ended generation task across the three datasets is depicted in Figure \ref{fig:bio_hist}, Figure \ref{fig:multi_hist}, and Figure \ref{fig:film_hist}, respectively.

\section{Examination of QA Types in Open-ended Generation QA Datasets} \label{sec:analysis_qa_type} 

\begin{table*}[th]
\setlength\tabcolsep{3pt}
  \centering
  \begin{threeparttable}
  \fontsize{9}{9}
  \selectfont
    \begin{tabular}{lcccccccc} 
    \toprule
        \multirow{3}{*}{\textbf{Dataset}}& \multirow{3}{*}{\makecell[c]{\textbf{QA Type}\\ \textbf{Instances}}}&
     \multicolumn{2}{c}{\textbf{QA Types}}
    &\multicolumn{2}{c}{\textbf{QA Types w/ Multiple Facts}}\cr\cmidrule(lr){3-4} \cmidrule(lr){5-6}
                                              &      & \textbf{Statistics} & \textbf{Top-5 Types} & \textbf{Statistics} & \textbf{Top-5 Types} \cr 
    \midrule
    \multicolumn{6}{c}{\textbf{Wiki-Newpages-2023-10-Bio (Single-domain)}}\cr 
    \midrule
    \multirow{5}{*}{\makecell[c]{Train}} & \multirow{5}{*}{\makecell[c]{Birth Date,\\Achievements,\\ Position, \\ \etc}}&\multirow{5}{*}{\makecell[c]{2014\\ (\# Types);\\ 6073\\ (\# Counts)}} & \multirow{5}{*}{\makecell[l]{ Birth Date (11.24\%)\\ Nationality (5.37\%) \\Profession (5.15\%) \\Team/Affiliation (3.05\%) \\Role/Position (2.56\%)}} & \multirow{5}{*}{\makecell[c]{158 \\(\# Types);\\ 265 \\(\# Counts) }} & \multirow{5}{*}{\makecell[l]{Birth \& Death Dates (0.93\%) \\ Birth Date \& Place (0.44\%) \\ Death Date \& Place (0.12\%) \\ Nationality \& Profession (0.10\%) \\Current Position \& Tenure (0.08\%)}} \cr \\ \cr \cr \cr
        \midrule
    \multirow{5}{*}{\makecell[c]{Test}} & \multirow{5}{*}{\makecell[c]{Full Name,\\Affiliation,\\ Residence, \\ \etc}}&\multirow{5}{*}{\makecell[c]{281\\ (\# Types);\\ 655 \\(\# Counts)}} & \multirow{5}{*}{\makecell[l]{ Birth Date (13.11\%)\\ Profession (6.18\%) \\Nationality (5.62\%) \\Team/Affiliation (4.49\%)\\Role/Position (3.00\%)}} & \multirow{5}{*}{\makecell[c]{16 \\(\# Types);\\ 30\\ (\# Counts) }} & \multirow{5}{*}{\makecell[l]{Birth Date \& Place (1.31\%) \\ Birth \& Death Dates (1.12\%) \\ Death Date \& Place (0.56\%) \\ Car Number \& Manufacturer (0.37\%) \\Current Club \& League (0.19\%)}} \cr \\ \cr \cr \cr
    \hdashline
    \noalign{\vskip 0.07cm} 
   \multicolumn{6}{l}{\textit{Within the train and test sets, there are 63 and 8 answers labeled as ``Information not provided/missing,'' respectively.}}\cr 
    \midrule
    \multicolumn{6}{c}{\textbf{Wiki-Newpages-2023-10-Multi (Multi-domain)}}\cr 
    \midrule
    \multirow{5}{*}{\makecell[c]{Train}} & \multirow{5}{*}{\makecell[c]{Album Source,\\Location,\\ Season \\Number, \etc}}&\multirow{5}{*}{\makecell[c]{4813 \\ (\# Types);\\ 9973 \\(\# Counts)}} & \multirow{5}{*}{\makecell[l]{ Birth Date (3.37\%)\\ Profession (1.76\%) \\ Nationality (1.47\%) \\Location (1.39\%) \\Release Date (1.27\%)}} & \multirow{5}{*}{\makecell[c]{303\\ (\# Types);\\ 371 \\(\# Counts) }} & \multirow{5}{*}{\makecell[l]{Birth \& Death Dates (0.32\%) \\ Birth Date \& Place (0.14\%) \\ Event Date \& Location (0.06\%) \\ Death Date \& Place (0.06\%) \\ Nationality \& Profession (0.05\%)}} \cr \\ \cr \cr \cr
        \midrule
    \multirow{5}{*}{\makecell[c]{Test}} & \multirow{5}{*}{\makecell[c]{Legacy/Impact,\\Purpose,\\ Leadership, \\ \etc}}&\multirow{5}{*}{\makecell[c]{924 \\(\# Types);\\ 1498 \\(\# Counts)}} & \multirow{5}{*}{\makecell[l]{ Birth Date (3.06\%)\\ Release Date (1.80\%) \\Profession (1.57\%) \\Nationality (1.25\%) \\Team/Affiliation (1.02\%)}} & \multirow{5}{*}{\makecell[c]{57\\ (\# Types);\\ 66\\ (\# Counts) }} & \multirow{5}{*}{\makecell[l]{ Birth \& Death Dates (0.31\%) \\ Birth Date \& Place (0.31\%) \\ Death Date \& Place (0.16\%) \\ Job Titles \& Affiliations (0.16\%) \\Language \& Genre (0.16\%)}} \cr \\ \cr \cr \cr
        \hdashline
    \noalign{\vskip 0.07cm} 
   \multicolumn{6}{l}{\textit{Within the train and test sets, there are 31 and 4 answers labeled as ``Information not provided/missing,'' respectively.}}\cr 
    \midrule
    \multicolumn{6}{c}{\textbf{Wiki-Newpages-2023-(9)10-Film (Single-domain)}}\cr 
    \midrule
    \multirow{5}{*}{\makecell[c]{Test}} & \multirow{5}{*}{\makecell[c]{Director,\\Actor,\\ Music \\ Composer, \etc}}&\multirow{5}{*}{\makecell[c]{339 \\ (\# Types);\\ 955 \\(\# Counts)}} & \multirow{5}{*}{\makecell[l]{ Director (9.07\%)\\ Release Date (7.23\%) \\ Genre (6.96\%) \\Cast (3.55\%) \\Language (2.76\%)}} & \multirow{5}{*}{\makecell[c]{13 \\(\# Types);\\ 15 \\(\# Counts) }} & \multirow{5}{*}{\makecell[l]{Title \& Release Year (0.39\%)\\ Milestone \& Historical Comparison (0.13\%)\\ Profession \& Industry (0.13\%)\\ Cast \& Roles (0.13\%)\\ Producer \& Production Banner (0.13\%)}} \cr \\ \cr \cr \cr
    \bottomrule  
    \end{tabular}
  \end{threeparttable}
  
  \caption{A comprehensive analysis of QA types related to factual information in open-ended generation QA datasets from Wiki-Newpages-2023-10-Bio (Wiki-Bio), Wiki-Newpages-2023-10-Multi (Wiki-Multi), and Wiki-Newpages-2023-(9)10-Film (Wiki-Film).}
  \label{tab:wiki_data_qa_details}
  \vspace{-1mm}
\end{table*}
 
We perform a detailed analysis of the QA types associated with the factual information in the open-ended generation QA datasets, as displayed in Table \ref{tab:wiki_data_qa_details}, by using the prompt in Table \ref{tab:wikiqa_type_prompt} with GPT-4.

\section{Detailed Templates used in the \selfteach{} Strategy} \label{sec:example_doc_task} 
\begin{table*}[ht]
\centering
\fontsize{9}{10}\selectfont
\begin{tabular}{p{3.5cm} p{3cm} p{8cm}}
\toprule
\textbf{Type} & \textbf{Task} & \textbf{Template} \\
\midrule
\textbf{\textit{Memorization}} \\
\cmidrule(lr){1-1}
Next-Token Prediction & Text-to-Text & <Document> \\
\midrule
\textbf{\textit{Comprehension}} \\
\cmidrule(lr){1-1}
\colorbox{color1!70}{\ding{172}}Summarization & Text-to-Topic & \makecell[l]{\textbf{Question}: Write a title: <Document>. \\ \textbf{Answer}: <Title>.} \\
\midrule
\colorbox{color2!70}{\ding{173}}Gist Identification & Text-to-Word & \makecell[l]{\textbf{Question}: Highlight the key information within \\the article: <Document>. \\ \textbf{Answer}: <Entity1>, <Entity2>, \etc} \\
\midrule
\colorbox{color3!70}{\ding{174}}Natural Language Inference & Text-to-Option & \makecell[l]{\textbf{Question}: <Document> Based on the article above \\can we conclude that <Sentence>. Options: -Yes; \\-It's impossible to say; -No. \\ \textbf{Answer}: Yes/It's impossible to say/No.} \\
\midrule
\textbf{\textit{Self-Reflection}} \\
\cmidrule(lr){1-1}
\colorbox{color4!55}{\ding{172}}``Teaching'' & Topic-to-Text & \makecell[l]{\textbf{Question}: Tell me about <Title>. \\ \textbf{Answer}: <Document>.} \\
\midrule
\colorbox{color5!55}{\ding{173}}``Flashcards'' & Word-to-Text & \makecell[l]{\textbf{Question}: Generate a concrete description about <Title>. \\based on the following keywords: <Entity>, \etc \\ \textbf{Answer}: <Document>.} \\
\midrule
\colorbox{color6!55}{\ding{174}} Fill-in-the-Blank & Cloze Sentence-to-Entity & \makecell[l]{\textbf{Question}: <Title> <Sentence\_Part1> -- <Sentence\_Part2>\\ (w/o <Entity>). \\ \textbf{Answer}: <Entity>.} \\
\midrule
\colorbox{color7!55}{\ding{175}}Multi-Choice QA & Cloze Sentence (w/ options)-to-Entity & \makecell[l]{\textbf{Question}: <Title> <Sentence\_Part1> -- <Sentence\_Part2>\\ (w/o <Entity>) Options: - <Entity1>; - <Entity2>, \etc  \\ \textbf{Answer}: <Entity>.} \\
\midrule
\colorbox{color8!55}{\ding{176}}Sentence Completion & Text-to-Text & \makecell[l]{\textbf{Question}: <Title> <Sentence\_Part1>: \\ \textbf{Answer}: <Sentence\_Part2>.} \\
\hline
\end{tabular}
\caption{The detailed templates for each task used in the \selfteach{} learning strategy.}
\label{tab:multi_task_list}
\end{table*}


 
\begin{table*}[ht]
\centering
\fontsize{9.5}{12}\selectfont
\begin{tabular}{p{3.5cm} p{11.5cm}}
\toprule
\textbf{Type} & \textbf{Example} \\
\midrule
\textbf{\textit{Memorization}} \\
\cmidrule(lr){1-1}
Next-Token Prediction & <Robert Anderson (artist)  - Wikipedia> Robert Alexander Anderson (born 1946) is an American portrait artist known for painting the official portraits of George W. Bush and Alan Greenspan as well as designing United States postage stamps. \\
\midrule
\textbf{\textit{Comprehension}} \\
\cmidrule(lr){1-1}
\colorbox{color1!70}{\ding{172}}Summarization & \makecell[l]{\textbf{Question}: Write a title: <Robert Anderson (artist) ... stamps. \\ \textbf{Answer}: Robert Anderson (artist).} \\
\midrule
\colorbox{color2!70}{\ding{173}}Gist Identification & \makecell[l]{\textbf{Question}: Highlight the key information within the article: <Robert Anderson \\(artist)  ... stamps. \\ \textbf{Answer}: United States; American; Alan Greenspan; George W. Bush; \\Robert Alexander Anderson; 1946} \\
\midrule
\colorbox{color3!70}{\ding{174}}Natural Language Inference & \makecell[l]{\textbf{Question}: <Robert Anderson (artist) ... stamps. Based on the article above can we\\ conclude that
<Robert Anderson (artist)> Robert Alexander Anderson (born 1946) is \\an American portrait artist known for painting the official portraits of George W. Bush\\ and Alan Greenspan as well as designing United States postage stamps.\\ Options:\\
- Yes\\
- It's impossible to say\\
- No\\
\textbf{Answer}: Yes} \\
\midrule
\textbf{\textit{Self-Reflection}} \\
\cmidrule(lr){1-1}
\colorbox{color4!55}{\ding{172}}``Teaching'' & \makecell[l]{\textbf{Question}: Tell me about Robert Anderson (artist). \\ \textbf{Answer}: Robert Alexander Anderson (born 1946) is ... stamps.} \\
\midrule
\colorbox{color5!55}{\ding{173}}``Flashcards'' & \makecell[l]{\textbf{Question}: Generate a concrete description about Robert Anderson (artist), based on \\the following keywords: United States; American; Alan Greenspan; George W. Bush; \\Robert Alexander Anderson; 1946 \\ \textbf{Answer}:  Robert Alexander Anderson (born 1946) is ... stamps.} \\
\midrule
\colorbox{color6!55}{\ding{174}}Fill-in-the-Blank & \makecell[l]{\textbf{Question}: <Robert Anderson (artist)> Robert Alexander Anderson (born 1946) is \\an American -- known for painting the official portraits of George W. Bush and Alan \\Greenspan as well as designing United States postage stamps. \\ \textbf{Answer}: Portrait artist.} \\
\midrule
\colorbox{color7!55}{\ding{175}}Multi-Choice QA & \makecell[l]{\textbf{Question}:<Robert Anderson (artist)> - (born 1946) is an American portrait artist \\known for painting the official portraits of George W. Bush and Alan Greenspan as well \\as designing United States postage stamps.\\
Options:\\
- Alan Greenspan\\
- 1946\\
- Robert Alexander Anderson\\
- George W. Bush\\
\textbf{Answer}: Robert Alexander Anderson.} \\
\midrule
\colorbox{color8!55}{\ding{176}}Sentence Completion  & \makecell[l]{\textbf{Question}: <Robert Anderson (artist)> Robert Alexander Anderson (born 1946) is \\an American portrait artist known for painting the official portraits of George W. Bush \\and Alan Greenspan as well as: \\ \textbf{Answer}: Designing United States postage stamps.} \\
\hline
\end{tabular}
\caption{An example of a training document from the Wiki-Newpages-2023-10-Bio train set, accompanied by related self-teaching tasks.}
\label{tab:train_doc_multi_task_list}
\end{table*}
 
We provide the detailed templates employed in the \selfteach{} strategy in Table \ref{tab:multi_task_list} and a complete example of a training document accompanied by its associated \selfteach{} tasks in Table \ref{tab:train_doc_multi_task_list}.

\section{Datasets and Evaluation Metrics}
\label{sec:setup_details} 
\paragraph{Evaluation on Knowledge Acquisition.} We assess the effectiveness of \selfpt{} in enhancing the model's knowledge acquisition capabilities on the curated Wiki-Newpages-QA datasets, concentrating on memorization, extraction, and reasoning. $(\RN{1})$ For memorization, we utilize test document datasets and report perplexity \cite{Jelinek1977PerplexityaMO}, which measures how well a language model predicts a text sample. $(\RN{2})$ For extraction, we employ test QA datasets for open-ended generation tasks. To evaluate the factual accuracy of the generated responses, we use exact match (EM), Recall, and F1 over words in the answer(s), following \citet{kwiatkowski-etal-2019-natural}. Additionally, we report Rouge-L \cite{lin-2004-rouge} to measure the overlap of n-grams between the generated and gold answers, accounting for minor lexical variations, following \citet{jiang2024instructiontuned}. We also assess accuracy by comparing each response's factual correctness to the gold answer, using the bidirectional entailment approach with the Deberta-Large-MNLI model \cite{he2021deberta}. We report the five-shot evaluation results on the open-ended generation tasks using the prompt in Table \ref{tab:prompt_five_shot}. $(\RN{3})$ Concerning reasoning, we utilize the test QA datasets for NLI tasks and report the accuracy by comparing the generated option with the gold option using EM. We present the zero-shot evaluation results on NLI tasks.

\paragraph{Evaluation on Knowledge Retention.} It is well-known that knowledge acquisition is often accompanied by catastrophic forgetting \cite{allenzhu2023physics, wang2023survey}. Therefore, we also provide the knowledge retention performance for a comprehensive investigation. Specifically, $(\RN{1})$ we verify the knowledge extraction performance on world knowledge using natural questions (NQ) \cite{kwiatkowski-etal-2019-natural} (\ie NQ-open \cite{min2021neurips} in the closed-book setting) and report EM and F1 scores. We report the five-shot evaluation results using the first five QA pairs in the dev sets as prompts. $(\RN{2})$ we assess the reasoning capability on Commonsense knowledge using CommonsenseQA (CSQA) \cite{talmor-etal-2019-commonsenseqa}, employing accuracy to assess the correctness of the selected option, calculated by comparing the generated option against the gold option using EM. We present the five-shot performance on the dev sets, as the test set does not contain golden annotations, and use the first five multi-choice QA pairs in the training set as prompts. We use these two datasets because they were curated before the cut-off time of \textsc{Llama2} family models (\ie year 2022), making it likely that the models have obtained relevant knowledge in these datasets during the pre-training stage, as evidenced by \citet{touvron2023llama}.

\section{Implementation Details}
\label{sec:implement_details}
\begin{figure}[!t]
\centering
\includegraphics[width=0.95\linewidth]{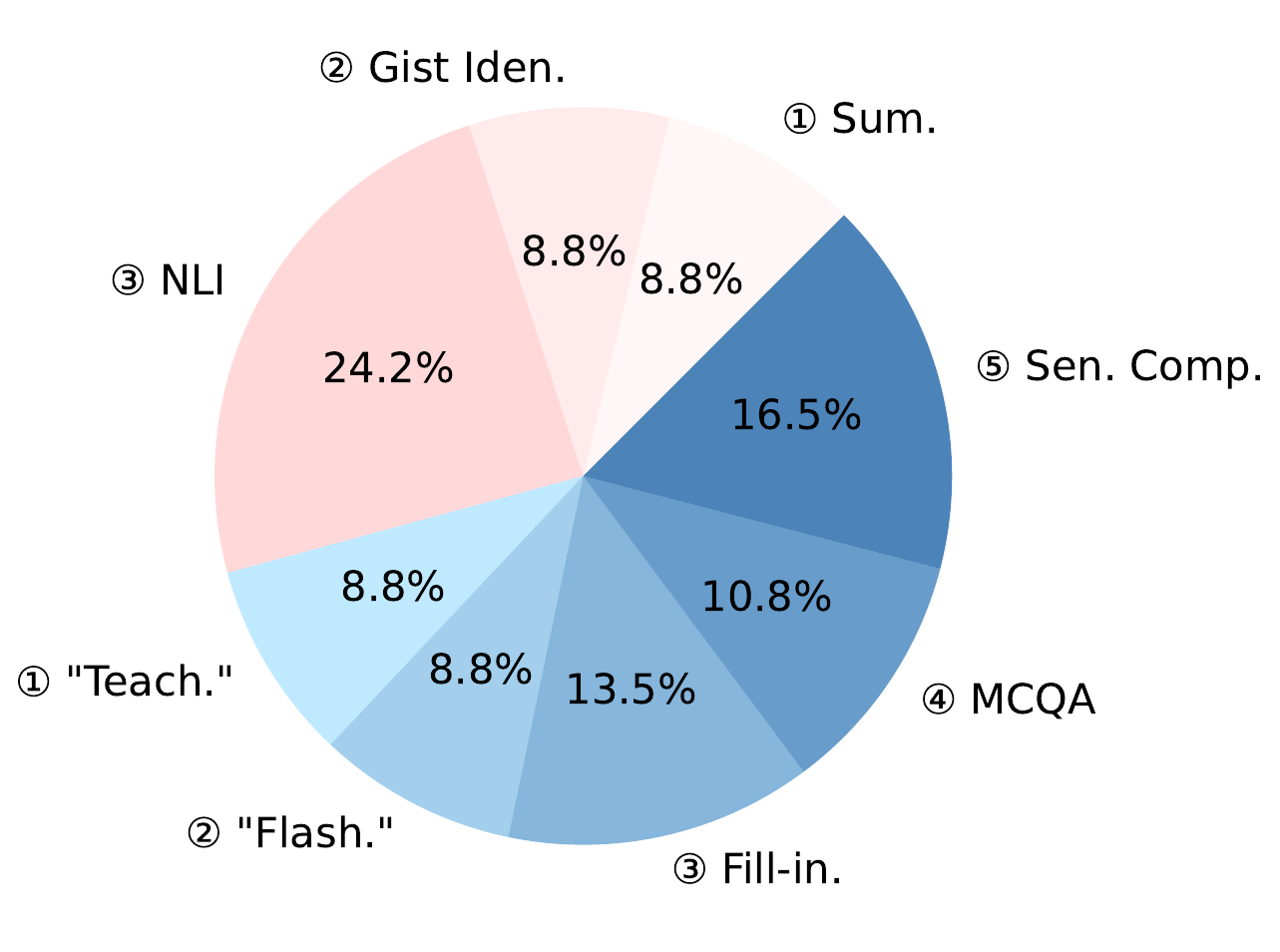}
\caption{The percentage of constructed examples of each task type in the self-teaching tasks on training documents in Wiki-Newpages-2023-10-Bio dataset.}
\label{fig:bio_task_percent}
\vspace{-3mm}
\end{figure}

\paragraph{Training Details.}

We utilize \llamaa{} for our investigation and provide analyses on Qwen2-7B, Mistral-7B-v0.1, Gemma-7B, \llamaathirteen{}, and \llamaachat{} for a comprehensive understanding. We use the following training objectives:
$(\RN{1})$ for training on document data $D^{Doc}$, we compute the standard next-token prediction loss by averaging over all tokens in the document $d$ (Equation \ref{eqa:x1}); $(\RN{2})$ for training on QA data $D^{QA}$, we compute the average
negative log-likelihood loss only on tokens in the answer $a$ given the question $q$ (Equation \ref{eqa:x2}), where $|d|$ and $|a|$ refer to the length of the tokenized document sequence and answer sequence, respectively.

\begin{align}
{L}_{\theta}(D^{Doc}) &=-\frac{1}{|d|} \sum_t \log p_{\theta}\left(d_t \mid d_{<t}\right)
\label{eqa:x1} \\
{L}_{\theta}(D^{QA}) &=-\frac{1}{|a|} \sum_t \log p_{\theta}\left(a_t \mid q, a_{<t}\right) 
\label{eqa:x2} 
\end{align}

We train \llamaa{}, Qwen2-7B, Mistral-7B-v0.1, Gemma-7B, and \llamaachat{} on 8 32GB Tesla V100 GPUs using a batch size of 8 and a learning rate of 5e-6. Additionally, we train \llamaathirteen{} on 8 A100-SXM4-40GB GPUs with a batch size of 8 and a learning rate of 5e-6. To ensure a fair comparison, all compared approaches train on the test documents for 3 epochs in total, regardless of the number of training stages. For continued pre-training, which is observed to struggle in grasping new knowledge, we train the models for 5 epochs. The specific number of training epochs used for each approach in Table \ref{tab:method_traing_stages_full} are as follows:
\begin{itemize}\setlength{\itemsep}{0pt}
    \item \textbf{Continued Pre-training} trains the model on the $D_{test}^{Doc}$ dataset for 5 epochs.
    \item \textbf{Standard Instruction-tuning} first trains on both $D_{train}^{Doc}$ and $D_{test}^{Doc}$ datasets, then fine-tunes on $D_{train}^{QA}$ dataset for 3 epochs.
    \item \textbf{PIT} \cite{jiang2024instructiontuned} first trains on $D_{train}^{QA}$ and $D_{train}^{Doc}$ datasets for 3 epochs, positioning the QA pairs right before the corresponding document texts, then trains on the $D_{test}^{Doc}$ data for 3 epochs.
    \item \textbf{\selfpt{}} (ours) first trains on $D_{train}^{QA}$ and $D_{train}^{Doc}$ with the created instruction-following dataset $D_{train}^{Self}$ (in the QA format) using the \selfteach{} strategy for 2 epochs, then continues training on $D_{test}^{Doc}$ data while reviewing the $D_{train}^{QA}$ data for 1 epoch, and finally continues training on $D_{test}^{Doc}$ data for 2 epochs. In addition, we provide the percentage of \selfteach{} task examples on training documents in Wiki-Newpages-2023-10-Bio dataset in Figure \ref{fig:bio_task_percent}.
\end{itemize}

Specifically, in the cross-domain setting, where there is a substantial difference between the domains of the training data and test documents, we continue training on $D_{test}^{Doc}$ data while reviewing the $D_{train}^{QA}$ data for 2 epochs after the initial training stage, followed by further training on $D_{test}^{Doc}$ data for 1 epoch. Furthermore, we adopt the same training strategy when dealing with \llamaachat{}, where the process of knowledge injection poses a significant challenge, as demonstrated by our experimental results. In accordance with \citet{jiang2024instructiontuned}, for PIT and \selfpt{}, we include 64 examples and 128 examples randomly sampled from $D_{train}^{QA}$ datasets, respectively, during the final training stages when solely training on the $D_{test}^{Doc}$ data, to prevent the model from losing its question-answering capabilities. It is important to note that all evaluation results are reported at the temperature $T=1$.

\paragraph{Training Details for \selfpt{} Variants.} 
\begin{itemize}\setlength{\itemsep}{0pt}
    \item \textbf{\selfpt{} w/o Review} first trains on $D_{train}^{QA}$ and $D_{train}^{Doc}$ with the created instruction-following dataset $D_{train}^{Self}$ (in the QA format) using the \selfteach{} strategy for 2 epochs, then continues training on $D_{test}^{Doc}$ data for 3 epochs.
    \item \textbf{\selfpt{} via Read.} initially trains on $D_{train}^{QA}$ and $D_{train}^{Doc}$ (in the read-comprehension format, as shown in Table \ref{tab:train_doc_read_compre_format} for 3 epochs, then trains on the $D_{test}^{Doc}$ data for 3 epochs.
    \item \textbf{\selfpt{} w/ Pre-Review} first trains on $D_{train}^{QA}$ and $D_{train}^{Doc}$ with the created instruction-following dataset $D_{train}^{Self}$ (in the QA format) using the \selfteach{} strategy for 2 epochs, then continues training on $D_{train}^{Doc}$ and $D_{train}^{QA}$ data for 1 epoch, and finally continues training on $D_{test}^{Doc}$ data for 3 epochs.
\end{itemize}

\paragraph{Training Details for Additional Compared Methods.} 
\begin{itemize}\setlength{\itemsep}{0pt}
    \item \textbf{Standard Instruction-Tuning w/o Forgetting} initially trains on the mixture of $D_{train}^{Doc}$ and $D_{test}^{Doc}$ for 3 epochs, then on $D_{train}^{QA}$ and $D_{test}^{Doc}$ datasets for 1 epoch.
    \item \textbf{PIT$^{++}$} \cite{jiang2024instructiontuned} initially trains on $D_{train}^{QA}$ for 1 epoch, then on $D_{train}^{QA}$ and $D_{train}^{Doc}$ datasets for 3 epochs, with the QA pairs placed right before the corresponding document texts, and finally, it trains on the $D_{test}^{Doc}$ data for 3 epochs.
    \item \textbf{Mixed Training} trains on mixture of the $D_{train}^{Doc}$, $D_{test}^{Doc}$ and $D_{train}^{QA}$ datasets simultaneously for 3 epochs.
\end{itemize}

Future research could explore the inclusion of segments from general domain datasets, such as Wiki data~\cite{zhang-etal-2024-self} and the Massive Multitask Language Understanding (MMLU)~\cite{hendrycks2021measuringmassivemultitasklanguage}, which were compiled prior to the pre-training cut-off date. Adopting this strategy may improve the model's capacity to retain learned knowledge and skills while reducing the risk of overfitting to novel information. In our current study, we deliberately avoid integrating extra data to ensure a precise assessment of knowledge injection, thereby preventing any biases that might arise from the inclusion of additional sources.

\paragraph{Prompts Employed in this Study.} 
The prompts used for constructing the QA datasets for open-ended generation and NLI tasks are presented in Table \ref{tab:wikiqa_gen_prompt} and Table \ref{tab:wikiqa_nli_prompt}, respectively. The prompt used during the evaluation process is displayed in Table \ref{tab:prompt_five_shot}. The prompt used by GPT-4 for annotating QA types in the open-ended generation tasks of the Wiki-Newpages-2023-QA datasets is presented in Table \ref{tab:wikiqa_type_prompt}.

\begin{table*}[th]
\centering
\fontsize{10}{10}\selectfont
\begin{tabular}{p{15.5cm}}
\toprule
\textbf{The prompt utilized by GPT-4 for building QA datasets for open-ended generation tasks} \\
\midrule
Below is a paragraph about the 51st International Emmy Awards ceremony. Your task is to formulate a detailed list of questions and corresponding answers that encompass all the information within the paragraph. To ensure clarity, each question should explicitly mention the 51st International Emmy Awards ceremony. Answers should be concise, consisting of a few short phrases separated by commas. For instance:\\
Paragraph: The 51st International Emmy Awards ceremony, presented by the International Academy of Television Arts and Sciences (IATAS), occurred on November 20, 2023, at the New York Hilton Midtown in New York City. It was held to acknowledge the best television programs initially produced and aired outside the United States in 2022. Nominations were announced on September 26, 2023.\\
Question: When was the 51st International Emmy Awards ceremony held?\\
Answer: November 20, 2023.\\
Question: Who was responsible for presenting the 51st International Emmy Awards ceremony?\\
Answer: The International Academy of Television Arts and Sciences (IATAS).\\
Question: Where was the 51st International Emmy Awards ceremony held? \\
Answer: The New York Hilton Midtown in New York City.\\
Question: What was the purpose of the 51st International Emmy Awards ceremony? \\
Answer: To recognize the best television programs initially produced and aired outside the United States in 2022.\\
Question: When were the nominations for the 51st International Emmy Awards announced? \\
Answer: September 26, 2023.\\
Below is a paragraph about \verb|{topic}|. Your task is to formulate a detailed list of questions and corresponding answers that encompass all the information within the paragraph. To ensure clarity, each question should explicitly mention \verb|{topic}|. Answers should be concise, consisting of a few short phrases separated by commas. For instance:\\
Paragraph: \verb|{paragraph}|\\
Question:\\
\bottomrule
\end{tabular}
\caption{The prompt utilized by GPT-4 for building QA datasets for open-ended generation tasks based on the gathered Wiki-Newpages documents.}
\label{tab:wikiqa_gen_prompt}
\end{table*}
\begin{table*}[th]
\centering
\fontsize{10}{10}\selectfont
\begin{tabular}{p{15.5cm}}
\toprule
\textbf{The prompt utilized by GPT-4 for building QA datasets for natural language inference tasks} \\
\midrule
Below is a paragraph about Luis Hugo Hernán Palma Pérez. Your task is to formulate a detailed list of natural language inference tasks with questions and corresponding answers based on the paragraph. For instance: \\
Paragraph: Luis Hugo Hernán Palma Pérez (born November 3, 1958) is a Chilean surgeon and politician, founding member of the Humanist Party of Chile. He is a deputy for the period 2022-2026, after being elected in the 2021 Chilean parliamentary elections. \\
Question: Based on the paragraph above can we conclude that Luis Hugo Hernán Palma Pérez was born in November. \\
Options: \\
- Yes \\
- It's impossible to say \\
- No \\
Answer: Yes \\
Question: Based on the paragraph above can we conclude that Luis Hugo Hernán Palma Pérez is a deputy for the period 2020-2024. \\
Options: \\
- Yes \\
- It's impossible to say \\
- No \\
Answer: No \\
Question: Based on the paragraph above can we conclude that The Humanist Party of Chile is a political party in Chile. \\
Options: \\
- Yes \\
- It's impossible to say \\
- No \\
Answer: Yes \\
Question: Based on the paragraph above can we conclude that Luis Hugo Hernán Palma Pérez is a dentist. \\
Options: \\
- Yes \\
- It's impossible to say \\
- No \\
Answer: No \\
Question: Based on the paragraph above can we conclude that Luis Hugo Hernán Palma Pérez was elected in the 2021 Chilean parliamentary elections. \\
Options: \\
- Yes \\
- It's impossible to say \\
- No \\
Answer: Yes \\
Below is a paragraph about \verb|{topic}|. Your task is to formulate a detailed list of natural language inference tasks with questions and corresponding answers based on the paragraph. For instance: \\
Paragraph: \verb|{paragraph}| \\
Question: \\

\bottomrule
\end{tabular}
\caption{The prompt utilized by GPT-4 for building QA datasets for natural language inference tasks based on the gathered Wiki-Newpages documents.}
\label{tab:wikiqa_nli_prompt}
\end{table*}
\begin{table*}[th]
\centering
\fontsize{10}{10}\selectfont
\begin{tabular}{p{15.5cm}}
\toprule
\textbf{The five-shot prompt used for assessing open-ended generation tasks} \\
\midrule
Question: Which animated film is included in the list of characters in the Zootopia franchise? \\
Answer: The animated film "Zootopia" (2016). \\

Question: Who were the coaches in The Voice Generations (Philippine TV series)? \\
Answer: Billy Crawford, Chito Miranda, Julie Anne San Jose, and Stell of SB19. \\

Question: Who is Cyrelle Saut? \\
Answer: A futsal and football player who has been associated with Tuloy Foundation and the Azkals Development team. \\

Question: What team does the 2023 Southern Miss Golden Eagles football team represent? \\
Answer: The University of Southern Mississippi. \\

Question: When was Kenneth Mitchell (basketball) born? \\
Answer: October 1, 1975. \\
\bottomrule
\end{tabular}
\caption{The five-shot prompt used for assessing open-ended generation tasks, which is derived from the gathered Wiki-Newpages-2024-03 documents.}
\label{tab:prompt_five_shot}
\end{table*}

\begin{table*}[th] 
\centering 
\fontsize{10}{10}
\selectfont \begin{tabular}{p{15.5cm}} 
\toprule 
\textbf{The prompt used by GPT-4 for annotating QA types in the open-ended generation tasks of the Wiki-Newpages-2023-QA datasets} \\
\midrule 
Below is a paragraph along with corresponding question and answer pairs. Your task is to analyze the paragraph and the question-answer pairs by categorizing the type of information they inquire about or provide. Use concise phrases to describe each category. For example:\\
Paragraph: <Andrew Turner (rugby union, born 2002) - Wikipedia> Andrew Turner (born 16 February 2002) is an English rugby union player, currently playing for the and . His preferred position is prop.\\
Question: When was Andrew Turner (rugby union, born 2002) born?\\
Answer: February 16, 2002.\\
Question: What nationality is Andrew Turner (rugby union, born 2002)?\\
Answer: English.\\
Question: What sport does Andrew Turner (rugby union, born 2002) play?\\
Answer: Rugby union.\\
Analysis: Types of question-answer pairs: (1) Birth date, (2) Nationality, (3) Sport/Profession.\\
Types of the paragraph: Biography - Biographical information about Andrew Turner, a rugby union player born in 2002, including his birth date, nationality, sport, and preferred position.\\
Below is a paragraph along with corresponding question and answer pairs. Your task is to analyze the paragraph and the question-answer pairs by categorizing the type of information they inquire about or provide. Use concise phrases to describe each category. For example:\\
Paragraph: \verb|{paragraph}|\\
\verb|{QA}|\\
Analysis: \\
\bottomrule 
\end{tabular} 
\caption{The prompt used by GPT-4 for annotating QA types in the open-ended generation tasks of the Wiki-Newpages-2023-QA datasets.} 
\label{tab:wikiqa_type_prompt} 
\end{table*}

\begin{table*}[th]
\centering
\fontsize{10}{10}\selectfont
\begin{tabular}{p{15.5cm}}
\toprule
\textbf{The prompt utilized by GPT-4o for building QA datasets for open-ended generation tasks} \\
\midrule
Your task is to rephrase the paragraph below to make it clearer and more concise. Then, create a detailed list of questions and corresponding answers that cover the factual information in the revised content. Answers should be concise, consisting of a few short phrases separated by commas. For example: \\
Paragraph: \\
6 VCs explain how startups can capture and defend marketshare in the AI era. Ninety-four percent of business leaders agree AI will be critical to all businesses’ success over the next five years, and total global spending on AI is expected to reach \$154 billion by the end of this year, a 27\% increase from 2022. \\
Revised Content: \\
Six venture capitalists (VCs) explain how startups can capture and defend market share in the AI era. Ninety-four percent of business leaders agree that AI will be critical to the success of all businesses over the next five years. Additionally, total global spending on AI is expected to reach \$154 billion by the end of this year, representing a 27\% increase from 2022. \\
Simple Question-Answering Pairs: \\
Question: How many VCs explain how startups can capture and defend market share in the AI era? \\
Answer: Six venture capitalists (VCs). \\
Question: What percentage of business leaders agree that AI will be critical to the success of all businesses over the next five years? \\
Answer: Ninety-four percent. \\
Question: Over what period do business leaders believe AI will be critical to all businesses' success? \\
Answer: Over the next five years. \\
Question: How much is the total global spending on AI expected to reach by the end of this year? \\
Answer: \$154 billion. \\
Question: By what percentage is the global spending on AI expected to increase from 2022? \\
Answer: Twenty-seven percent. \\
Your task is to rephrase the paragraph below to make it clearer and more concise. Then, create a detailed list of simple questions and corresponding answers that cover the information in the revised content. Answers should be concise, consisting of a few short phrases separated by commas. For example: \\
Paragraph: \\
\verb|{paragraph}| \\
Revised Content: \\
\bottomrule
\end{tabular}
\caption{The prompt utilized by GPT-4o for building QA datasets for open-ended generation tasks based on the gathered WebNews documents.}
\label{tab:news_gen_prompt}
\end{table*}

\section{Implementation Details of Evaluation on Varying Models and Corpora}
\label{sec:evaluate_varied}

\subsection{Evaluation on Different Models}
For the evaluation using different models, specifically Qwen2-7B~\cite{yang2024qwen2technicalreport}, we collected articles published on Wikipedia NewPages from June 2024 to September 2024 to minimize overlap with the pre-training corpus. We randomly selected 146 biographies from the collected articles, following the data construction pipeline described in Appendix~\ref{sec:details_data_construction}, to create a new question-answering dataset for an open-ended generation task. This resulted in a test set, named WikiBio-2024, comprising 146 documents and a total of 827 QA pairs.

\subsection{Evaluation on Varied Corpora}
For the evaluation using varied corpora, we utilized the news data collected by \citet{tang2024multihopragbenchmarkingretrievalaugmentedgeneration} using the mediastack API\footnote{\url{https://mediastack.com/}}. Specifically, this dataset includes articles published from September 26, 2023, to December 26, 2023, which is beyond the pre-training cutoff time of \llamaa{}. The dataset covers a range of news categories, such as entertainment, business, technology, and science.

For each factual sentence extracted from the original articles by~\citet{tang2024multihopragbenchmarkingretrievalaugmentedgeneration}, we concatenated the article title with the fact to create a knowledge snippet. Following \citet{tang2024multihopragbenchmarkingretrievalaugmentedgeneration}, we used GPT-4o~\cite{gpt4o} (version dated 2024-02-01) to first paraphrase these snippets to make them clearer and more concise, and then generate relevant QA pairs. The prompt utilized can be found in Table~\ref{tab:news_gen_prompt}. Using the data construction pipeline described in Appendix~\ref{sec:details_data_construction}, we generated a new training set, \ie WebNews-2023, consisting of 1,800 training documents and 6,038 QA pairs, as well as a testing set with 400 testing documents and 1,350 QA pairs.

\section{A Sample Training Document in the Reading-Comprehension Format} 
\label{sec:train_read_format}
Drawing inspiration from \citet{cheng2024adapting}, we restructure the training document in the reading-comprehension text format. Each raw text is enriched with a series of tasks related to its content, constructed using our proposed \selfteach{} strategy. An example of a training document is provided in Table \ref{tab:train_doc_read_compre_format}.

\begin{table*}[!t]
\centering
\fontsize{10}{10}\selectfont
\begin{tabular}{p{15.5cm}}
\toprule
\textbf{A training document example in the reading-comprehension format} \\
\midrule
<Robert Anderson (artist)  - Wikipedia> Robert Alexander Anderson (born 1946) is an American portrait artist known for painting the official portraits of George W. Bush and Alan Greenspan as well as designing United States postage stamps.\\

Answer the questions based on the article:\\

Question: Write a title:\\
Answer:Robert Anderson (artist)\\

Question: Highlight the key information within the article:\\
Answer:United States; American; Alan Greenspan; George W. Bush; Robert Alexander Anderson; 1946\\

Question: Based on the article above can we conclude that\\
<Robert Anderson (artist)> Robert Alexander Anderson (born 1946) is an American portrait artist known for painting the official portraits of George W. Bush and Alan Greenspan as well as designing United States postage stamps.\\

Options:\\
- Yes\\
- It's impossible to say\\
- No\\

Answer: Yes\\

Question: Tell me about Robert Anderson (artist).\\
Answer:Robert Alexander Anderson (born 1946) is an American portrait artist known for painting the official portraits of George W. Bush and Alan Greenspan as well as designing United States postage stamps.\\

Question: Generate a concrete description about Robert Anderson (artist) based on the following keywords:\\
United States; American; Alan Greenspan; George W. Bush; Robert Alexander Anderson; 1946\\
Answer:Robert Alexander Anderson (born 1946) is an American portrait artist known for painting the official portraits of George W. Bush and Alan Greenspan as well as designing United States postage stamps.\\

Question: <Robert Anderson (artist)> Robert Alexander Anderson (born 1946) is an American -- known for painting the official portraits of George W. Bush and Alan \\Greenspan as well as designing United States postage stamps.\\

Answer: Portrait artist.\\

Question: <Robert Anderson (artist)> - (born 1946) is an American portrait artist known for painting the official portraits of George W. Bush and Alan Greenspan as well as designing United States postage stamps.\\
Options:\\
- Alan Greenspan\\
- 1946\\
- Robert Alexander Anderson\\
- George W. Bush\\
Answer:Robert Alexander Anderson\\

Question: <Robert Anderson (artist)> Robert Alexander Anderson (born 1946) is an American portrait artist known for painting the official portraits of George W. Bush and Alan Greenspan as well as:\\
Answer:designing United States postage stamps\\

\bottomrule
\end{tabular}
\caption{An example of a training document from the Wiki-Newpages-2023-10-Bio train set, presented in a reading-comprehension format.}
\label{tab:train_doc_read_compre_format}
\end{table*}

\label{sec:appendix}

\end{document}